\documentclass[10pt,twocolumn,letterpaper]{article}

\usepackage{cvpr}              %
\newcommand{\metric}{\textcolor{black}{ID-Sim}}

\usepackage[table]{xcolor}
\definecolor{incOne}{HTML}{4E95D9}
\definecolor{incTwoA}{HTML}{94D4D4}
\definecolor{incTwoB}{HTML}{ECCA89}

\expandafter\def\expandafter\normalsize\expandafter{%
    \normalsize%
    \setlength\abovedisplayskip{-1pt}%
    \setlength\belowdisplayskip{8pt}%
    \setlength\abovedisplayshortskip{-8pt}%
    \setlength\belowdisplayshortskip{2pt}%
}

\definecolor{cvprblue}{rgb}{0.21,0.49,0.74}
\usepackage[pagebackref,breaklinks,colorlinks,allcolors=cvprblue]{hyperref}
\usepackage{booktabs}   %
\usepackage{makecell}   %
\usepackage{pifont}
\usepackage{multirow}
\usepackage{float}
\usepackage{caption}
\usepackage{subcaption}
\newcommand{\xmark}{\ding{55}}

\title{ID-Sim: An Identity-Focused Similarity Metric}

\vspace{-1mm}
\author{
Julia Chae$^{1,\dagger}$ \quad
Nicholas Kolkin$^{2}$ \quad
Jui-Hsien Wang$^{2}$ \quad
Richard Zhang$^{2}$ \quad
Sara Beery$^{1, *}$ \quad
Cusuh Ham$^{2, *}$\\[0.6em]
$^{1}$MIT CSAIL \quad\quad
$^{2}$Adobe Research
}
\begin{document}

\twocolumn[{%
    \renewcommand\twocolumn[1][]{#1}%
    \maketitle
    \begin{center}
        \centering
        \captionsetup{type=figure}
        \vspace{-20pt}
        \includegraphics[width=\textwidth]{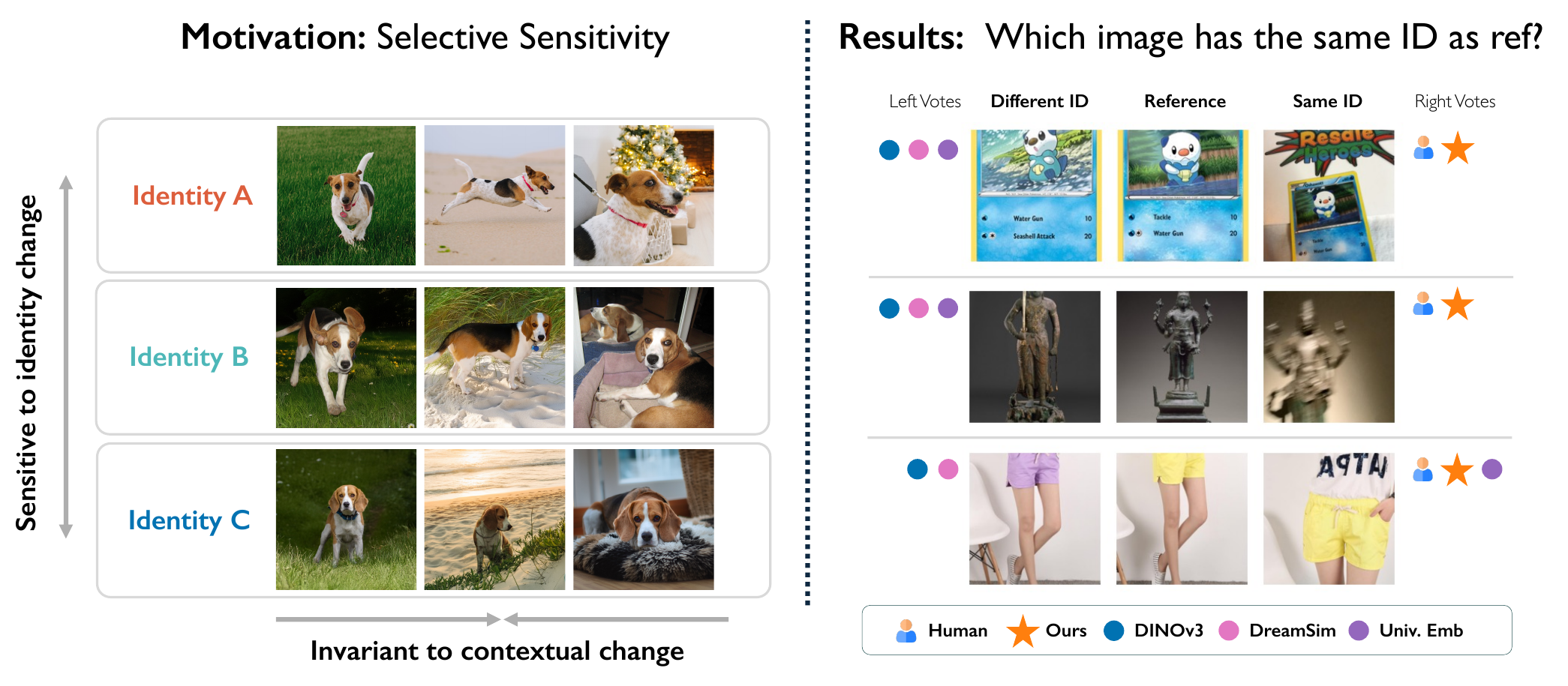}
        \vspace{-16pt}
        \captionof{figure}{\textbf{ID-Sim motivation \& results.} (Left) An identity-focused metric should exhibit \textit{selective sensitivity:} invariant to contextual changes (e.g. background, pose, lighting), yet sensitive to subtle identity-altering changes. (Right) We present \textsc{\metric{}}, which captures this property more effectively than existing metrics, and achieves strong improvements across a diverse set of identity-focused tasks. }
        \label{fig:teaser}
    \end{center}
    
}]

\setcounter{footnote}{0}
\renewcommand{\thefootnote}{\fnsymbol{footnote}}
\footnotetext{$^{*}$Equal advising, randomly ordered.}
\footnotetext{$^{\dagger}$Work done while at Adobe as an intern.}

\begin{abstract}
Humans have remarkable \textit{selective sensitivity} to identities--easily distinguishing between highly similar identities, even across significantly different contexts such as diverse viewpoints or lighting. Vision models have struggled to match this capability, and progress towards identity-focused tasks such as personalized image generation is slowed by a lack of identity-focused evaluation metrics. To help facilitate progress, we propose \textbf{\textsc{\metric}}, a feed-forward metric designed to faithfully reflect human selective sensitivity. To build \metric, we curate a high-quality training set of images spanning diverse real-world domains, augmented with generative synthetic data that provides controlled, fine-grained identity and contextual variations. We evaluate our metric on a new unified evaluation benchmark for assessing consistency with human annotations across identity-focused recognition, retrieval, and generative tasks. Our project page is \href{https://juliachae.github.io/id_sim.github.io/}{here}.
\end{abstract}

\vspace{-8pt}
\section{Introduction}

Humans readily recognize the same individual or object across large variations in viewpoint, illumination, pose, and context while remaining highly sensitive to subtle differences that signal identity changes \cite{dicarlo2007untangling,biederman1987recognition,logothetis1995psychophysical,palmeri2004visual}. This balance, which we term \textit{selective sensitivity}, enables both robust generalization and fine-grained discrimination--we recognize a familiar character from an unusual angle, identify a personal item under new lighting, or pick our own pet out of a crowd~\cite{rosch1975cognitive,biederman1987recognition,palmeri2000role}.
From a cognitive perspective, this corresponds to learning representations in which diverse appearances of the same identity cluster tightly while distinct identities remain well separated~\cite{dicarlo2007untangling,logothetis1995psychophysical}.

Existing ``identity''- or ``instance''-focused works in computer vision employ widely varying definitions of what this means, from broad semantic categories (e.g., cities or product types) to unique physical objects. To reduce ambiguity, we adopt a specific, property-based definition for this work. We first define the concept of \textit{visual identity}, and then use it to define an \textit{instance}. 
\vspace{-1mm}
\begingroup
\setlength{\fboxsep}{6pt} %
\begin{center}
\fbox{
\begin{minipage}{0.83\linewidth}
\textbf{Visual identity:} An object's unique set of intrinsic visual
properties (e.g., shape, texture, color). \\
\textbf{Instance:} Objects sharing the same visual identity.
\end{minipage}
}
\end{center}
\endgroup
\vspace{-1mm}

Despite remarkable progress in visual representation learning, vision systems still struggle with identity-focused tasks. Even foundation models trained on massive datasets \cite{siméoni2025dinov3, radford2021learning, ilharco_gabriel_2021_5143773} fail to recognize the same object under moderate transformations (e.g., changes in viewpoint or illumination) and confuse identities that share superficial visual features like the background (see examples in \Cref{fig:teaser}). Specialized systems for instance retrieval \cite{shao2023guie}, re-identification \cite{trein2025siamese, zheng2016person, adam2025wildlifereid}, or personalized evaluation \cite{peng2024dreambench, eldesokey2025mindtheglitchvisualcorrespondencedetecting}, address aspects of this challenge, but typically in narrow, domain-specific contexts. None provide a general measure of identity consistency that captures when a transformation preserves, versus alters, an identity.

Historically, advances in perceptual metrics have catalyzed progress in computer vision. The shift from signal-based measures (PSNR and SSIM \cite{wang2004image}) to learned perceptual metrics like LPIPS \cite{zhang2018unreasonable} and DISTS \cite{ding2020image} transformed how visual similarity is quantified, enabling models that better align with human judgments of appearance. However, these metrics are focused on \textit{appearance similarity}, not \textit{identity}.
To catalyze progress on identity-focused tasks, we propose a new perceptual metric 
that explicitly prioritizes selective sensitivity.
We curate a diverse, instance-level training dataset that unifies and extends existing benchmarks across domains, augmented with a generative editing pipeline for controlled identity-preserving and identity-altering transformations. We train our model using complementary global and local contrastive objectives to balance invariance and discrimination, and evaluate and analyze our metric across diverse identity-focused tasks. 

\noindent
Our main contributions are:
\begin{itemize}
    \item \textbf{A new identity-focused perceptual metric ID-Sim,} trained to mimic human selective sensitivity via curated real and synthetic instance-level data.

    \item A \textbf{comprehensive benchmark for identity perception}, combining existing instance-level tasks across domains with a new human-annotated generative evaluation dataset (\textsc{Subjects2k}).
    
    \item A \textbf{systematic sensitivity analysis} using controlled generative edits, revealing the influence of viewpoint, lighting, and contextual changes on perceived identity consistency.
    
\end{itemize}

\vspace{-0.4em}
\section{Related Works}
\label{sec:relatedworks}

\subsection{Identity-focused tasks}

\noindent\textbf{Re-identification (Re-ID)} aims to identify the same individual across contexts ~\cite{zheng2016person,schneider2019past}. Deep metric models for Re-ID are: (i) highly specialized to specific domains (e.g., animals~\cite{adam2025wildlifereid,otarashvili2024multispecies,schneider2022similarity}, humans~\cite{schroff2015facenet, deng2019arcface, wang2018cosface, liu2017sphereface,sun2018beyond, he2019foreground, song2019generalizable}), with models trained on domain X failing on domain Y ~\cite{shaked2024minimizing, ye2021deep,kim2024pose,otarashvili2024multispecies}, (ii)~require extensive domain-specific fine-grained annotations, and (iii)~optimize for discrimination (maximizing inter-class margins) rather than perceptual alignment (matching human similarity judgments). 

\vspace{3pt}
\noindent\textbf{Instance retrieval} 
entails finding matches to an example object from within a large candidate pool~\cite{zheng2017sift,chen2022deep}. Recent works like UnED~\cite{ypsilantis2023towards} and GPR-1200~\cite{gpr1200} have pushed towards generalizing instance retrieval across categories, from products to landmarks. Many prominent models train on data that conflate fine-grained classification with instance identity, which may limit their ability to differentiate two visually similar but distinct objects, as observed in \Cref{fig:teaser}. Related work \cite{wu2025instancelevelgenerationrepresentationlearning} explores training an instance-retrieval representation using generative edited data. While the approach is promising, the model is evaluated only on retrieval benchmarks and not on a broader range of identity-focused tasks.

\vspace{3pt}
\noindent\textbf{Personalized vision} works~\cite{sundaram2024personalizedrepresentationpersonalizedgeneration,jiang2025personalizedvisionvisualincontext, samuel2024whereswaldodiffusionfeatures} adapt large models to a user-specified concept for tasks like subject-driven generation~\cite{ruiz2023dreambooth, gal2022image, kumari2023multi, ham2024personalized, he2025conceptrol, tan2025ominicontrol, ye2023ip} or personalized segmentation~\cite{zhang2023personalizesegmentmodelshot}. Personalized generation faces a core challenge with identity fidelity, as models often struggle to faithfully preserve a subject's unique features. This failure of preservation also makes robust evaluation hard, creating a clear need for an approach that can reliably measure fine-grained identity similarity. Tasks like personalized segmentation (e.g., PerSAM~\cite{zhang2023personalizesegmentmodelshot}) pursue different goals, such as producing a pixel-level mask for the target subject, rather than quantifying its identity consistency.

\subsection{Visual similarity metrics} \label{sec:rw-metrics}

\noindent\textbf{Perceptual metrics.}
SSIM~\cite{wang2004image}, PSNR~\cite{hore2010image}, and other classical perceptual metrics~\cite{sampat2009complex,zhang2011fsim,wang2003multiscale} are hand-designed, and often fail to capture the complex nuances of human perceptual similarity~\cite{zhang2018unreasonable}.
Alternatively, learning-based methods (e.g., LPIPS~\cite{zhang2018unreasonable}, PieAPP~\cite{prashnani2018pieapp}, DreamSim~\cite{fu2023dreamsim}, DISTS~\cite{ding2020image}) show that embeddings from deep networks~\cite{krizhevsky2012imagenet, simonyan2014very}
can be calibrated or trained on perceptual judgments, and even align well with human perceptual judgments~\cite{zhang2018unreasonable}. This observation extends to other modalities, such as stereo~\cite{tamir2025makes} and audio~\cite{manocha2020differentiable}. DiffSim \cite{song2025diffsim} has also found that diffusion model features align well with human judgments of perceptual similarity. Since these metrics optimize for \textit{overall} similarity rather than identity consistency, they are influenced by contextual changes that are irrelevant for identity-focused tasks.

\vspace{3pt}
\noindent\textbf{Contrastive representations.}
The distance between contrastive representations is often used to quantify visual similarity. Vision models trained with self-supervised contrastive objectives~\cite{oord2018representation,wu2018unsupervised,hjelm2018learning,tian2020contrastive, he2020momentum, chen2020simple, grill2020bootstrap} learn by attracting representations of augmented views of the same image and repelling those of different images. Thus, the representations capture the broad semantics of an image while ignoring the effects of transformations used as positive augmentations. For example, SimCLR~\cite{chen2020simple} and MoCo~\cite{he2020momentum} use cropping, color jittering, and blurring, encouraging invariance to low-level global changes. Similarly, the DINO model family~\cite{caron2021emerging, oquab2023dinov2,siméoni2025dinov3} and CLIP~\cite{radford2021learning} apply contrastive learning at scale, with CLIP aligning images to text—often compressing fine-grained visual differences in favor of higher-level semantic similarity.

\vspace{3pt}
\noindent \textbf{Applications of visual similarity metrics.} Metrics aligned with human perception have been shown to benefit downstream tasks like segmentation and instance retrieval \cite{sundaram2024doesperceptualalignmentbenefit}. As mentioned above, another primary application is evaluating subject-driven generation, where identity fidelity is crucial. However, general perceptual metrics are often insufficient, as they can confuse high visual similarity (e.g., two similar purses in the same pose) with true identity preservation (e.g., the same purse in a different pose). MLLMs (e.g., GPT-4V \cite{openai2023gpt4v, peng2024dreambench++}) are also used and align well with human judgments, but they face issues with prompt sensitivity, stochasticity, and scalability \cite{shi2024judging}. This necessitates an efficient metric focused on instance-level identity. Concurrent work also proposed specialized metrics for detecting generative inconsistencies \cite{eldesokey2025mindtheglitchvisualcorrespondencedetecting}, but may falter under occlusion or lighting changes.

\section{Methods}
\label{sec:methods}
\subsection{Characterizing our definition of an instance}
\label{subsec:instance_def}

Under our definitions, two images depict the \textit{same instance} when they show visually indistinguishable objects, such as two factory-identical screwdrivers, even when these objects are transformed by \textit{extrinsic} variations (e.g., pose, viewpoint, or lighting).  
Conversely, two images depict \textit{different instances} if their visual identities differ, including clearly different objects, significant temporal changes (e.g., a kitten aging to a cat), and physical alterations (e.g., a repainted chair).

\subsection{Training data curation} \label{sec:dataset}
\begin{figure}[t]
  \centering
  \includegraphics[width=\linewidth]{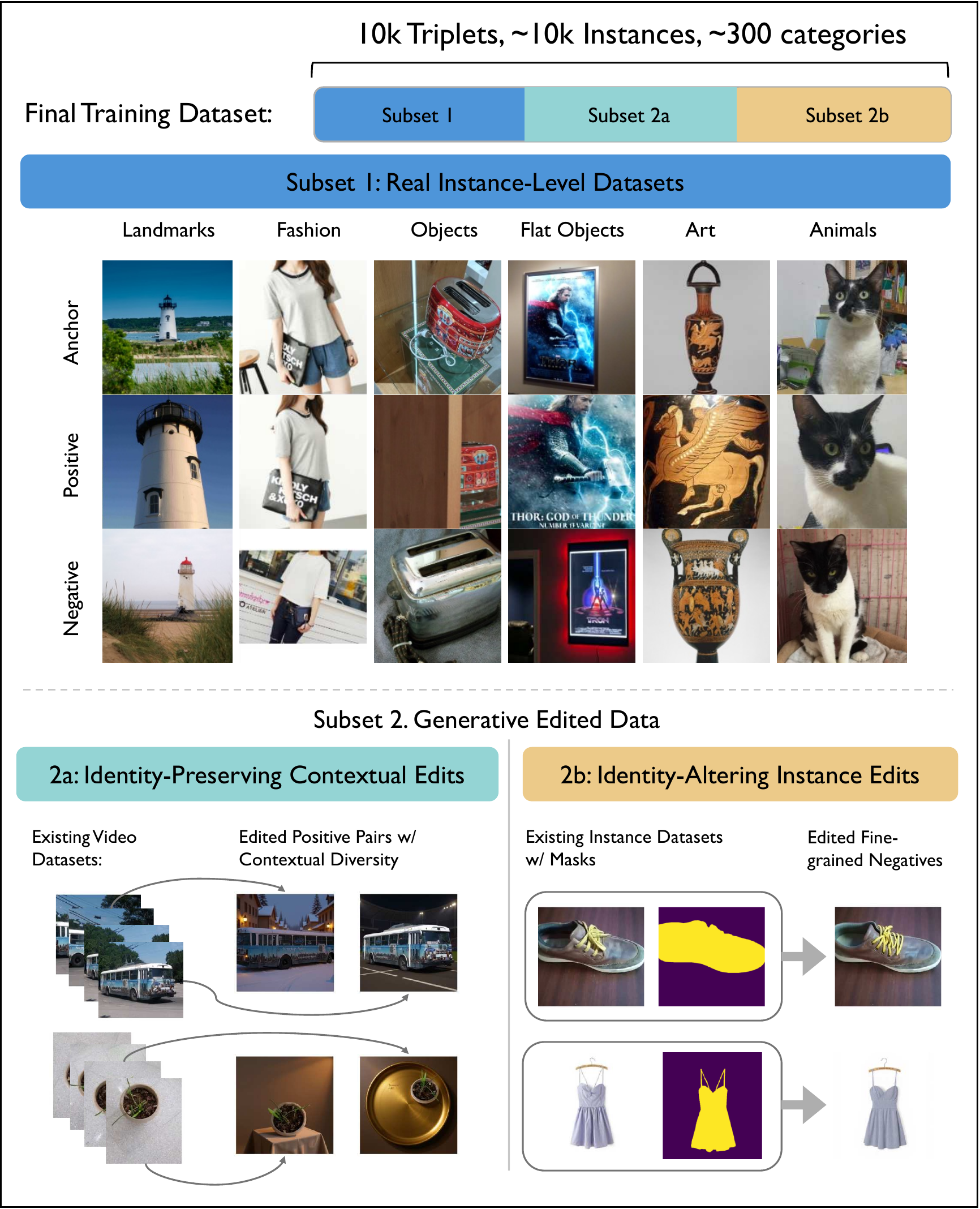}
  \caption{\textbf{Dataset curation pipeline.} We highlight the different real and synthetic data subsets that enable ID-Sim training. Together, they provide high context, domain, and visual identity diversities.}
  \label{fig:dataset-pipeline}
  \vspace{-3mm}
\end{figure}

\begin{table}[t]
\centering
\scriptsize
\setlength{\tabcolsep}{5.5pt}
\renewcommand{\arraystretch}{1.05}

\begin{tabular}{l c c c c c}
\toprule
\textbf{Dataset} & \textbf{Type} & \textbf{Objects} &
\textbf{\#Cat} & \textbf{Included in} & \textbf{\#Inst} \\
\midrule
ILIAS \cite{kordopatis2025ilias}      & Img & General     & N/A  & {\textbf{\textcolor{incOne}{S1}}}            & 281 \\
FORB \cite{wu2023forb}       & Img & Flat Obj    & 7    & {\textbf{\textcolor{incOne}{S1}}}            & 761 \\
MET \cite{ypsilantis2021met}        & Img & Artworks    & 1    & {\textbf{\textcolor{incOne}{S1}}}            & 226 \\
GLDv2 \cite{weyand2020google}     & Img & Landmarks   & 12   & {\textbf{\textcolor{incOne}{S1}}}            & 769 \\
Dogs \cite{dogfacenet}      & Img & Animal      & 1    & {\textbf{\textcolor{incOne}{S1}}}            & 494 \\
Cats \cite{Cermak_2024_WACV}      & Img & Animal      & 1    & {\textbf{\textcolor{incOne}{S1}}}            & 140 \\
DF2   \cite{ge2019deepfashion2}     & Img & Fashion     & 13   &
{\textbf{\textcolor{incOne}{S1}}}, {\textbf{\textcolor{incTwoB}{S2b}}}
& 2466 \\
UCO3D  \cite{liu24uco3d}    & Vid & General     & 146  &
{\textbf{\textcolor{incTwoA}{S2a}}}, {\textbf{\textcolor{incTwoB}{S2b}}}
& 3884 \\
LASOT  \cite{fan2020lasothighqualitylargescalesingle}    & Vid & General     & 34   & {\textbf{\textcolor{incTwoA}{S2a}}}          & 101 \\
YouTubeVIS \cite{yang2019videoinstancesegmentation}& Vid & General     & 35   & {\textbf{\textcolor{incTwoA}{S2a}}}          & 414 \\
GOT10k   \cite{Huang_2021}  & Vid & General     & 72   & {\textbf{\textcolor{incTwoA}{S2a}}}          & 604 \\
\bottomrule

\end{tabular}

\caption{\textbf{Overview of datasets used for training set curation.} \# Cat and \# Inst refer to number of categories and instances respectively. Colored text indicates different subsets that the dataset images appear in: \textcolor{incOne}{S1}, 
\textcolor{incTwoA}{S2a}, 
\textcolor{incTwoB}{S2b}, which can be matched to \Cref{fig:dataset-pipeline}}
\label{tab:datasets}
\vspace{-3mm}
\end{table}

\begin{figure*}[t]
\centering
\includegraphics[width=\textwidth]{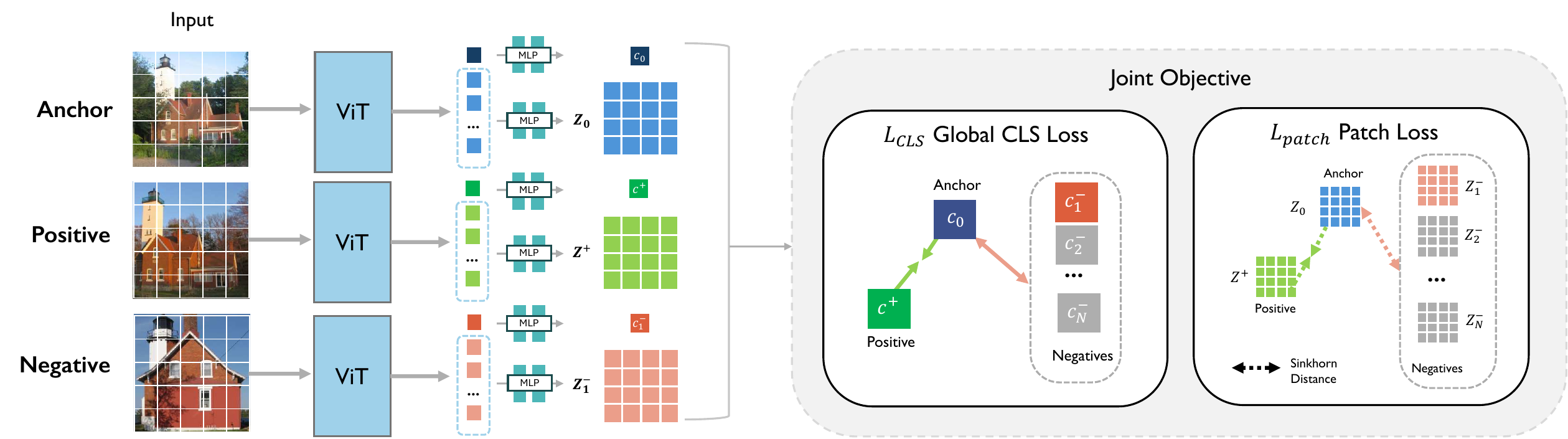}
\vspace{-6pt}
\caption{
\textbf{\metric~training pipeline.} We train our metric with dual contrastive supervision. At the global level, CLS-token projections for anchor–positive pairs are contrasted against one hard negative and additional batch negatives using InfoNCE. At the patch level, projected patch tokens are compared using Sinkhorn distance for the same instance pairs.}
\label{fig:training_overview}
\vspace{-6pt}
\end{figure*}
To train a metric that mimics selective sensitivity, we need data with three complementary signals: 

\begin{itemize}
    \item \textit{Context diversity} supporting invariance to different backgrounds, lighting, and viewpoints
    \item \textit{Visual identity diversity} enabling sensitivity to subtle appearance differences
    \item \textit{Domain diversity} ensuring generalization beyond specific categories
\end{itemize}

No existing datasets provide all three simultaneously, so we curate a training set using: \textbf{(Subset 1)} existing real instance-level datasets, and \textbf{(Subset 2)} synthetic data with: \textbf{(a)} \textit{contextual edits} that diversify the contexts in which instances appear, and \textbf{(b) }\textit{identity edits} that perturb visual identity (see \Cref{fig:dataset-pipeline}). These generative edits (S2a and S2b) expand the training pool, alleviating the limited diversity of real-world data, which is difficult to collect and annotate at scale.

We formulate our training data using triplets (an anchor image, a positive ID match, and a negative non-match), a standard structure for learning similarity metrics. Positives come from real instance images (S1) or identity-preserving contextual edits (S2a), while negatives come from different real instances (S1) or identity-altering edits (S2b). Our training set contains 10k triplets (30k images) spanning \(\sim\)10k instances across 10 datasets, with an even split between triplets containing only real images, generative identity-preserving positives with real negatives, and real positives with identity-altering negatives. \Cref{tab:datasets} provides an overview of the dataset composition. We analyze the effects of dataset scale and composition in \Cref{sec:data_balance}, and include additional experiments and full details on the source datasets, splits, and editing pipelines in the Supplemental~\ref{supp:training_data}.

\subsection{\metric~Training}
\label{sec:trainingsub}

\noindent \textbf{Data formulation for contrastive learning.} As seen in \Cref{fig:training_overview}, we follow the supervised contrastive learning framework \cite{khosla2021supervisedcontrastivelearning}, training our metric with positive (identity-preserving) pairs and negative (identity-breaking) pairs. We build our training batches from the dataset $\mathcal{D}$ introduced in \Cref{sec:dataset} comprised of $M$ instances $\{X_j\}_j^M$, and use these instances to curate triplets $(x_0, x^+, \{x_i^-\}_{i=1}^N)$ for training. Two images from the same instance are sampled as the anchor $x_0$ and positive $x^+$. We then sample a hard negative $x^-_1$, which may be either an identity-altering edit from S2b or a mined real negative from S1. For real negatives, we mine visually similar yet distinct instances using the nearest neighbors in the pretrained DINOv3 embedding space \cite{siméoni2025dinov3}. The remaining $N-1$ negatives $\{x^-_i\}_{i=2}^{N}$ are sampled from other instances within the batch.

\vspace{3pt}
\noindent \textbf{Joint objective.} We build upon a vision transformer (ViT) \cite{dosovitskiy2021imageworth16x16words} backbone $f_\theta$, following recent works \cite{oquab2024dinov2learningrobustvisual, radford2021learningtransferablevisualmodels, ilharco_gabriel_2021_5143773}. Each image is passed through $f_\theta$ to obtain the global $\textsc{CLS}$ token $c'$ and a set of patch tokens $Z'$. Since these representations capture complementary global and local information, we project them into separate embedding spaces using a dual-headed MLP: $c = \mathrm{MLP}_{\text{CLS}}(c')$ and $Z = \mathrm{MLP}_{\text{Patch}}(z')$. We train using a joint supervised contrastive objective that combines global and local terms:
\begin{equation}
\mathcal{L}_{\text{total}}
    = \mathcal{L}_{\text{CLS}}(c) \;+\; \lambda\, \mathcal{L}_{\text{Patch}}(Z),
\end{equation}

\noindent We opt for this joint objective instead of supervising only on the global token since patch embeddings provide complementary spatial signals for dense downstream tasks. 

\vspace{3pt}
\noindent \textbf{1. Global CLS Loss.} $\mathcal{L}_{\text{CLS}}$ is the standard InfoNCE objective \cite{oord2019representationlearningcontrastivepredictive} applied to the projected $\textsc{CLS}$ tokens:

\begin{equation}
\begin{aligned}
\mathcal{L}_{\text{CLS}} &=
-\log 
\frac{e^{\,s^{+}}}
     {e^{\,s^{+}} + \sum_{i=1}^{N} e^{\,s_i^{-}}}, \\[4pt]
s^{+} &= \mathrm{sim}(c_0,c^{+})/\tau,\quad
s_i^{-} = \mathrm{sim}(c_0,c_i^{-})/\tau
\end{aligned}
\label{eq:infonce}
\end{equation}

\noindent where $\mathrm{sim}(\cdot,\cdot)$ is cosine similarity, $\tau$ is a temperature parameter, and $c_0, c^{+}, c_i^-$ are the projected $\textsc{CLS}$ tokens for the anchor, positive, and $i$-th negative, respectively.

\vspace{3pt}
\noindent \textbf{2. Local patch loss.}
Patch tokens encode fine-grained local cues, but the spatial layouts of instances across images are often misaligned due to viewpoint or context changes, making direct position-wise comparisons unreliable. We therefore treat patch tokens between two images as an unordered set of local descriptors and measure their similarity via soft alignment. Given projected patch embeddings $A, B \in \mathbb{R}^{P \times D}$, we define their similarity as the negative entropically regularized optimal transport (OT) distance:

\begin{equation}
    \mathrm{sim}_{\text{patch}}(A,B) = -\,\mathcal{S}_\varepsilon(A,B),
\end{equation}
where $\mathcal{S}_\varepsilon$ is the Sinkhorn distance computed with uniform weights over patches, using \textsc{GeomLoss}~\cite{feydy2019interpolating}. 

Unlike DenseCL \cite{wang2021densecontrastivelearningselfsupervised}, which builds hard nearest-neighbor correspondences using augmented views of the same image, our objective operates across different images of the same instance, learning correspondences implicitly through a soft global OT plan. $\mathcal{L}_{\text{Patch}}$ is obtained by substituting $\mathrm{sim}_{\text{patch}}(\cdot,\cdot)$ into the InfoNCE objective.

\vspace{3pt}
\noindent \textbf{Using representations as an image similarity metric.} \label{sec:image-similarity} 
Using a trained $f_\theta$, similarity between images $x$ and $y$ can be measured as: 

\begin{equation}
    D(x, y; f_\theta) = 1 - \text{sim}\big(f_\theta(x),\, f_\theta(y)\big),
\end{equation} \label{eq:generic-similarity}
\noindent where the choice of feature representation $f_\theta(\cdot)$ and similarity function $sim(\cdot, \cdot)$ can vary. Since ID-Sim is a ViT-based metric, common features include the global CLS token or various patch token representations (e.g., aggregated or localized sets). The similarity function is typically cosine similarity. 
We explore alternative combinations of feature and similarity functions, which can enable different types of downstream tasks, in \Cref{sec:results}.

\vspace{-2mm}
\section{Experiments}

We evaluate \metric~against 7 baselines across instance \textit{recognition}, \textit{retrieval}, and \textit{preservation} tasks on 7 datasets, all disjoint from the training set.

\subsection{Experimental setup}
\noindent \textbf{Network architecture and training details.}
We select DINOv3 ViT-L~\cite{siméoni2025dinov3} at $448\times448$ resolution as the backbone $f_\theta$, chosen for strong instance-level performance on our validation set (described below). We freeze the backbone and finetune only: (i) lightweight 2-layer dual MLP projection heads, and (ii) rank 16 LoRA adapters \cite{hu2021loralowrankadaptationlarge} on attention and feedforward MLP layers. Training uses standard augmentations (color jitter, Gaussian noise, random cropping).

Hyperparameter tuning and checkpoint selection are performed on a held-out validation set drawn from the training data domains. We also construct an ``identity ablation set'', a small Flux-generated \cite{flux2024} synthetic dataset of 5 instances with identity-preserving and identity-altering edits. Full training details and ablation studies are in the Supplemental~\ref{supp:ablations_main}. 

\vspace{3pt}
\noindent \textbf{Baselines.}
We test 7 baselines in three categories: (1) \textit{perceptual metrics} (DreamSim \cite{fu2023dreamsim}, LPIPS \cite{zhang2018unreasonable}, DiffSim \cite{song2025diffsim}), (2) \textit{foundation models} (DINOv3 \cite{siméoni2025dinov3}, CLIP \cite{radford2021learning}, OpenCLIP \cite{ilharco_gabriel_2021_5143773}), and (3) an \textit{image retrieval model} -- the 1st-place solution \cite{shao2023guie} from Google's Universal Embedding (UNED) challenge \cite{ypsilantis2023towards}. All models use the ViT-L architecture except for \cite{shao2023guie} (larger ViT-H), DreamSim (ViT-B), and DiffSim (U-Net).

\begin{figure*}[t]
\centering
\includegraphics[width=\textwidth]{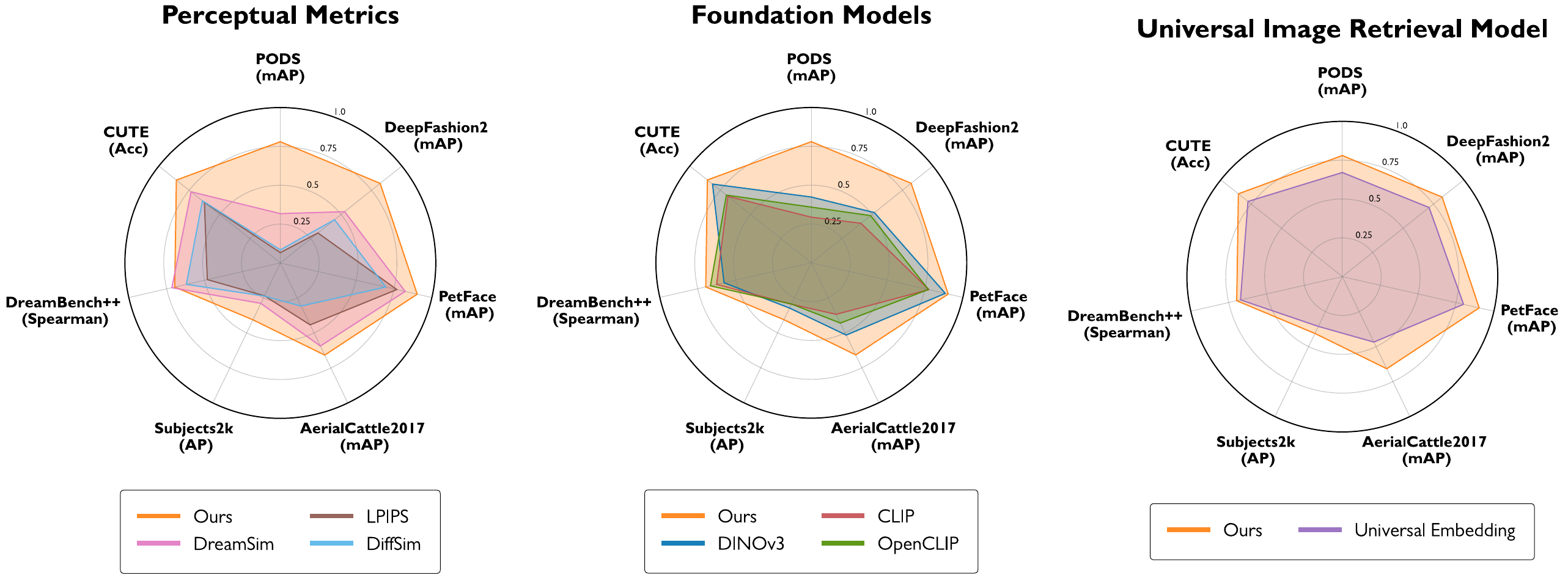}
\vspace{-6pt}
\caption{
\textbf{Performance of \metric~vs. baseline models}. We compare \metric~against standard perceptual metrics, large-scale vision foundation models, and a supervised “Universal Embedding” model (the top entry in Google’s universal embedding challenge). Across tasks -- instance retrieval, concept preservation, and re-identification -- \metric~consistently outperforms all baselines, including the instance-retrieval-focused model, despite using over 100× less labeled data and a smaller backbone (our \textsc{ViT-L} vs. \textsc{ViT-H}). Full results and seed variance are reported in the Supplemental~\ref{supp:full_results}. 
}
\label{fig:overview}
\vspace{-6pt}
\end{figure*}
\newcommand{\persamrow}[3]{#1 & #2 & #3\\}

\begin{table*}[t]
\centering
\scriptsize
\renewcommand{\arraystretch}{1.15}

\begin{subtable}[t]{0.28\textwidth}
\centering
{%
\setlength{\tabcolsep}{3pt}%
\begin{tabular}{lccc}
\toprule
\textbf{Method} & \textbf{Model} & \textbf{Subjects2k (AP)} & \textbf{DreamBench} \\
\midrule
Ours       & ViT-L   & \textbf{0.4063} & 0.697 \\
MLLM*      & GPT-4o  & 0.2901 & \textbf{0.748} \\
MLLM       & GPT-5   & 0.3159 & 0.3554 \\
MLLM       & Gemini  & 0.3354 & 0.70 \\
\bottomrule
\end{tabular}
}%
\caption{\textbf{Comparison with MLLMs on concept preservation.} MLLM* uses the original Subjects200K and DreamBench++ prompts and models respectively; MLLM rows use a controlled identity-preservation prompt for both datasets.}
\label{tab:mllm_comparison}
\end{subtable}
\hfill
\begin{subtable}[t]{0.40\textwidth}
\centering
{%
\setlength{\tabcolsep}{3pt}%
\begin{tabular}{l c c c c}
\toprule
\textbf{Dataset} & \textbf{Metric} & \textbf{DINOv3} & \textbf{Ours (no patch)} & \textbf{Ours} \\
\midrule
DF2       & mAP      & 0.4071 & 0.4765 & \textbf{0.7967} \\
AC2017   & mAP      & 0.4516 & 0.5471 & \textbf{0.6245} \\
CUTE               & Acc      & 0.6561 & 0.6439 & \textbf{0.8189} \\
DB++       & Spearman & 0.5479 & 0.5913 & \textbf{0.6834} \\
PetFace            & mAP      & 0.7849 & 0.8377 & \textbf{0.8446} \\
PODS               & mAP      & 0.5825 & \textbf{0.8181} & 0.7907 \\
S2k       & AP       & 0.2314 & 0.2348 & \textbf{0.3674} \\
\bottomrule
\end{tabular}
}%
\caption{\textbf{Patch-level Performance} of ID-Sim, ID-Sim without patch supervision, and DINOv3 across tasks}
\label{tab:patch_retrieval}
\end{subtable}
\hfill
\begin{subtable}[t]{0.24\textwidth}
\centering
{%
\setlength{\tabcolsep}{3pt}%
\begin{tabular}{lcc}
\toprule
\textbf{Method} & \textbf{mAP} & \textbf{F1} \\
\midrule
\persamrow{PerSAM + DINOv3}{0.153}{0.18}
\persamrow{PerSAM + Ours w/o patch sup}{0.214}{0.235}
\persamrow{PerSAM + Ours}{\textbf{0.436}}{\textbf{0.409}}
\bottomrule
\end{tabular}
}%
\caption{\textbf{Personalized segmentation (PerSAM)} performance on PODS with varying metrics.}
\label{tab:persam_main}
\end{subtable}
\vspace{-2mm}
\caption{
\textbf{Overview of results.} 
(Left): comparison with MLLMs on concept preservation. 
(Middle): performance across recognition and retrieval datasets. 
(Right): transfer to personalized segmentation with PerSAM.
}
\vspace{-2mm}
\label{tab:all_results}
\end{table*}

\subsection{Benchmarks}

\noindent\textbf{1. Concept preservation evaluation} aims to quantify how well a model is able to generate images of a reference instance while preserving its visual appearance. We evaluate this using two benchmarks. 

First, we report Spearman's $\rho$ correlation against human judgments on DreamBench++ \cite{peng2024dreambench}, a public benchmark for subject-driven generation. However, we found its human preference labels to be noisy, stemming from sparse annotations (see Supplemental~\ref{supp:subjects2k_pipeline}). Thus, we introduce \textsc{Subjects2k}, a new human-annotated subset of Subjects200k \cite{tan2024omini}. We collected new binary (same/different instance) human annotations to improve and evaluate the original dataset's GPT-4v \cite{openai2023gpt4v} labels. On \textsc{Subjects2k}, we report average precision (AP). 

\label{sec:dataset}
\begin{figure}[H]
  \centering
  \vspace{-2mm}
  \includegraphics[width=\linewidth]{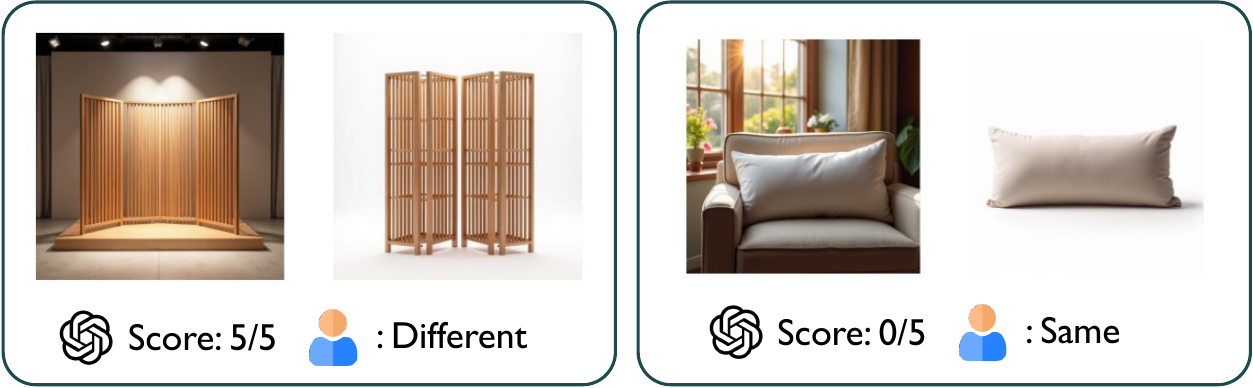}
  \caption{\textbf{Newly annotated Subjects2k.} We release a 2k high-quality human annotations with a subset of Subjects200k to serve as a new challenging concept preservation eval benchmark.}
  \label{fig:subjects2k}
  \vspace{-3mm}
\end{figure}

\noindent\textbf{2. Instance retrieval} tests the ability to find images of a given reference object from a pool of distractors. We report mean AP (mAP), averaged across each instance in the datasets on: (a) PODS \cite{sundaram2024personalizedrepresentationpersonalizedgeneration}, a dataset of household objects for instance-level retrieval and recognition under fixed distribution shifts, and (b) DeepFashion2 \cite{ge2019deepfashion2}, a fashion dataset designed to match in-store clothing items to in-the-wild consumer images.

\vspace{3pt}
\noindent\textbf{3. Re-identification (Re-ID) / instance classification} assesses whether individuals can be consistently recognized across viewpoints and conditions. We evaluate using: (a) mAP on PetFace \cite{shinoda2024petfacelargescaledatasetbenchmark}, a multi-species pet re-ID dataset, (b) mAP on AerialCattle \cite{Andrew_2021}, consisting of 23 individual cattle captured from aerial viewpoints, and, following the protocol from DiffSim, (c) accuracy on CUTE \cite{kotar2023cute}, where the model must identify which instance out of a pair of candidates matches an anchor object.

We also evaluate results on additional metrics (e.g. AUROC or NDCG for ranking \cite{wang2013theoreticalanalysisndcgtype}) for all tasks in the Supplemental~\ref{supp:full_results}.

\subsection{Results}

\label{sec:results}

\noindent \textbf{Improved identity-alignment across tasks.}
We evaluate ID-Sim across diverse domains and task types (\Cref{fig:overview}), using the global CLS token for similarity computation across all ViT-based methods. Across 49 evaluation setups, ID-Sim outperforms prior work in 48 cases. 

The strongest gains emerge along two axes of selective sensitivity: (1) \textit{Recognizing instances across contextual changes}, and (2) \textit{discriminating small visual identity changes}. This challenge of (1) is prominent in datasets such as PODS and DeepFashion2, where in addition to requiring fine-grained discrimination, positive instances are explicitly observed in different contexts (background, pose, and distractors in PODS; in-store vs in-the-wild for DeepFashion2). With \metric{}, we see some of the strongest relative improvements in these cases, with $+0.11$ and $+0.30$ gains in mAP over the second-best and the third-best models for both cases.
For (2), the Subjects2k benchmark presents some of the most challenging examples of fine-grained identity variation across datasets, with hundreds of visually similar negative instance pairs distinguished only by subtle details. On this benchmark, \metric{} outperforms the second-best metric by $+0.05$ mAP.

\noindent \textbf{Comparing metrics.} 
Across baselines, clear trends emerge in the strengths and limitations. Perceptual metrics generally underperform on identity-focused tasks, as they capture perceptual similarity rather than identity discrimination (though DreamSim performs best on DreamBench++, consistent with its human-aligned objective). Foundation models like DINOv3 perform well on datasets like CUTE and PetFace that primarily test identity similarity under lighting variations, but struggle to maintain identity similarity under other context shifts such as background variation, and also struggle with retrieval tasks. The Universal Embedding model achieves the second-strongest overall performance, but benefits from a larger backbone (ViT-H) and millions of labelled instance-level and fine-grained examples. 
\metric{} delivers consistently strong performance across all datasets, indicating broader generalization and a more unified notion of identity-alignment. 

\noindent \textbf{Comparison to MLLMs for concept preservation.} 
Multimodal LLMs (MLLMs) have shown strong potential for identity-based evaluation, often aligning more closely with humans than DINO or CLIP ~\cite{peng2024dreambench}. Therefore, we compare \metric{} against MLLMs using structured evaluation protocols consistent with prior work as shown in Table~\ref{tab:mllm_comparison}. As shown, \metric{} performs competitively and even surpasses MLLMs on Subjects2k, our more fine-grained concept-preservation benchmark.  Notably, MLLM performance is sensitive to prompt and model choice: DreamBench++ accuracy drops substantially when its original rubric-guided prompts are replaced with controlled identity-preservation prompts, whereas \metric{} remains stable across evaluations. MLLMs also introduce practical limitations, including stochastic outputs and reliance on pairwise comparisons that increase cost at scale, which is challenging for tasks like retrieval. In contrast, \metric{} provides deterministic, feed-forward evaluations that match or exceed MLLM performance with significantly lower computational overhead. Full prompting details and MLLM evaluation settings are provided in the Supplemental~\ref{supp:mllm_eval}.

\vspace{3pt}
\noindent \textbf{Beyond global similarity: Patch-level embeddings and localization power.}
While the global CLS token used in \Cref{fig:overview} captures a holistic representation, ViT patch tokens offer complementary, spatially localized features essential for fine-grained correspondence and region-level discrimination. We compare \metric{}'s patch embeddings against DINOv3~\cite{siméoni2025dinov3}, the strongest baseline with well-established patch embeddings, and ablate patch-level supervision to assess its contribution.

\Cref{tab:patch_retrieval} shows performance across tasks when similarity is computed using patch embeddings. \metric{} significantly outperforms DINOv3 across all datasets, indicating that it learns stronger and more discriminative local representations. While the variant trained only with CLS supervision improves performance by $13\%$ over DINOv3, explicit patch-level supervision substantially amplifies these gains, yielding a $40\%$ relative improvement.

To further assess whether our patch embeddings encode spatially meaningful information, we evaluate \metric{} within the state-of-the-art personalized segmentation framework, PerSAM~\cite{zhang2023personalizesegmentmodelshot}, which uses patch-token similarity to localize SAM point prompts and score segmentation predictions.  As shown in Table~\ref{tab:persam_main}, our patch features improve segmentation mAP significantly from 0.153 to 0.436 and F1 from 0.18 to 0.409 over DINOv3. Even without explicit patch supervision, \metric{} features improve over DINOv3 (0.214 mAP, 0.235 F1). Our patch embeddings capture both aggregated and spatially coherent information for precise localization and discrimination of identities.

\begin{table}[t]
\centering
\scriptsize
\setlength{\tabcolsep}{4pt} %
\vspace{2pt}
\begin{tabular}{lccccc}

\toprule
\textbf{Dataset} & \textbf{Bal.} & \textbf{Pos.} & \textbf{Neg.} & \textbf{Ratio} & \textbf{Val} \\
\textbf{Group} & & \textbf{Edit} & \textbf{Edit} & & \textbf{Score} \\
\midrule
All datasets         & \xmark & \xmark & \xmark & --    & 0.693 \\
All datasets         & \checkmark & \xmark & \xmark & --    & 0.752 \\
Filtered datasets    & \checkmark & \xmark & \xmark & --    & 0.890 \\
Filtered datasets    & \checkmark & \checkmark & \xmark & 1:1   & 0.937 \\
Filtered datasets    & \checkmark & \checkmark & \checkmark & 1:1:1 & \textbf{0.965} \\
\bottomrule
\end{tabular}
\caption{\textbf{Ablation of dataset composition and editing strategies.} Balancing and targeted editing of positive and negative samples improve performance.}
\label{tab:data_ablation}
\vspace{-6pt}
\end{table}

\subsection{Analysis}
\label{sec:results}
\begin{figure*}[t]
\centering
\includegraphics[width=\textwidth]{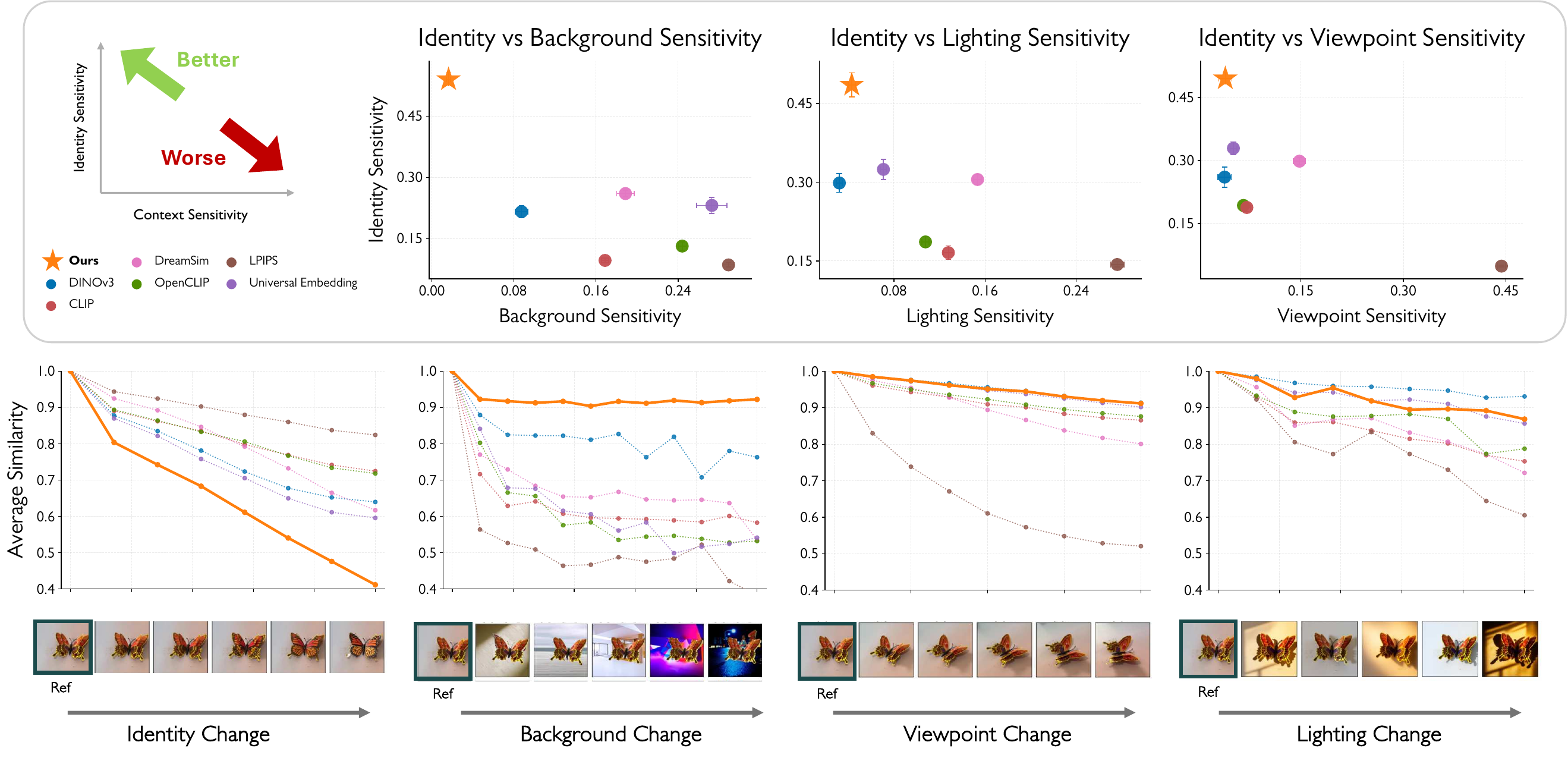}
\vspace{-6pt}
\caption{
\textbf{Selective sensitivity analysis.} We evaluate model sensitivity across four axes of visual change: identity, background, viewpoint, and lighting. For 100 anchor instances, we generate controlled variations and compute both sensitivity scores and similarity trends. \textbf{(Top row.)} Compared with baseline methods, our model is notably more sensitive to identity differences while remaining stable under background, viewpoint, and lighting changes. \textbf{(Bottom row.)} When systematically increasing variations across each dimension, we see that, as desired, only identity changes significantly reduce similarity measured by \metric{}.
}
\label{fig:sensitivity-analysis}
\vspace{-6pt}
\end{figure*}
\noindent \textbf{What makes for the best training data?}
\label{sec:data_balance}
While developing \metric, we systematically explored different strategies for curating and prioritizing high-value, identity-focused training data. Results are shown in \Cref{tab:data_ablation}, demonstrating that these choices significantly impact metric performance. We find that \textit{balanced composition is crucial}. Ensuring balanced positive and negative samples prevents overfitting to dominant instances and leads to more stable convergence. Additionally, \textit{dataset quality matters:} filtering out noisy or inconsistent instance-level samples significantly improves generalization. This matches prior literature--high-quality data is particularly vital for fine-grained tasks~\cite{cole2022does}.
Finally, we find that \textit{synthetic data boosts performance}: incorporating edited samples enhances both diversity and robustness—positive edits improve intra-instance consistency and edited negatives sharpen inter-instance discrimination.

\vspace{3pt}
\noindent \textbf{Exploring sensitivity to visual variation.}
In order to isolate the visual factors that metrics are most sensitive to, we conduct a systematic sensitivity analysis measuring how similarity scores change with respect to four dimensions of variation: identity, background, viewpoint, and lighting. We use 100 diverse objects from MVImgNet \cite{mvimagenet}, a multi-view dataset not used in training or evaluation, which provides 180 views per object on a clean surface with natural viewpoint variation. For the other dimensions, we apply generative edits: identity changes are simulated by editing the foreground with Qwen-Edit-Inpainting \cite{wu2025qwenimagetechnicalreport} (varying noise strengths), background changes via inpainting with 14 scene prompts, and lighting variations using Qwen-Edit \cite{wu2025qwenimagetechnicalreport} with nine illumination prompts. For each reference, we construct an edit grid varying jointly along identity and one other factor and compute the similarity of each image back to its original anchor. Sensitivity scores are then estimated by fitting a regression model to quantify the similarity decrease per unit change in each dimension. Final scores are averaged across instances, with uncertainty estimated via bootstrapped confidence intervals.

 \vspace{-1mm}
Figure \ref{fig:sensitivity-analysis} summarizes our sensitivity analysis across these four factors and shows that \metric{} achieves the most desirable balance: high identity sensitivity and low contextual sensitivity. Performance of other metrics varies across these challenges. DreamSim exhibits moderate identity sensitivity but remains similarly sensitive to background and lighting variation. In contrast, the Universal Embedding model and DINOv3 show greater invariance to viewpoint and lighting but are more sensitive to background changes. CLIP, OpenCLIP, and LPIPS show the weakest identity sensitivity, measuring semantic or image-level similarity rather than identity similarity. Examining the similarity scores in the bottom row of Figure \ref{fig:sensitivity-analysis} offers a complementary perspective: compared to other metrics, ID-Sim exhibits the largest similarity drop in response to identity changes while maintaining invariance to other factors, supporting its stronger identity sensitivity. ID-Sim is slightly less robust to lighting variation than DINOv3, reflecting a tradeoff to preserve fine-grained color cues for identity.

\vspace{-3pt}
\section{Limitations, Future Work, and Conclusions}
\label{sec:limitations}

\noindent \textbf{Limitations.}
Our instance definition relies on consistent visual identity and therefore does not fully capture broader notions of identity that may require user-specified invariances (e.g., aging, accessories, or stylistic changes). Also, ID-Sim is a global prompt-free metric and does not resolve the identity to target in multi-entity scenes; doing so requires external conditioning, either using spatial cues (e.g., masks) or text prompts, as explored by concurrent work Omni-Attribute~\cite{chen2025omniattributeopenvocabularyattributeencoder}. We show in the Supplemental~\ref{supp:dense_results} that our localized patch embeddings provide a natural foundation for more flexible, spatially-conditioned identity specification.

\vspace{3pt}
\noindent \textbf{Future work.}
Recent work personalization works \cite{sohn2023styledrop, wu2025less} has used synthetic data to bootstrap training, improving generalization and reducing overfitting. However, automating this has been difficult and error-prone, lacking the general, selectively sensitive identity embeddings that our work (ID-Sim) introduces. We believe leveraging ID-Sim for this task is a promising direction. In addition, conditioning signals can be incorporated for selective identity specification.

\vspace{3pt}
\noindent \textbf{Conclusions.}
Our results demonstrate that by combining a carefully curated dataset (\Cref{sec:dataset}) and training formulation (\Cref{sec:trainingsub}), it is possible to train a general purpose identity-focused similarity metric with state of the art performance across a wide variety of tasks, all at a fraction of the inference costs, training costs, and data requirements of MLLM foundation models. \metric~produces both global and local embeddings that can be easily plugged into any application that requires identity sensitivity and robustness to contextual changes (e.g., pose, background, lighting).

\paragraph{Acknowledgements} This work was supported by an NSERC PGS-D, a Schmidt Science AI2050 Early Career Fellowship, NSF CAREER Award No. 2441060, the NSF and NSERC AI and Biodiversity Change Global Center (NSF Award No. 2330423
and NSERC Award No. 585136), the MIT Generative AI Consortium, and the Department of the Air Force Artificial Intelligence Accelerator and was accomplished under Cooperative Agreement Number FA8750-19-2-1000. The views and conclusions contained in this document are those of the authors and should not be interpreted as representing the official policies, either expressed or implied, of the Department of the Air Force or the U.S. Government. The U.S. Government is authorized to reproduce and distribute reprints for Government purposes notwithstanding any copyright notation herein.

\clearpage

{
    \small
    \bibliographystyle{ieeenat_fullname}
    \bibliography{main}

@String(ECCV= {Eur. Conf. Comput. Vis.})

@String(CVPRW= {IEEE Conf. Comput. Vis. Pattern Recog. Worksh.})

@String(ECCV  = {ECCV})

@String(CVPRW= {CVPRW})

@article{dicarlo2007untangling,
  title={Untangling invariant object recognition},
  author={DiCarlo, James J and Cox, David D},
  journal={Trends in cognitive sciences},
  volume={11},
  number={8},
  pages={333--341},
  year={2007},
  publisher={Elsevier}
}

@inproceedings{mvimagenet,
  title={Mvimgnet: A large-scale dataset of multi-view images},
  author={Yu, Xianggang and Xu, Mutian and Zhang, Yidan and Liu, Haolin and Ye, Chongjie and Wu, Yushuang and Yan, Zizheng and Zhu, Chenming and Xiong, Zhangyang and Liang, Tianyou and others},
  booktitle={Proceedings of the IEEE/CVF conference on computer vision and pattern recognition},
  pages={9150--9161},
  year={2023}
}

@inproceedings{ge2019deepfashion2,
  title={Deepfashion2: A versatile benchmark for detection, pose estimation, segmentation and re-identification of clothing images},
  author={Ge, Yuying and Zhang, Ruimao and Wang, Xiaogang and Tang, Xiaoou and Luo, Ping},
  booktitle={Proceedings of the IEEE/CVF conference on computer vision and pattern recognition},
  pages={5337--5345},
  year={2019}
}

@article{trein2025siamese,
  title={Siamese Networks for Cat Re-Identification: Exploring Neural Models for Cat Instance Recognition},
  author={Trein, Tobias and Garcia, Luan Fonseca},
  journal={arXiv preprint arXiv:2501.02112},
  year={2025}
}

@inproceedings{schroff2015facenet,
  title={Facenet: A unified embedding for face recognition and clustering},
  author={Schroff, Florian and Kalenichenko, Dmitry and Philbin, James},
  booktitle={Proceedings of the IEEE conference on computer vision and pattern recognition},
  pages={815--823},
  year={2015}
}

@inproceedings{deng2019arcface,
  title={Arcface: Additive angular margin loss for deep face recognition},
  author={Deng, Jiankang and Guo, Jia and Xue, Niannan and Zafeiriou, Stefanos},
  booktitle={Proceedings of the IEEE/CVF conference on computer vision and pattern recognition},
  pages={4690--4699},
  year={2019}
}

@inproceedings{sun2018beyond,
  title={Beyond part models: Person retrieval with refined part pooling (and a strong convolutional baseline)},
  author={Sun, Yifan and Zheng, Liang and Yang, Yi and Tian, Qi and Wang, Shengjin},
  booktitle={Proceedings of the European conference on computer vision (ECCV)},
  pages={480--496},
  year={2018}
}

@inproceedings{chen2020simple,
  title={A simple framework for contrastive learning of visual representations},
  author={Chen, Ting and Kornblith, Simon and Norouzi, Mohammad and Hinton, Geoffrey},
  booktitle={International conference on machine learning},
  pages={1597--1607},
  year={2020},
  organization={PmLR}
}

@article{oquab2023dinov2,
  title={Dinov2: Learning robust visual features without supervision},
  author={Oquab, Maxime and Darcet, Timoth{\'e}e and Moutakanni, Th{\'e}o and Vo, Huy and Szafraniec, Marc and Khalidov, Vasil and Fernandez, Pierre and Haziza, Daniel and Massa, Francisco and El-Nouby, Alaaeldin and others},
  journal={arXiv preprint arXiv:2304.07193},
  year={2023}
}

@inproceedings{cole2022does,
  title={When does contrastive visual representation learning work?},
  author={Cole, Elijah and Yang, Xuan and Wilber, Kimberly and Mac Aodha, Oisin and Belongie, Serge},
  booktitle={Proceedings of the IEEE/CVF conference on computer vision and pattern recognition},
  pages={14755--14764},
  year={2022}
}

@article{grill2020bootstrap,
  title={Bootstrap your own latent-a new approach to self-supervised learning},
  author={Grill, Jean-Bastien and Strub, Florian and Altch{\'e}, Florent and Tallec, Corentin and Richemond, Pierre and Buchatskaya, Elena and Doersch, Carl and Avila Pires, Bernardo and Guo, Zhaohan and Gheshlaghi Azar, Mohammad and others},
  journal={Advances in neural information processing systems},
  volume={33},
  pages={21271--21284},
  year={2020}
}

@inproceedings{he2020momentum,
  title={Momentum contrast for unsupervised visual representation learning},
  author={He, Kaiming and Fan, Haoqi and Wu, Yuxin and Xie, Saining and Girshick, Ross},
  booktitle={Proceedings of the IEEE/CVF conference on computer vision and pattern recognition},
  pages={9729--9738},
  year={2020}
}

@article{zheng2016person,
  title={Person re-identification: Past, present and future},
  author={Zheng, Liang and Yang, Yi and Hauptmann, Alexander G},
  journal={arXiv preprint arXiv:1610.02984},
  year={2016}
}

@inproceedings{hore2010image,
  title={Image quality metrics: PSNR vs. SSIM},
  author={Hore, Alain and Ziou, Djemel},
  booktitle={2010 20th international conference on pattern recognition},
  pages={2366--2369},
  year={2010},
  organization={IEEE}
}

@article{gal2022image,
  title={An image is worth one word: Personalizing text-to-image generation using textual inversion},
  author={Gal, Rinon and Alaluf, Yuval and Atzmon, Yuval and Patashnik, Or and Bermano, Amit H and Chechik, Gal and Cohen-Or, Daniel},
  journal={arXiv preprint arXiv:2208.01618},
  year={2022}
}

@inproceedings{kumari2023multi,
  title={Multi-concept customization of text-to-image diffusion},
  author={Kumari, Nupur and Zhang, Bingliang and Zhang, Richard and Shechtman, Eli and Zhu, Jun-Yan},
  booktitle={Proceedings of the IEEE/CVF conference on computer vision and pattern recognition},
  pages={1931--1941},
  year={2023}
}

@inproceedings{ruiz2023dreambooth,
  title={Dreambooth: Fine tuning text-to-image diffusion models for subject-driven generation},
  author={Ruiz, Nataniel and Li, Yuanzhen and Jampani, Varun and Pritch, Yael and Rubinstein, Michael and Aberman, Kfir},
  booktitle={Proceedings of the IEEE/CVF conference on computer vision and pattern recognition},
  pages={22500--22510},
  year={2023}
}

@article{schneider2019past,
  title={Past, present and future approaches using computer vision for animal re-identification from camera trap data},
  author={Schneider, Stefan and Taylor, Graham W and Linquist, Stefan and Kremer, Stefan C},
  journal={Methods in Ecology and Evolution},
  volume={10},
  number={4},
  pages={461--470},
  year={2019},
  publisher={Wiley Online Library}
}

@article{chen2022deep,
  title={Deep learning for instance retrieval: A survey},
  author={Chen, Wei and Liu, Yu and Wang, Weiping and Bakker, Erwin M and Georgiou, Theodoros and Fieguth, Paul and Liu, Li and Lew, Michael S},
  journal={IEEE Transactions on Pattern Analysis and Machine Intelligence},
  volume={45},
  number={6},
  pages={7270--7292},
  year={2022},
  publisher={IEEE}
}

@article{zheng2017sift,
  title={SIFT meets CNN: A decade survey of instance retrieval},
  author={Zheng, Liang and Yang, Yi and Tian, Qi},
  journal={IEEE transactions on pattern analysis and machine intelligence},
  volume={40},
  number={5},
  pages={1224--1244},
  year={2017},
  publisher={IEEE}
}

@inproceedings{adam2025wildlifereid,
  title={WildlifeReID-10k: Wildlife re-identification dataset with 10k individual animals},
  author={Adam, Luk{\'a}{\v{s}} and {\v{C}}erm{\'a}k, Vojt{\v{e}}ch and Papafitsoros, Kostas and Picek, Lukas},
  booktitle={2025 IEEE/CVF Conference on Computer Vision and Pattern Recognition Workshops (CVPRW)},
  pages={2090--2100},
  year={2025},
  organization={IEEE}
}

@article{ye2021deep,
  title={Deep learning for person re-identification: A survey and outlook},
  author={Ye, Mang and Shen, Jianbing and Lin, Gaojie and Xiang, Tao and Shao, Ling and Hoi, Steven CH},
  journal={IEEE transactions on pattern analysis and machine intelligence},
  volume={44},
  number={6},
  pages={2872--2893},
  year={2021},
  publisher={IEEE}
}

@article{kim2024pose,
  title={Pose-dIVE: Pose-Diversified Augmentation with Diffusion Model for Person Re-Identification},
  author={Kim, In{\`e}s Hyeonsu and Lee, JoungBin and Jin, Woojeong and Son, Soowon and Cho, Kyusun and Seo, Junyoung and Kwak, Min-Seop and Cho, Seokju and Baek, JeongYeol and Lee, Byeongwon and others},
  journal={arXiv preprint arXiv:2406.16042},
  year={2024}
}

@article{schneider2022similarity,
  title={Similarity learning networks for animal individual re-identification: an ecological perspective},
  author={Schneider, Stefan and Taylor, Graham W and Kremer, Stefan C},
  journal={Mammalian Biology},
  volume={102},
  number={3},
  pages={899--914},
  year={2022},
  publisher={Springer}
}

@article{otarashvili2024multispecies,
  title={Multispecies animal re-id using a large community-curated dataset},
  author={Otarashvili, Lasha and Subramanian, Tamilselvan and Holmberg, Jason and Levenson, JJ and Stewart, Charles V},
  journal={arXiv preprint arXiv:2412.05602},
  year={2024}
}

@inproceedings{he2019foreground,
  title={Foreground-aware pyramid reconstruction for alignment-free occluded person re-identification},
  author={He, Lingxiao and Wang, Yinggang and Liu, Wu and Zhao, He and Sun, Zhenan and Feng, Jiashi},
  booktitle={Proceedings of the IEEE/CVF international conference on computer vision},
  pages={8450--8459},
  year={2019}
}

@inproceedings{song2019generalizable,
  title={Generalizable person re-identification by domain-invariant mapping network},
  author={Song, Jifei and Yang, Yongxin and Song, Yi-Zhe and Xiang, Tao and Hospedales, Timothy M},
  booktitle={Proceedings of the IEEE/CVF conference on Computer Vision and Pattern Recognition},
  pages={719--728},
  year={2019}
}

@inproceedings{wang2018cosface,
  title={Cosface: Large margin cosine loss for deep face recognition},
  author={Wang, Hao and Wang, Yitong and Zhou, Zheng and Ji, Xing and Gong, Dihong and Zhou, Jingchao and Li, Zhifeng and Liu, Wei},
  booktitle={Proceedings of the IEEE conference on computer vision and pattern recognition},
  pages={5265--5274},
  year={2018}
}

@inproceedings{liu2017sphereface,
  title={Sphereface: Deep hypersphere embedding for face recognition},
  author={Liu, Weiyang and Wen, Yandong and Yu, Zhiding and Li, Ming and Raj, Bhiksha and Song, Le},
  booktitle={Proceedings of the IEEE conference on computer vision and pattern recognition},
  pages={212--220},
  year={2017}
}

@article{wu2023forb,
  title={FORB: a flat object retrieval benchmark for universal image embedding},
  author={Wu, Pengxiang and Wang, Siman and Dela Rosa, Kevin and Hu, Derek},
  journal={Advances in Neural Information Processing Systems},
  volume={36},
  pages={25448--25460},
  year={2023}
}

@inproceedings{kordopatis2025ilias,
  title={Ilias: Instance-level image retrieval at scale},
  author={Kordopatis-Zilos, Giorgos and Stojni{\'c}, Vladan and Manko, Anna and Suma, Pavel and Ypsilantis, Nikolaos-Antonios and Efthymiadis, Nikos and Laskar, Zakaria and Matas, Jiri and Chum, Ondrej and Tolias, Giorgos},
  booktitle={Proceedings of the Computer Vision and Pattern Recognition Conference},
  pages={14777--14787},
  year={2025}
}

@inproceedings{ypsilantis2021met,
  title={The met dataset: Instance-level recognition for artworks},
  author={Ypsilantis, Nikolaos-Antonios and Garcia, Noa and Han, Guangxing and Ibrahimi, Sarah and Van Noord, Nanne and Tolias, Giorgos},
  booktitle={Thirty-fifth conference on neural information processing systems datasets and benchmarks track (Round 2)},
  year={2021}
}

@inproceedings{weyand2020google,
  title={Google landmarks dataset v2-a large-scale benchmark for instance-level recognition and retrieval},
  author={Weyand, Tobias and Araujo, Andre and Cao, Bingyi and Sim, Jack},
  booktitle={Proceedings of the IEEE/CVF conference on computer vision and pattern recognition},
  pages={2575--2584},
  year={2020}
}

@article{biederman1987recognition,
  title={Recognition-by-components: a theory of human image understanding.},
  author={Biederman, Irving},
  journal={Psychological review},
  volume={94},
  number={2},
  pages={115},
  year={1987},
  publisher={American Psychological Association}
}

@article{logothetis1995psychophysical,
  title={Psychophysical and physiological evidence for viewer-centered object representations in the primate},
  author={Logothetis, Nikos K and Pauls, Jon},
  journal={Cerebral cortex},
  volume={5},
  number={3},
  pages={270--288},
  year={1995},
  publisher={Oxford University Press}
}

@article{palmeri2000role,
  title={The role of background knowledge in speeded perceptual categorization},
  author={Palmeri, Thomas J and Blalock, Celina},
  journal={Cognition},
  volume={77},
  number={2},
  pages={B45--B57},
  year={2000},
  publisher={Elsevier}
}

@article{rosch1975cognitive,
  title={Cognitive representations of semantic categories.},
  author={Rosch, Eleanor},
  journal={Journal of experimental psychology: General},
  volume={104},
  number={3},
  pages={192},
  year={1975},
  publisher={American Psychological Association}
}

@article{palmeri2004visual,
  title={Visual object understanding},
  author={Palmeri, Thomas J and Gauthier, Isabel},
  journal={Nature Reviews Neuroscience},
  volume={5},
  number={4},
  pages={291--303},
  year={2004},
  publisher={Nature Publishing Group UK London}
}

@misc{song2015deepmetriclearninglifted,
      title={Deep Metric Learning via Lifted Structured Feature Embedding}, 
      author={Hyun Oh Song and Yu Xiang and Stefanie Jegelka and Silvio Savarese},
      year={2015},
      eprint={1511.06452},
      archivePrefix={arXiv},
      primaryClass={cs.CV},
      url={https://arxiv.org/abs/1511.06452}, 
}

@article{ding2020image,
  title={Image quality assessment: Unifying structure and texture similarity},
  author={Ding, Keyan and Ma, Kede and Wang, Shiqi and Simoncelli, Eero P},
  journal={IEEE transactions on pattern analysis and machine intelligence},
  volume={44},
  number={5},
  pages={2567--2581},
  year={2020},
  publisher={IEEE}
}

@inproceedings{zhang2018unreasonable,
  title={The unreasonable effectiveness of deep features as a perceptual metric},
  author={Zhang, Richard and Isola, Phillip and Efros, Alexei A and Shechtman, Eli and Wang, Oliver},
  booktitle={Proceedings of the IEEE conference on computer vision and pattern recognition},
  pages={586--595},
  year={2018}
}

@article{fu2023dreamsim,
  title={Dreamsim: Learning new dimensions of human visual similarity using synthetic data},
  author={Fu, Stephanie and Tamir, Netanel and Sundaram, Shobhita and Chai, Lucy and Zhang, Richard and Dekel, Tali and Isola, Phillip},
  journal={arXiv preprint arXiv:2306.09344},
  year={2023}
}

@inproceedings{caron2021emerging,
  title={Emerging properties in self-supervised vision transformers},
  author={Caron, Mathilde and Touvron, Hugo and Misra, Ishan and J{\'e}gou, Herv{\'e} and Mairal, Julien and Bojanowski, Piotr and Joulin, Armand},
  booktitle={Proceedings of the IEEE/CVF international conference on computer vision},
  pages={9650--9660},
  year={2021}
}

@article{peng2024dreambench++,
  title={Dreambench++: A human-aligned benchmark for personalized image generation},
  author={Peng, Yuang and Cui, Yuxin and Tang, Haomiao and Qi, Zekun and Dong, Runpei and Bai, Jing and Han, Chunrui and Ge, Zheng and Zhang, Xiangyu and Xia, Shu-Tao},
  journal={arXiv preprint arXiv:2406.16855},
  year={2024}
}

@inproceedings{radford2021learning,
  title={Learning transferable visual models from natural language supervision},
  author={Radford, Alec and Kim, Jong Wook and Hallacy, Chris and Ramesh, Aditya and Goh, Gabriel and Agarwal, Sandhini and Sastry, Girish and Askell, Amanda and Mishkin, Pamela and Clark, Jack and others},
  booktitle={International conference on machine learning},
  pages={8748--8763},
  year={2021},
  organization={PmLR}
}

@inproceedings{GPR1200,
    author = {Schall, Konstantin and Barthel, Kai Uwe and Hezel, Nico and Jung, Klaus},
    title = {GPR1200: A Benchmark for General-Purpose Content-Based Image Retrieval},
    year = {2022},
    isbn = {978-3-030-98357-4},
    publisher = {Springer-Verlag},
    address = {Berlin, Heidelberg},
    url = {https://doi.org/10.1007/978-3-030-98358-1_17},
    doi = {10.1007/978-3-030-98358-1_17},
    booktitle = {MultiMedia Modeling: 28th International Conference, MMM 2022, Phu Quoc, Vietnam, June 6–10, 2022, Proceedings, Part I},
    pages = {205–216},
    numpages = {12},
    location = {Phu Quoc, Vietnam}
}

@article{hurst2024gpt,
  title={Gpt-4o system card},
  author={Hurst, Aaron and Lerer, Adam and Goucher, Adam P and Perelman, Adam and Ramesh, Aditya and Clark, Aidan and Ostrow, AJ and Welihinda, Akila and Hayes, Alan and Radford, Alec and others},
  journal={arXiv preprint arXiv:2410.21276},
  year={2024}
}

@article{shaked2024minimizing,
  title={Minimizing Embedding Distortion for Robust Out-of-Distribution Performance},
  author={Shaked, Tom and Goldman, Yuval and Shayer, Oran},
  journal={arXiv preprint arXiv:2409.07582},
  year={2024}
}

@inproceedings{ypsilantis2023towards,
  title={Towards universal image embeddings: A large-scale dataset and challenge for generic image representations},
  author={Ypsilantis, Nikolaos-Antonios and Chen, Kaifeng and Cao, Bingyi and Lipovsk{\`y}, M{\'a}rio and Dogan-Sch{\"o}nberger, Pelin and Makosa, Grzegorz and Bluntschli, Boris and Seyedhosseini, Mojtaba and Chum, Ond{\v{r}}ej and Araujo, Andr{\'e}},
  booktitle={Proceedings of the ieee/cvf international conference on computer vision},
  pages={11290--11301},
  year={2023}
}

@article{shi2024judging,
  title={Judging the judges: A systematic study of position bias in llm-as-a-judge},
  author={Shi, Lin and Ma, Chiyu and Liang, Wenhua and Diao, Xingjian and Ma, Weicheng and Vosoughi, Soroush},
  journal={arXiv preprint arXiv:2406.07791},
  year={2024}
}

@article{sampat2009complex,
  title={Complex wavelet structural similarity: A new image similarity index},
  author={Sampat, Mehul P and Wang, Zhou and Gupta, Shalini and Bovik, Alan Conrad and Markey, Mia K},
  journal={IEEE transactions on image processing},
  volume={18},
  number={11},
  pages={2385--2401},
  year={2009},
  publisher={IEEE}
}

@article{zhang2011fsim,
  title={FSIM: A feature similarity index for image quality assessment},
  author={Zhang, Lin and Zhang, Lei and Mou, Xuanqin and Zhang, David},
  journal={IEEE transactions on Image Processing},
  volume={20},
  number={8},
  pages={2378--2386},
  year={2011},
  publisher={IEEE}
}

@article{wang2004image,
  title={Image quality assessment: from error visibility to structural similarity},
  author={Wang, Zhou and Bovik, Alan C and Sheikh, Hamid R and Simoncelli, Eero P},
  journal={IEEE transactions on image processing},
  volume={13},
  number={4},
  pages={600--612},
  year={2004},
  publisher={IEEE}
}

@inproceedings{prashnani2018pieapp,
  title={Pieapp: Perceptual image-error assessment through pairwise preference},
  author={Prashnani, Ekta and Cai, Hong and Mostofi, Yasamin and Sen, Pradeep},
  booktitle={Proceedings of the IEEE Conference on Computer Vision and Pattern Recognition},
  pages={1808--1817},
  year={2018}
}

@article{krizhevsky2012imagenet,
  title={Imagenet classification with deep convolutional neural networks},
  author={Krizhevsky, Alex and Sutskever, Ilya and Hinton, Geoffrey E},
  journal={Advances in neural information processing systems},
  volume={25},
  year={2012}
}

@article{simonyan2014very,
  title={Very deep convolutional networks for large-scale image recognition},
  author={Simonyan, Karen and Zisserman, Andrew},
  journal={arXiv preprint arXiv:1409.1556},
  year={2014}
}

@inproceedings{wang2003multiscale,
  title={Multiscale structural similarity for image quality assessment},
  author={Wang, Zhou and Simoncelli, Eero P and Bovik, Alan C},
  booktitle={The thrity-seventh asilomar conference on signals, systems \& computers, 2003},
  volume={2},
  pages={1398--1402},
  year={2003},
  organization={Ieee}
}

@inproceedings{tamir2025makes,
  title={What Makes for a Good Stereoscopic Image?},
  author={Tamir, Netanel and Amir, Shir and Itzhaky, Ranel and Atia, Noam and Sundaram, Shobhita and Fu, Stephanie and Sokolovsky, Ron and Isola, Phillip and Dekel, Tali and Zhang, Richard and others},
  booktitle={Proceedings of the Computer Vision and Pattern Recognition Conference},
  pages={261--272},
  year={2025}
}

@article{manocha2020differentiable,
  title={A differentiable perceptual audio metric learned from just noticeable differences},
  author={Manocha, Pranay and Finkelstein, Adam and Zhang, Richard and Bryan, Nicholas J and Mysore, Gautham J and Jin, Zeyu},
  journal={arXiv preprint arXiv:2001.04460},
  year={2020}
}

@inproceedings{tian2020contrastive,
  title={Contrastive multiview coding},
  author={Tian, Yonglong and Krishnan, Dilip and Isola, Phillip},
  booktitle={European conference on computer vision},
  pages={776--794},
  year={2020},
  organization={Springer}
}

@article{oord2018representation,
  title={Representation learning with contrastive predictive coding},
  author={Oord, Aaron van den and Li, Yazhe and Vinyals, Oriol},
  journal={arXiv preprint arXiv:1807.03748},
  year={2018}
}

@inproceedings{wu2018unsupervised,
  title={Unsupervised feature learning via non-parametric instance discrimination},
  author={Wu, Zhirong and Xiong, Yuanjun and Yu, Stella X and Lin, Dahua},
  booktitle={Proceedings of the IEEE conference on computer vision and pattern recognition},
  pages={3733--3742},
  year={2018}
}

@article{hjelm2018learning,
  title={Learning deep representations by mutual information estimation and maximization},
  author={Hjelm, R Devon and Fedorov, Alex and Lavoie-Marchildon, Samuel and Grewal, Karan and Bachman, Phil and Trischler, Adam and Bengio, Yoshua},
  journal={arXiv preprint arXiv:1808.06670},
  year={2018}
}

@InProceedings{Cermak_2024_WACV,
    author    = {\v{C}erm\'ak, Vojt\v{e}ch and Picek, Luk\'a\v{s} and Adam, Luk\'a\v{s} and Papafitsoros, Kostas},
    title     = {{WildlifeDatasets: An Open-Source Toolkit for Animal Re-Identification}},
    booktitle = {Proceedings of the IEEE/CVF Winter Conference on Applications of Computer Vision (WACV)},
    month     = {January},
    year      = {2024},
    pages     = {5953-5963}
}

@inproceedings{liu24uco3d,
                Author = {Liu, Xingchen and Tayal, Piyush and Wang, Jianyuan
                          and Zarzar, Jesus and Monnier, Tom and Tertikas, Konstantinos
                          and Duan, Jiali and Toisoul, Antoine and Zhang, Jason Y.
                          and Neverova, Natalia and Vedaldi, Andrea
                          and Shapovalov, Roman and Novotny, David},
                Booktitle = {arXiv},
                Title = {UnCommon Objects in 3D},
                Year = {2024},
            }

@misc{fan2020lasothighqualitylargescalesingle,
      title={LaSOT: A High-quality Large-scale Single Object Tracking Benchmark}, 
      author={Heng Fan and Hexin Bai and Liting Lin and Fan Yang and Peng Chu and Ge Deng and Sijia Yu and Harshit and Mingzhen Huang and Juehuan Liu and Yong Xu and Chunyuan Liao and Lin Yuan and Haibin Ling},
      year={2020},
      eprint={2009.03465},
      archivePrefix={arXiv},
      primaryClass={cs.CV},
      url={https://arxiv.org/abs/2009.03465}, 
}

@misc{yang2019videoinstancesegmentation,
      title={Video Instance Segmentation}, 
      author={Linjie Yang and Yuchen Fan and Ning Xu},
      year={2019},
      eprint={1905.04804},
      archivePrefix={arXiv},
      primaryClass={cs.CV},
      url={https://arxiv.org/abs/1905.04804}, 
}

@misc{wu2025qwenimagetechnicalreport,
      title={Qwen-Image Technical Report}, 
      author={Chenfei Wu and Jiahao Li and Jingren Zhou and Junyang Lin and Kaiyuan Gao and Kun Yan and Sheng-ming Yin and Shuai Bai and Xiao Xu and Yilei Chen and Yuxiang Chen and Zecheng Tang and Zekai Zhang and Zhengyi Wang and An Yang and Bowen Yu and Chen Cheng and Dayiheng Liu and Deqing Li and Hang Zhang and Hao Meng and Hu Wei and Jingyuan Ni and Kai Chen and Kuan Cao and Liang Peng and Lin Qu and Minggang Wu and Peng Wang and Shuting Yu and Tingkun Wen and Wensen Feng and Xiaoxiao Xu and Yi Wang and Yichang Zhang and Yongqiang Zhu and Yujia Wu and Yuxuan Cai and Zenan Liu},
      year={2025},
      eprint={2508.02324},
      archivePrefix={arXiv},
      primaryClass={cs.CV},
      url={https://arxiv.org/abs/2508.02324}, 
}

@misc{eldesokey2025mindtheglitchvisualcorrespondencedetecting,
      title={Mind-the-Glitch: Visual Correspondence for Detecting Inconsistencies in Subject-Driven Generation}, 
      author={Abdelrahman Eldesokey and Aleksandar Cvejic and Bernard Ghanem and Peter Wonka},
      year={2025},
      eprint={2509.21989},
      archivePrefix={arXiv},
      primaryClass={cs.CV},
      url={https://arxiv.org/abs/2509.21989}, 
}

@misc{dosovitskiy2021imageworth16x16words,
      title={An Image is Worth 16x16 Words: Transformers for Image Recognition at Scale}, 
      author={Alexey Dosovitskiy and Lucas Beyer and Alexander Kolesnikov and Dirk Weissenborn and Xiaohua Zhai and Thomas Unterthiner and Mostafa Dehghani and Matthias Minderer and Georg Heigold and Sylvain Gelly and Jakob Uszkoreit and Neil Houlsby},
      year={2021},
      eprint={2010.11929},
      archivePrefix={arXiv},
      primaryClass={cs.CV},
      url={https://arxiv.org/abs/2010.11929}, 
}

@misc{oquab2024dinov2learningrobustvisual,
      title={DINOv2: Learning Robust Visual Features without Supervision}, 
      author={Maxime Oquab and Timothée Darcet and Théo Moutakanni and Huy Vo and Marc Szafraniec and Vasil Khalidov and Pierre Fernandez and Daniel Haziza and Francisco Massa and Alaaeldin El-Nouby and Mahmoud Assran and Nicolas Ballas and Wojciech Galuba and Russell Howes and Po-Yao Huang and Shang-Wen Li and Ishan Misra and Michael Rabbat and Vasu Sharma and Gabriel Synnaeve and Hu Xu and Hervé Jegou and Julien Mairal and Patrick Labatut and Armand Joulin and Piotr Bojanowski},
      year={2024},
      eprint={2304.07193},
      archivePrefix={arXiv},
      primaryClass={cs.CV},
      url={https://arxiv.org/abs/2304.07193}, 
}

@misc{radford2021learningtransferablevisualmodels,
      title={Learning Transferable Visual Models From Natural Language Supervision}, 
      author={Alec Radford and Jong Wook Kim and Chris Hallacy and Aditya Ramesh and Gabriel Goh and Sandhini Agarwal and Girish Sastry and Amanda Askell and Pamela Mishkin and Jack Clark and Gretchen Krueger and Ilya Sutskever},
      year={2021},
      eprint={2103.00020},
      archivePrefix={arXiv},
      primaryClass={cs.CV},
      url={https://arxiv.org/abs/2103.00020}, 
}

@software{ilharco_gabriel_2021_5143773,
  author       = {Ilharco, Gabriel and
                  Wortsman, Mitchell and
                  Wightman, Ross and
                  Gordon, Cade and
                  Carlini, Nicholas and
                  Taori, Rohan and
                  Dave, Achal and
                  Shankar, Vaishaal and
                  Namkoong, Hongseok and
                  Miller, John and
                  Hajishirzi, Hannaneh and
                  Farhadi, Ali and
                  Schmidt, Ludwig},
  title        = {OpenCLIP},
  month        = jul,
  year         = 2021,
  note         = {If you use this software, please cite it as below.},
  publisher    = {Zenodo},
  version      = {0.1},
  doi          = {10.5281/zenodo.5143773},
  url          = {https://doi.org/10.5281/zenodo.5143773}
}

@misc{zhang2023personalizesegmentmodelshot,
      title={Personalize Segment Anything Model with One Shot}, 
      author={Renrui Zhang and Zhengkai Jiang and Ziyu Guo and Shilin Yan and Junting Pan and Xianzheng Ma and Hao Dong and Peng Gao and Hongsheng Li},
      year={2023},
      eprint={2305.03048},
      archivePrefix={arXiv},
      primaryClass={cs.CV},
      url={https://arxiv.org/abs/2305.03048}, 
}

@misc{khosla2021supervisedcontrastivelearning,
      title={Supervised Contrastive Learning}, 
      author={Prannay Khosla and Piotr Teterwak and Chen Wang and Aaron Sarna and Yonglong Tian and Phillip Isola and Aaron Maschinot and Ce Liu and Dilip Krishnan},
      year={2021},
      eprint={2004.11362},
      archivePrefix={arXiv},
      primaryClass={cs.LG},
      url={https://arxiv.org/abs/2004.11362}, 
}

@misc{oord2019representationlearningcontrastivepredictive,
      title={Representation Learning with Contrastive Predictive Coding}, 
      author={Aaron van den Oord and Yazhe Li and Oriol Vinyals},
      year={2019},
      eprint={1807.03748},
      archivePrefix={arXiv},
      primaryClass={cs.LG},
      url={https://arxiv.org/abs/1807.03748}, 
}

@inproceedings{feydy2019interpolating,
    title={Interpolating between Optimal Transport and MMD using Sinkhorn Divergences},
    author={Feydy, Jean and S{\'e}journ{\'e}, Thibault and Vialard, Fran{\c{c}}ois-Xavier and Amari, Shun-ichi and Trouve, Alain and Peyr{\'e}, Gabriel},
    booktitle={The 22nd International Conference on Artificial Intelligence and Statistics},
    pages={2681--2690},
    year={2019}
}

@misc{hu2021loralowrankadaptationlarge,
      title={LoRA: Low-Rank Adaptation of Large Language Models}, 
      author={Edward J. Hu and Yelong Shen and Phillip Wallis and Zeyuan Allen-Zhu and Yuanzhi Li and Shean Wang and Lu Wang and Weizhu Chen},
      year={2021},
      eprint={2106.09685},
      archivePrefix={arXiv},
      primaryClass={cs.CL},
      url={https://arxiv.org/abs/2106.09685}, 
}

@misc{shao2023guie,
  author       = {Shao, Shihao and Cui, Qinghua},
  title        = {1st Solution in Google Universal Image Embedding},
  howpublished = {\url{https://www.kaggle.com/datasets/louieshao/guieweights0732}},
  year         = {2023}
}

@misc{siméoni2025dinov3,
      title={DINOv3}, 
      author={Oriane Siméoni and Huy V. Vo and Maximilian Seitzer and Federico Baldassarre and Maxime Oquab and Cijo Jose and Vasil Khalidov and Marc Szafraniec and Seungeun Yi and Michaël Ramamonjisoa and Francisco Massa and Daniel Haziza and Luca Wehrstedt and Jianyuan Wang and Timothée Darcet and Théo Moutakanni and Leonel Sentana and Claire Roberts and Andrea Vedaldi and Jamie Tolan and John Brandt and Camille Couprie and Julien Mairal and Hervé Jégou and Patrick Labatut and Piotr Bojanowski},
      year={2025},
      eprint={2508.10104},
      archivePrefix={arXiv},
      primaryClass={cs.CV},
      url={https://arxiv.org/abs/2508.10104}, 
}

@software{openai2023gpt4v,
  author       = {OpenAI},
  title        = {GPT-4V (Vision): Multimodal GPT-4 with image and text input},
  year         = {2023},
  howpublished = {\url{https://openai.com/research/gpt-4v-system-card}},
  note         = {Accessed: 2025-11-13}
}

@inproceedings{peng2024dreambench,
  author={Yuang Peng and Yuxin Cui and Haomiao Tang and Zekun Qi and Runpei Dong and Jing Bai and Chunrui Han and Zheng Ge and Xiangyu Zhang and Shu-Tao Xia},
  title={DreamBench++: A Human-Aligned Benchmark for Personalized Image Generation},
  booktitle={The Thirteenth International Conference on Learning Representations},
  url={https://openreview.net/forum?id=4GSOESJrk6},
  year={2025},
}

@article{sohn2023styledrop,
  title={Styledrop: Text-to-image generation in any style},
  author={Sohn, Kihyuk and Ruiz, Nataniel and Lee, Kimin and Chin, Daniel Castro and Blok, Irina and Chang, Huiwen and Barber, Jarred and Jiang, Lu and Entis, Glenn and Li, Yuanzhen and others},
  journal={arXiv preprint arXiv:2306.00983},
  year={2023}
}

@article{wu2025less,
  title={Less-to-more generalization: Unlocking more controllability by in-context generation},
  author={Wu, Shaojin and Huang, Mengqi and Wu, Wenxu and Cheng, Yufeng and Ding, Fei and He, Qian},
  journal={arXiv preprint arXiv:2504.02160},
  year={2025}
}

@misc{wang2013theoreticalanalysisndcgtype,
      title={A Theoretical Analysis of NDCG Type Ranking Measures}, 
      author={Yining Wang and Liwei Wang and Yuanzhi Li and Di He and Tie-Yan Liu and Wei Chen},
      year={2013},
      eprint={1304.6480},
      archivePrefix={arXiv},
      primaryClass={cs.LG},
      url={https://arxiv.org/abs/1304.6480}, 
}

@article{
  tan2024omini,
  title={OminiControl: Minimal and Universal Control for Diffusion Transformer},
  author={Zhenxiong Tan and Songhua Liu and Xingyi Yang and Qiaochu Xue and Xinchao Wang},
  journal={arXiv preprint arXiv:2411.15098},
  year={2024}
}

@misc{sundaram2024personalizedrepresentationpersonalizedgeneration,
      title={Personalized Representation from Personalized Generation}, 
      author={Shobhita Sundaram and Julia Chae and Yonglong Tian and Sara Beery and Phillip Isola},
      year={2024},
      eprint={2412.16156},
      archivePrefix={arXiv},
      primaryClass={cs.CV},
      url={https://arxiv.org/abs/2412.16156}, 
}

@misc{wu2025instancelevelgenerationrepresentationlearning,
      title={Instance-Level Generation for Representation Learning}, 
      author={Yankun Wu and Zakaria Laskar and Giorgos Kordopatis-Zilos and Noa Garcia and Giorgos Tolias},
      year={2025},
      eprint={2510.09171},
      archivePrefix={arXiv},
      primaryClass={cs.CV},
      url={https://arxiv.org/abs/2510.09171}, 
}

@inproceedings{song2025diffsim,
  title={Diffsim: Taming diffusion models for evaluating visual similarity},
  author={Song, Yiren and Liu, Xiaokang and Shou, Mike Zheng},
  booktitle={Proceedings of the IEEE/CVF International Conference on Computer Vision},
  pages={16904--16915},
  year={2025}
}

@inproceedings{ham2024personalized,
  title={Personalized residuals for concept-driven text-to-image generation},
  author={Ham, Cusuh and Fisher, Matthew and Hays, James and Kolkin, Nicholas and Liu, Yuchen and Zhang, Richard and Hinz, Tobias},
  booktitle={Proceedings of the IEEE/CVF Conference on Computer Vision and Pattern Recognition},
  pages={8186--8195},
  year={2024}
}

@inproceedings{tan2025ominicontrol,
  title={Ominicontrol: Minimal and universal control for diffusion transformer},
  author={Tan, Zhenxiong and Liu, Songhua and Yang, Xingyi and Xue, Qiaochu and Wang, Xinchao},
  booktitle={Proceedings of the IEEE/CVF International Conference on Computer Vision},
  pages={14940--14950},
  year={2025}
}

@article{he2025conceptrol,
  title={Conceptrol: Concept Control of Zero-shot Personalized Image Generation},
  author={He, Qiyuan and Yao, Angela},
  journal={arXiv preprint arXiv:2503.06568},
  year={2025}
}

@article{ye2023ip,
  title={Ip-adapter: Text compatible image prompt adapter for text-to-image diffusion models},
  author={Ye, Hu and Zhang, Jun and Liu, Sibo and Han, Xiao and Yang, Wei},
  journal={arXiv preprint arXiv:2308.06721},
  year={2023}
}

@misc{sundaram2024doesperceptualalignmentbenefit,
      title={When Does Perceptual Alignment Benefit Vision Representations?}, 
      author={Shobhita Sundaram and Stephanie Fu and Lukas Muttenthaler and Netanel Y. Tamir and Lucy Chai and Simon Kornblith and Trevor Darrell and Phillip Isola},
      year={2024},
      eprint={2410.10817},
      archivePrefix={arXiv},
      primaryClass={cs.CV},
      url={https://arxiv.org/abs/2410.10817}, 
}

@misc{jiang2025personalizedvisionvisualincontext,
      title={Personalized Vision via Visual In-Context Learning}, 
      author={Yuxin Jiang and Yuchao Gu and Yiren Song and Ivor Tsang and Mike Zheng Shou},
      year={2025},
      eprint={2509.25172},
      archivePrefix={arXiv},
      primaryClass={cs.CV},
      url={https://arxiv.org/abs/2509.25172}, 
}

@misc{samuel2024whereswaldodiffusionfeatures,
      title={Where's Waldo: Diffusion Features for Personalized Segmentation and Retrieval}, 
      author={Dvir Samuel and Rami Ben-Ari and Matan Levy and Nir Darshan and Gal Chechik},
      year={2024},
      eprint={2405.18025},
      archivePrefix={arXiv},
      primaryClass={cs.CV},
      url={https://arxiv.org/abs/2405.18025}, 
}

@misc{flux2024,
    author={Black Forest Labs},
    title={FLUX},
    year={2024},
    howpublished={\url{https://github.com/black-forest-labs/flux}},
}

@misc{shinoda2024petfacelargescaledatasetbenchmark,
      title={PetFace: A Large-Scale Dataset and Benchmark for Animal Identification}, 
      author={Risa Shinoda and Kaede Shiohara},
      year={2024},
      eprint={2407.13555},
      archivePrefix={arXiv},
      primaryClass={cs.CV},
      url={https://arxiv.org/abs/2407.13555}, 
}

@article{Andrew_2021,
   title={Visual identification of individual Holstein-Friesian cattle via deep metric learning},
   volume={185},
   ISSN={0168-1699},
   url={http://dx.doi.org/10.1016/j.compag.2021.106133},
   DOI={10.1016/j.compag.2021.106133},
   journal={Computers and Electronics in Agriculture},
   publisher={Elsevier BV},
   author={Andrew, William and Gao, Jing and Mullan, Siobhan and Campbell, Neill and Dowsey, Andrew W. and Burghardt, Tilo},
   year={2021},
   month=jun, pages={106133} }

@inproceedings{kotar2023cute,
  title={Are These the Same Apple? Comparing Images Based on Object Intrinsics},
  author={Klemen Kotar and Stephen Tian and Hong-Xing Yu and Daniel L.K. Yamins and Jiajun Wu},
  journal={Neural Information Processing Systems Datasets and Benchmarks Track},
  year={2023}
}

@InProceedings{dogfacenet,
author="Mougeot, Guillaume and Li, Dewei and Jia, Shuai",
editor="Nayak, Abhaya C. and Sharma, Alok",
title="A Deep Learning Approach for Dog Face Verification and Recognition",
booktitle="PRICAI 2019: Trends in Artificial Intelligence",
year="2019",
publisher="Springer International Publishing",
address="Cham",
pages="418--430",
isbn="978-3-030-29894-4"
}

@article{Huang_2021,
   title={GOT-10k: A Large High-Diversity Benchmark for Generic Object Tracking in the Wild},
   volume={43},
   ISSN={1939-3539},
   url={http://dx.doi.org/10.1109/TPAMI.2019.2957464},
   DOI={10.1109/tpami.2019.2957464},
   number={5},
   journal={IEEE Transactions on Pattern Analysis and Machine Intelligence},
   publisher={Institute of Electrical and Electronics Engineers (IEEE)},
   author={Huang, Lianghua and Zhao, Xin and Huang, Kaiqi},
   year={2021},
   month=may, pages={1562–1577} }

@software{adobephotoshop,
  author = {{Adobe Inc.}},
  title = {Adobe Photoshop},
  url = {https://www.adobe.com/products/photoshop.html},
  version = {CC 2019},
  date = {2019-03-06},
}

@misc{rosebrock2015blur,
  author       = {Rosebrock, Adrian},
  title        = {Blur detection with OpenCV},
  year         = {2015},
  howpublished = {\url{https://pyimagesearch.com/2015/09/07/blur-detection-with-opencv/}},
  note         = {Accessed: 2021-07-12}
}

@misc{chen2025omniattributeopenvocabularyattributeencoder,
      title={Omni-Attribute: Open-vocabulary Attribute Encoder for Visual Concept Personalization}, 
      author={T.S. Chen et al.},
      year={2025},
      eprint={2512.10955},
      archivePrefix={arXiv},
      primaryClass={cs.CV},
      url={https://arxiv.org/abs/2512.10955}, 
}

@misc{wang2021densecontrastivelearningselfsupervised,
      title={Dense Contrastive Learning for Self-Supervised Visual Pre-Training}, 
      author={X. Wang et al.},
      year={2021},
      eprint={2011.09157},
      archivePrefix={arXiv},
      primaryClass={cs.CV},
      url={https://arxiv.org/abs/2011.09157}, 
}
}

\clearpage
\setcounter{page}{1}
\maketitlesupplementary
\appendix
\setcounter{table}{0}

\noindent\textbf{Appendix Contents}
\vspace{2mm}

\begin{itemize}[leftmargin=6mm,itemsep=1mm]
    \item \textbf{\hyperref[supp:training_data]{A. Training Data Curation}}
    \begin{itemize}[leftmargin=5mm,itemsep=0.5mm]
        \item \hyperref[supp:subset1]{A.1 Subset 1: Real instance-level data}
        \item \hyperref[supp:subset2]{A.2 Subset 2: Synthetic data}
        \begin{itemize}[leftmargin=5mm,itemsep=0.5mm]
            \item \hyperref[supp:subset2a]{A.2.1 Subset 2a: Contextual Edits for Generative Synthetic Positives}
            \item \hyperref[supp:subset2b]{A.2.2 Subset 2b: Identity-Altering Edits for Hard Negatives}
        \end{itemize}
    \end{itemize}

    \item \textbf{\hyperref[supp:ablations_main]{B. Ablation Studies}}
    \begin{itemize}[leftmargin=5mm,itemsep=0.5mm]
        \item \hyperref[supp:ablation_datasets]{B.1 Ablation Datasets}
        \item \hyperref[supp:ablations]{B.2 Training Dataset Ablation}
        \item \hyperref[supp:training_ablation]{B.3 Training Ablation}
        \begin{itemize}[leftmargin=5mm,itemsep=0.5mm]
            \item \hyperref[supp:backbone_resolution]{B.3.1 Backbone and Input Resolution}
            \item \hyperref[supp:cls_patch_joint]{B.3.2 CLS vs.\ Patch vs.\ Joint Training}
            \item \hyperref[supp:loss_patch_metric]{B.3.3 Loss Function and Patch Metric}
            \item \hyperref[supp:ablation_summary]{B.3.4 Overall Summary}
        \end{itemize}
    \end{itemize}

    \item \textbf{\hyperref[supp:training_details]{C. Training Details}}
    \begin{itemize}[leftmargin=5mm,itemsep=0.5mm]
        \item \hyperref[supp:model_config]{C.1 Model Configuration}
        \item \hyperref[supp:optimization]{C.2 Optimization}
        \item \hyperref[supp:loss]{C.3 Loss}
        \item \hyperref[supp:data_aug]{C.4 Data Augmentations}
        \item \hyperref[supp:sinkhorn]{C.5 Sinkhorn Patch Metric}
        \item \hyperref[supp:data_loading]{C.6 Data Loading}
    \end{itemize}

    \item \textbf{\hyperref[supp:evaluation]{D. Evaluation}}
    \begin{itemize}[leftmargin=5mm,itemsep=0.5mm]
        \item \hyperref[supp:eval_datasets]{D.1 Evaluation Dataset Details}
        \item \hyperref[supp:subjects2k_pipeline]{D.2 Subjects2k Human Annotation Pipeline}
        \item \hyperref[supp:mllm_eval]{D.3 MLLM Evaluation Criteria}
    \end{itemize}

    \item \textbf{\hyperref[supp:results]{E. Results}}
    \begin{itemize}[leftmargin=5mm,itemsep=0.5mm]
        \item \hyperref[supp:dense_results]{E.1 Dense Results}
        \item \hyperref[supp:full_results]{E.2 Full Results}
    \end{itemize}

    \item \textbf{\hyperref[supp:sensitivity]{F. Analysis}}
    \begin{itemize}[leftmargin=5mm,itemsep=0.5mm]
        \item \hyperref[supp:analysis_image_gen]{F.1 Analysis Image Generation}
    \end{itemize}
\end{itemize}

\vspace{4mm}

\section{Training Data Curation}
\label{supp:training_data}
This section provides technical details of the full curation process: (i) real instance-level data curation decisions such as dataset balancing procedures, criteria used to filter data sources, and more detailed dataset scale information, 
(ii) generative editing pipeline implementation details for subsets~2a and 2b. 

\subsection{Subset 1: Real instance-level data}
\label{supp:subset1}
\paragraph{Initial candidate pool for instance-level data.}

\Cref{tab:initial_pool} lists every dataset originally considered, including
several large instance-level datasets not included in the final main training set
(e.g., Stanford Online Products \cite{song2015deepmetriclearninglifted}, MVImgNet \cite{mvimagenet}, Wildlife-ReID \cite{adam2025wildlifereid} datasets). We focus on instance-level datasets available under research-approved licenses that contain more than 500 unique instances. For each dataset we report:
\begin{itemize}[leftmargin=6mm]
    \item Total number of images, instances, and categories
    \item Per-instance image count range
    \item Type of instance annotation (object ID, catalog ID, animal ID, etc.)
    \item Known annotation issues (if any)
\end{itemize}

\begin{table}[h]
\centering
\scriptsize
\setlength{\tabcolsep}{5pt}
\renewcommand{\arraystretch}{0.9}
\begin{tabular}{lccccc}
\toprule
Dataset & Inst. & Imgs & Cats & Img/Inst & Domain \\
\midrule
MET \cite{ypsilantis2021met}            & 734     & 3,429    & --     & 2--17      & Art \\
ILIAS \cite{kordopatis2025ilias}           & 900     & 5,326    & --     & 2--35      & Everyday Obj \\
FORB  \cite{wu2023forb}           & 4,050   & 16,781   & 7      & 2--22      & Flat Obj \\
GLDv2  \cite{weyand2020google}          & 4,503   & 81,964   & --     & 2--6,272   & Landmarks \\
WildlifeReID10k \cite{adam2025wildlifereid} & 9,756   & 126,302  & 22     & 1--411     & Animals \\
SOP \cite{song2015deepmetriclearninglifted}            & 11,318  & 59,551   & 12     & 2--12      & Products \\
MVImgNet2* \cite{mvimagenet}     & 20,000+ & 689,003  & 300+   & 3--33      & Multi-view Obj \\
DeepFashion  \cite{ge2019deepfashion2}   & 30,018  & 77,221   & 13     & 1--10      & Fashion \\
\midrule
\textbf{Total}  & 80k+    & --       & --     & --         \\
\bottomrule
\end{tabular}
\caption{\textbf{Initial real instance-level dataset pool.} MVImgNet2* is a subset of MVImgNet2 composed of the first two released parts, as the full released dataset contains 180k+ videos. We also use the train\_clean split of GLDv2.}
\label{tab:initial_pool}
\end{table}

\paragraph{Random vs. balanced sampling.}
Because the number of instances across datasets is highly skewed (e.g., 734 instances in MET vs. 30k+ in DeepFashion2), we first evaluated whether
training on a heavily imbalanced mixture would bias the model toward the largest domains. Since ID-Sim is intended to generalize across many visual identities and contexts, we aim to avoid over-representing any specific dataset or domain.

To study the effect of dataset composition, we fix a target pool of 10,000 triplets (30k images) and compare two sampling strategies:
(i) proportional sampling based on raw dataset size, and
(ii) a balanced sampling strategy designed to equalize per-dataset contribution.

As reported in the main paper, balancing improves the
validation ROC AUC from 0.69 to 0.75. We provide the full procedure used to construct the balanced pool below.

\paragraph{Balanced sampling procedure.}
\label{sec:balanced_sampling}
We sample unique instances rather than individual images to maximize identity diversity. The process is as follows:

\begin{enumerate}[leftmargin=6mm]
    \item \textbf{Initial allocation.}
    Each dataset is allocated a quota of 
    $11{,}000 / N_{\text{datasets}}$ instances ($10{,}000$ for the training set and $1{,}000$ for validation),
    giving each dataset an equal starting contribution.

    \item \textbf{Small-dataset allocation.}
    Datasets with fewer instances than their quota (e.g., MET, ILIAS)
    contribute all available instances and are excluded from later steps.

    \item \textbf{Redistribution.}
    The remaining instance budget is divided equally among the remaining datasets.
    This redistribution is repeated until the full $11{,}000$ instance target is reached.

    \item \textbf{Per-instance sampling.}
    For each selected instance, we uniformly sample two images from all available images (one anchor, one positive).

    \item \textbf{Train / validation split.}
    The datasets are then randomly split into $10,000$ instance train set and $1,000$ instance val set. 
\end{enumerate}

After selecting the instances and sampling images, negative pairs are created using 
\textit{hard-negative mining}: for each anchor image, we search the training pool for the nearest neighbor in DINOv3 embedding space. 

This procedure yields two comparable datasets--one proportional and one balanced--
whose final instance counts for the training set are:

\begin{table}[h]
\centering
\scriptsize
\setlength{\tabcolsep}{6pt}
\renewcommand{\arraystretch}{1.05}
\label{tab:balanced_vs_unbalanced}
\begin{tabular}{lcc}
\toprule
Dataset & Unbalanced & Balanced \\
\midrule
MET                    & 105  & 663  \\
ILIAS                  & 111  & 804  \\
FORB                   & 592  & 1428 \\
GLDv2                  & 625  & 1419 \\
WildlifeReID10k        & 1450 & 1418 \\
StanfordOnlineProducts & 1566 & 1435 \\
MVImgNet2              & 3191 & 1411 \\
DeepFashion2           & 2360 & 1422 \\
\midrule
Validation ROC AUC & 0.69 & \textbf{0.75} \\
\bottomrule
\end{tabular}
\caption{\textbf{Instance counts in the unbalanced vs.\ balanced 10k training mixtures.}}
\end{table}

\paragraph{Other dataset filtering criteria and impact on performance.}
We observed inconsistencies in how some datasets defined an ``instance'',
especially relative to the definition used in the main paper (shared visual identity). To evaluate whether 
these inconsistencies affected training quality, we ran an ablation where we applied 
strict filtering rules to remove ambiguous or overly broad instance labels.
Our filtering rules were designed to be simple and reproducible. At a high level, we
removed (i) classes where one instance label covered visually different objects, 
(ii) identities that were extremely difficult to match reliably, and 
(iii) datasets lacking sufficient contextual variation. Below we describe the exact
decisions applied to each dataset.

\begin{figure}[h]
  \centering
  \includegraphics[width=\linewidth]{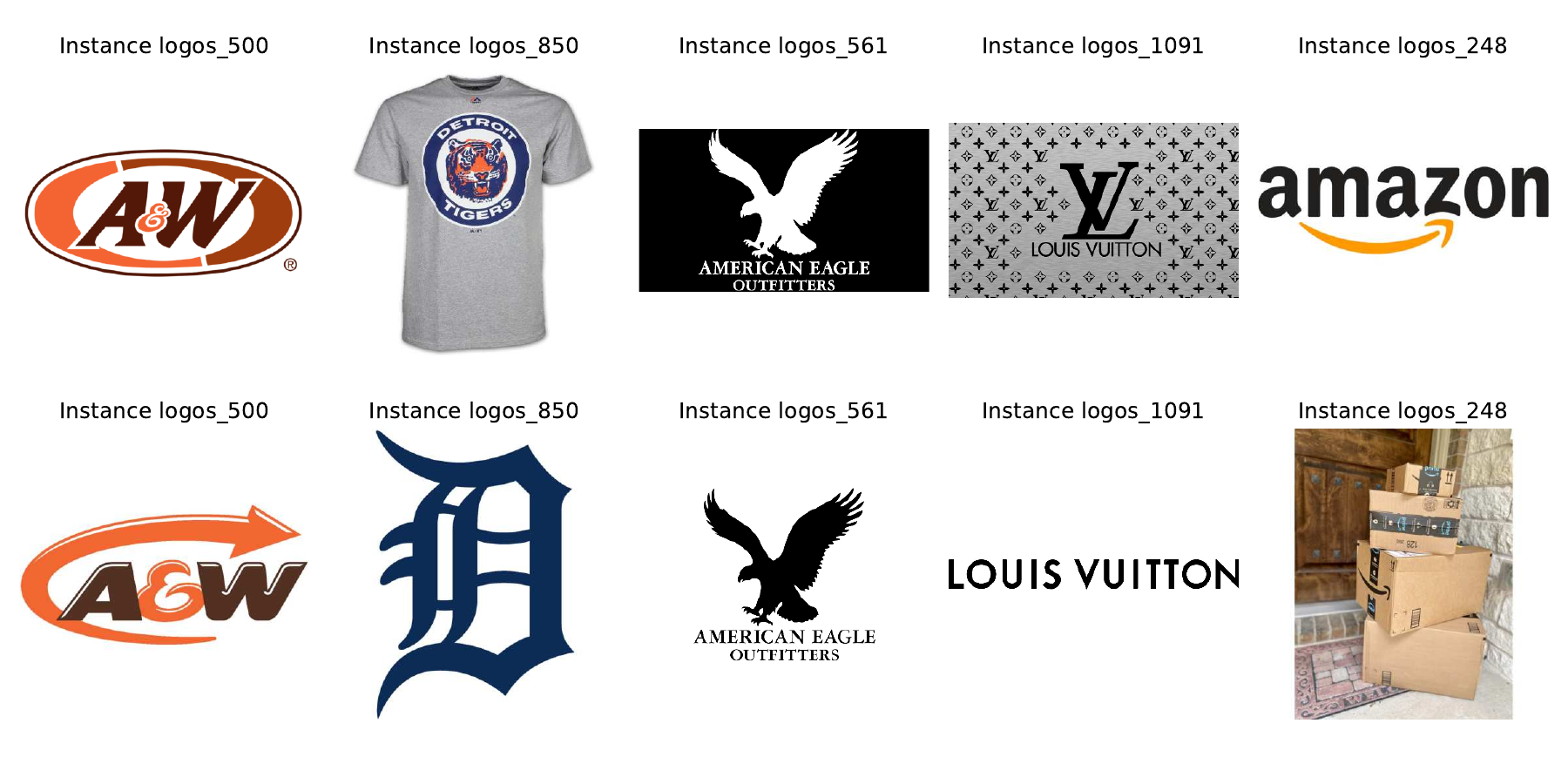}
  \caption{\textbf{Filtered out FORB \texttt{logo} category.} We observe consistent appearance inconsistencies between the same "instance" category in FORB's "logo" class.}
  \label{fig:forb-filtered}
  \vspace{-3mm}
\end{figure}

\begin{figure}[h]
  \centering
  \includegraphics[width=\linewidth]{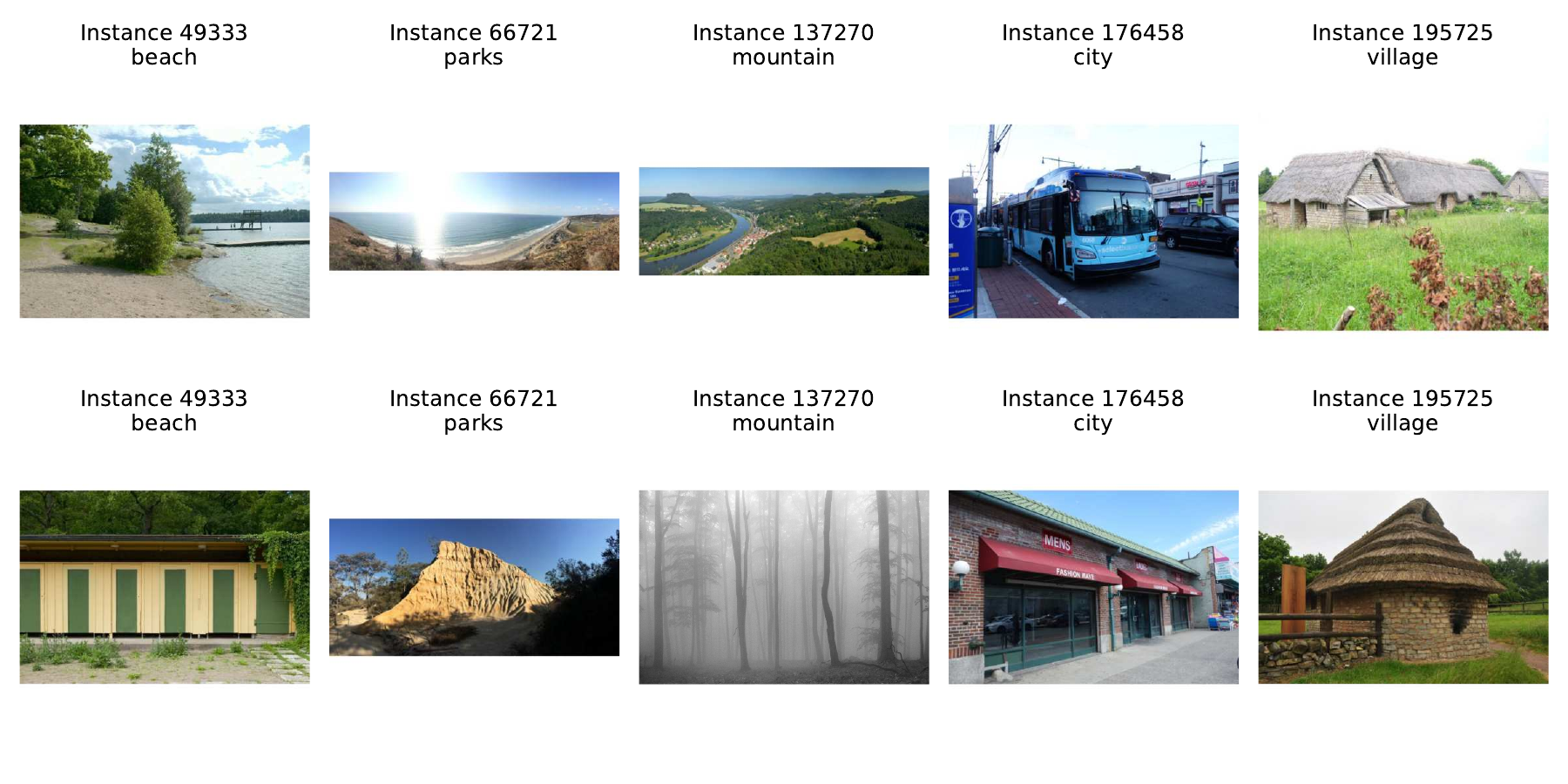}
  \caption{\textbf{Filtered GLDv2 categories.} Many GLDv2 classes cover broad geographic areas rather than a single localized site, building or an object, making it difficult for a class to correspond to a consistent visual identity.}
  \label{fig:gldv2-filtered}
  \vspace{-3mm}
\end{figure}

\begin{enumerate}[leftmargin=6mm]
    \item \textbf{Incorrect instance granularity.}
    Several datasets grouped visually distinct objects under the same instance.
    \begin{itemize}
        \item \textbf{FORB}: We removed the \texttt{Logo} category because different logo
        styles (e.g., the ``LV'' monogram vs. full ``Louis Vuitton'' text) appeared
        under one instance label \Cref{fig:forb-filtered}.
        \item \textbf{GLDv2}: Many GLDv2 categories are too broad to represent a single
        object or a consistent visual identity (see \Cref{fig:gldv2-filtered}). We kept
        only landmark classes where two random images are likely to show the same physical
        structure. Specifically, we retained the following hierarchical labels:
        \texttt{[house, lighthouse, tower, skyscraper, observatory,
        fountain, windmill, sculpture, boat, school, cross, pyramid]}.
        Broad geographic categories such as \texttt{cities}, \texttt{mountains}, and \texttt{villages} were removed.
        \item \textbf{SOP}: Product instances in this dataset often included different colors or 
        versions grouped under the same ID. Because these violate our instance 
        definition, we removed SOP entirely.
    \end{itemize}

    \item \textbf{Hard-to-match or viewpoint-inconsistent identities.}
    Some identities were not mislabeled but were visually too difficult for consistent  
    matching, either due to limited texture cues or extreme viewpoint differences.
    \begin{itemize}
        \item \textbf{WildlifeReID10k}: Certain animal identities (e.g., belugas,
        dolphins) in this dataset have little to no distinctive patterning and appear nearly 
        indistinguishable across individuals. Others include opposite-sided 
        views of the same animal under the same identity, making consistent matching unreliable. To avoid 
        these failure cases, we retained only the DogFaceNet and CatIndividualID 
        subsets, which have stable markings and consistent viewpoints.
    \end{itemize}

    \item \textbf{Insufficient contextual variation.}
    \begin{itemize}
        \item \textbf{MVImgNet}: Although MVImgNet provides rich multi-view rotation,
        it contains very limited background or lighting variation within a single instance sequence since it is a multi-view dataset. Because our 
        training objective requires seeing the same instance under diverse contexts,
        we removed MVImgNet for insufficient contextual diversity.
    \end{itemize}
\end{enumerate}

These filtering steps remove ambiguous labels and ensure that the remaining datasets
align with our visual instance definition. After this filtering, we perform the balancing again following the procedure outlined in \Cref{sec:balanced_sampling}. Together with the sampling, this process produces a cleaner and more consistent dataset, significantly boosting evaluation validation performance from 0.75 to 0.89 (see \Cref{tab:filtered_balanced_dataset}).  

\begin{table}[h]
\centering
\scriptsize
\setlength{\tabcolsep}{6pt}
\renewcommand{\arraystretch}{1.05}
\begin{tabular}{lc}
\toprule
Dataset & Filtered\\
\midrule
MET                    & 671 \\
ILIAS                  & 826 \\
WildlifeReID10k (Dogs and Cats)  & 1501\\
FORB (Filtered)                  & 2346 \\
GLDv2 (Filtered)                  & 2315 \\
DeepFashion2           & 2341 \\
\midrule
Validation ROC AUC & \textbf{0.89} \\
\bottomrule
\end{tabular}
\caption{\textbf{Instance counts for the filtered and balanced 10k training mixture.}}
\label{tab:filtered_balanced_dataset}
\end{table}

\subsection{Subset 2: Synthetic data}
\label{supp:subset2}
While the real instance-level datasets in Subset~1 provide strong coverage of
identity-preserving variation, they underrepresent many forms of contextual change
(e.g., background, lighting, scene geometry). To address this limitation, we augment
our training set with synthetic data generated through controlled editing. These
edits preserve the object's visual identity while introducing new contexts that
rarely appear in the original datasets.

We use two complementary sources of synthetic data. \textbf{Subset 2a} applies generative
contextual edits (background and lighting changes) to isolated frames sampled from
video datasets.
\textbf{Subset 2b} applies generative foreground edits to create hard-negative examples for contrastive triplets.

\subsubsection{Subset 2a: Contextual Edits for Generative Synthetic Positives}
\label{supp:subset2a}
\paragraph{Base datasets.}
Subset 2a is constructed from sequential video datasets rather than independent
images of an instance.  Video sources are particularly valuable because they capture the
same instance under natural pose variation but often lack diversity in background
and illumination. Generative editing conditioned on these real frames therefore
injects controlled contextual diversity while maintaining identity fidelity. We use LaSOT~\cite{fan2020lasothighqualitylargescalesingle},
GOT10k~\cite{Huang_2021}, YouTubeVIS~\cite{yang2019videoinstancesegmentation}, and
UCO3D~\cite{liu24uco3d}, chosen for their instance diversity, scale, and availability
of mask annotations.  To obtain a diverse and high-quality set of frames per instance, we use a
simple dataset-adaptive sampling strategy:

\begin{itemize}[leftmargin=6mm]
    \item \textbf{Long videos ($>$6s):} divide each sequence into 5--6 equal-duration
    parts and sample one valid frame from each part
    \item \textbf{Short videos ($\leq$6s):} sample every 
    $k_{\text{annotated}} \times \text{annotation\_stride}$ frames
\end{itemize}
Dataset-specific values for frame rate, annotation stride, window size, and number
of segments are provided in \Cref{tab:video_sampling_params}. These parameters are
chosen so that each sampling window corresponds to roughly 1--2 seconds of video, preventing oversampling of near-duplicate frames.

Within each sampling window, we apply the following quality filters:
\begin{itemize}[leftmargin=6mm]
    \item \textbf{Foreground coverage:} between 10\% and 90\%
    \item \textbf{Sharpness:} blur score \cite{rosebrock2015blur}. $>$50, where the blur score is computed as the 
    variance of the Laplacian (higher variance indicates a sharper image with stronger edges)
\end{itemize}

If multiple frames satisfy these criteria, we randomly select one to encourage
temporal diversity. Instances are retained only if at least two valid frames are
obtained.

\begin{table}[h]
\centering
\scriptsize
\setlength{\tabcolsep}{2pt}
\renewcommand{\arraystretch}{0.9}
\label{tab:video_sampling_params}
\begin{tabular}{l c c c c c}
\toprule
Dataset &
\makecell{Frames per\\Second (FPS)} &
\makecell{Frame\\Stride} &
\makecell{Number of\\Frames ($k$)} &
\makecell{Window\\Size} &
\makecell{Number of\\Sampled \\Parts} \\
\midrule
LaSOT        & 30 & 5 & 6  & 30 (1s)   & 6 \\
GOT-10k      & 10 & 5 & 2  & 10 (1s)   & 5 \\
YouTubeVIS   & 6  & 5 & 2  & 10 (1.7s) & 5 \\
UCO3D        & 30 & 1 & 30 & 30 (1s)   & 5 \\
\bottomrule
\end{tabular}
\caption{\textbf{Dataset-specific parameters for video frame sampling.} About 5-6 frames are sampled from each instance sequence, at intervals that are at $\sim$1 second apart.}
\end{table}

\paragraph{Editing model and pipeline.}
Contextual edits are produced using the Qwen-Image-Edit \cite{wu2025qwenimagetechnicalreport} diffusion model
(\texttt{Qwen/Qwen-Image-Edit}) with Lightning LoRA weights, enabling 8-step
inference. All generations use bfloat16 precision and a
\texttt{FlowMatchEulerDiscreteScheduler}. A fixed generator seed ensures
determinism.

\paragraph{Preprocessing.}
Each selected frame is paired with its binary foreground mask. Before editing, we
apply:

\begin{itemize}[leftmargin=6mm]
    \item Foreground crop preserving a 2:3 or 3:2 aspect ratio.
    \item Resize so the longer side is at most 1248~px and both dimensions are divisible
    by 32.
    \item Foreground scaling: if the mask covers less than 10\%, scale up to exactly
    10\%; otherwise randomly select a scale factor so the new coverage lies between 10\%
    and the original value. Each instance is assigned a scale mode (small or large).
    \item Placement of the scaled foreground onto a white canvas without border clipping.
    \item Composition of the foreground onto the blank canvas to form the editing input.
\end{itemize}

\paragraph{Prompts.}
Each frame receives a unique background–lighting combination. Background prompts are
sampled from a supercategory-specific list using
\texttt{category\_to\_supercat.json} and
\texttt{supercat\_to\_backgrounds.json}. Lighting prompts come from
\texttt{lighting\_prompts.json} and are conditioned on the selected background using
\texttt{background\_to\_scene.json}. The prompts are generated using GPT-4o \cite{hurst2024gpt}. All component files are included in the
supplemental.

The prompt used for all contextual edits is:

\begin{quote}
\small
\texttt{Replace only the white background pixels with \{background\}; keep the foreground objects and text completely unchanged in size, position, orientation, and appearance (except lighting); preserve original text, composition, proportions, alignment, and text properties; seamlessly blend the new background with simulated \{lighting\_prompt\} to match scene lighting, shadows, and reflections; ensure natural integration without duplication, movement, or distortion of the foreground; maintain original dimensions, aspect ratio, and focal center; adjust foreground lighting for seamless blending.}
\end{quote}

\paragraph{Parameter sampling.}
Contextual edits use fixed settings: 8 inference steps, guidance scale of 1.0, and a
fixed generator seed. Background and lighting indices are sampled per frame.

\paragraph{Generation and output.}
The model replaces only white-background pixels with the selected scene while
blending foreground lighting to match. Outputs include the edited RGB image and
updated mask, saved with deterministic filenames encoding all edit parameters.

\begin{figure}[h]
  \centering
  \includegraphics[width=\linewidth]{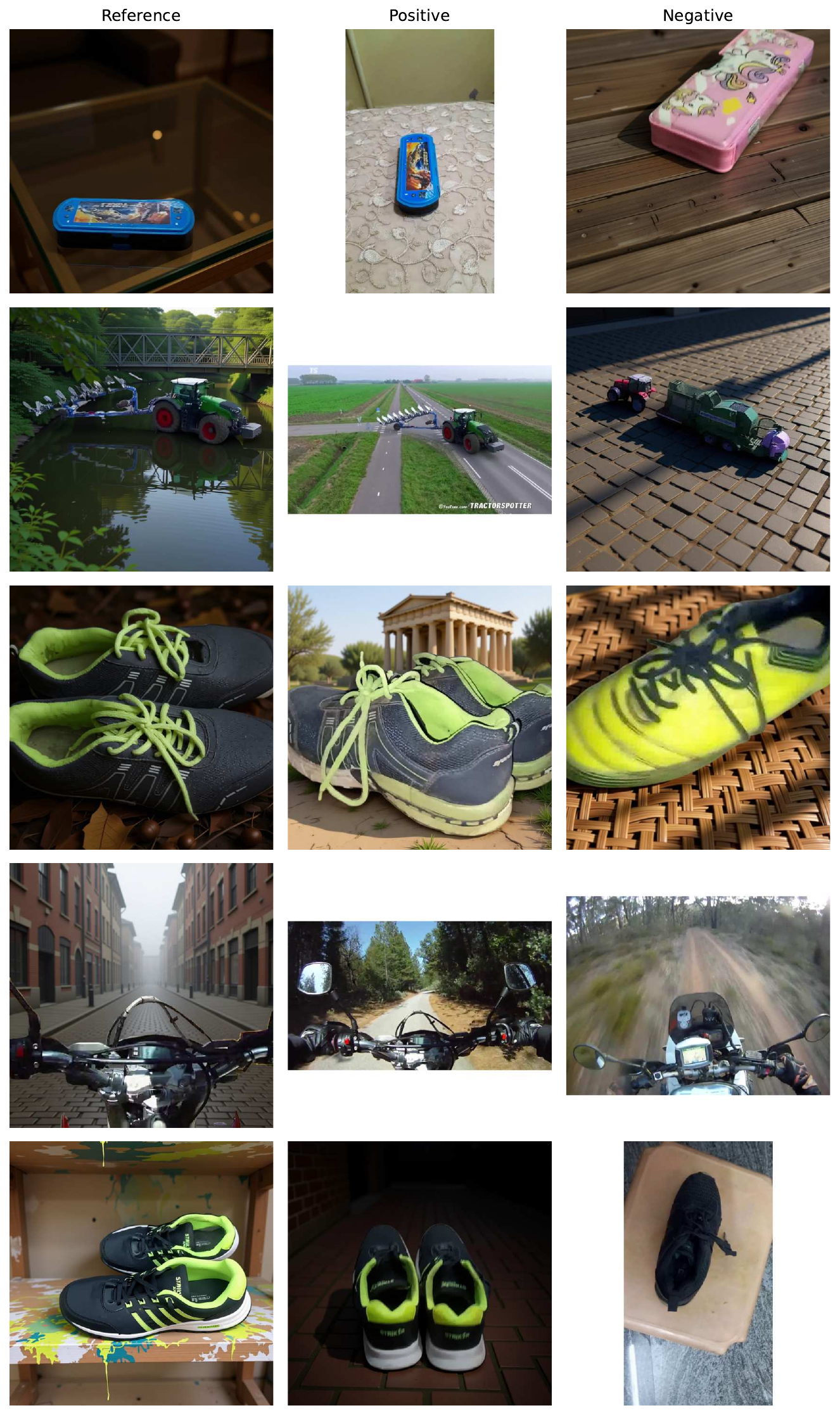}
  \caption{\textbf{Generative Contextual Edited Images} }
  \label{fig:pos-edit}
  \vspace{-3mm}
\end{figure}

\paragraph{Use in training.}
During training, triplets
are formed by mixing original and edited views of the same instance. Each positive pair
is chosen uniformly from:
\begin{enumerate}[leftmargin=6mm]
    \item original anchor + edited positive
    \item edited anchor + original positive
    \item edited anchor + edited positive
\end{enumerate}
The negative is drawn from either an edited or original view of a different instance.
Examples are shown in Figure~\ref{fig:pos-edit}. Adding this dataset in a 1:1 mix with Subset 1 results in a \textbf{validation ROC AUC improvement from 0.89 to 0.937}, as reported in the main paper, suggesting that the diversification of contextual edits helps. The ablation on the dataset composition is in \Cref{supp:ablations}.

\subsubsection{Subset 2b: Identity-Altering Edits for Hard Negatives}
\label{supp:subset2b}

Subset~2b introduces controlled foreground edits that alter identity-defining
features while maintaining class-level semantics. These edits produce realistic but
non-matching variants of each object and are used exclusively as hard negatives.

\paragraph{Base datasets.}
We apply identity edits only to datasets with high-quality segmentation masks:
DeepFashion2~\cite{ge2019deepfashion2} and UCO3D~\cite{liu24uco3d}. These datasets
provide clean boundaries and stable viewpoints. The remaining video datasets contain
motion blur or noisy masks, and the real instance-level datasets from Subset~1 do not
provide per-object masks.

\paragraph{Model and pipeline.}
Foreground edits are generated using the \texttt{FluxFillPipeline}
(\texttt{black-forest-labs/FLUX.1-Fill-dev})~\cite{flux2024}. The model operates in
bfloat16 precision and performs inpainting only where the input mask is white. The
background is preserved unchanged.

\begin{figure}[t]
  \centering
  \includegraphics[width=\linewidth]{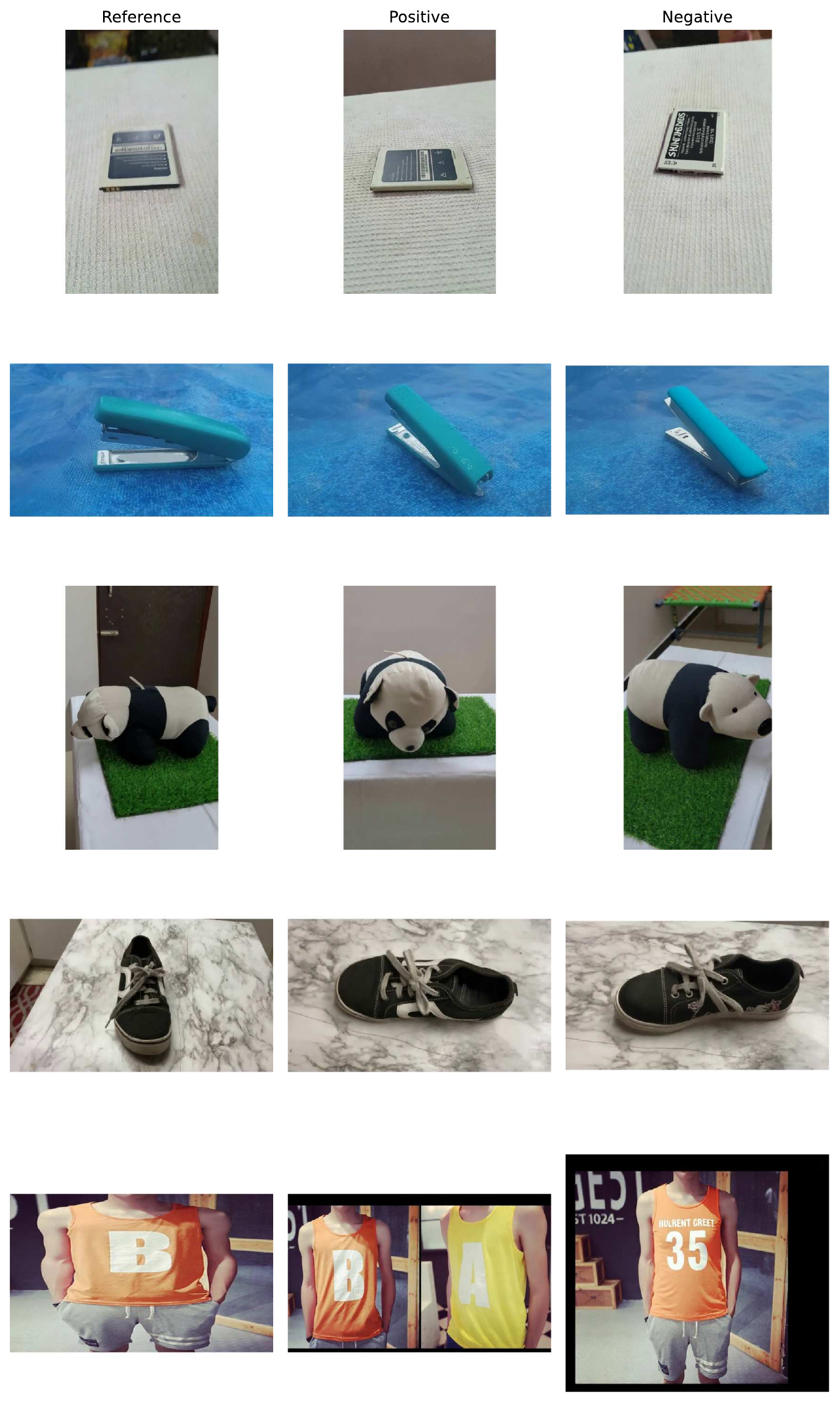}
  \caption{\textbf{Generative Edited Hard-Negatives} }
  \label{fig:neg-edit}
  \vspace{-3mm}
\end{figure}

\paragraph{Preprocessing.}
All images and masks are resized so the longer side is at most 720~px, while preserving
aspect ratio and ensuring both dimensions are divisible by eight. We apply randomized partial masking by removing 40–60\% of the
foreground region along a horizontal or vertical axis to introduce occlusion and
increase edit diversity.

\paragraph{Prompts.}
All identity edits use a class-specific prompt of the form:

\begin{quote}
\small
\texttt{Photo of a <object>}
\end{quote}

The object name is drawn from \texttt{instance\_class\_to\_name.json}. These prompts
encourage edits that remain faithful to class semantics while modifying fine-grained
appearance cues.

\paragraph{Parameter sampling.}
For each frame we sample:
\begin{itemize}[leftmargin=6mm]
    \item Strength in the range 0.5--0.8
    \item 50 inference steps
    \item Guidance scale of 2.5
\end{itemize}
Sampling is controlled by a per-frame seed computed as \texttt{seed + row\_id}. The
generation step uses a fixed internal seed of zero, enabling fully deterministic output.

\paragraph{Generation and output.}
The pipeline inpaints only the masked foreground, producing realistic but
identity-altered variants. Outputs are resized back to the original crop resolution
and saved with deterministic filenames encoding all parameters.

\paragraph{Use in training.}
Identity-edited images are used only as hard negatives. To avoid the model relying on
generative artifacts to identify these negatives, we add mild generative noise
(strength~0.1) to the anchor and positive whenever a triplet includes an
identity-edited negative. This noise does not change image content but prevents
artifact-based shortcuts. Negatives are sampled from both edited and original images
of other instances. Examples appear in Figure~\ref{fig:neg-edit}. Adding this dataset in a 1:1:1 mix with Subset 1 results in a \textbf{validation ROC AUC improvement from 0.937 to 0.965}, as reported in the main paper, suggesting that the edited hard-negatives provide additional signal. The ablation on the dataset composition is in the following section.

\section{Ablation Studies}
\label{supp:ablations_main}

\subsection{Ablation Datasets}
\label{supp:ablation_datasets}

For model selection we rely on two complementary validation sets.  
First, we use the validation split of the curated real instance dataset from Subset~1, which we refer to as \textbf{Real Instance Validation}. The metric on this set is accuracy as it is in a triplet format.  
Second, we construct a small identity-focused validation set composed of five Flux-generated base instances, each edited with a mixture of identity-preserving and identity-altering Photoshop \cite{adobephotoshop} modifications.  
We evaluate binary identity classification on this set and refer to the resulting score as \textbf{Identity Validation}.  

This identity-focused set is intentionally small, does not overlap with any test data, and provides an additional targeted signal that complements the broader Real Instance Validation set.  
Throughout the ablation experiments, we report results on both validation sets and use them jointly to identify the best-performing configuration.

\label{supp:ablations}
\subsection{Training Dataset Ablation.}
To determine the appropriate training scale and composition, we evaluate both factors using two complementary validation metrics. As shown in Table~\ref{tab:dataset_scale_ablation}, performance increases gradually when scaling from 5k to 20k instances, but the gains are modest and within the range of expected variance. Beyond 20k, identity-validation performance decreases substantially. This pattern suggests that the main benefits are already achieved at moderate dataset sizes, and that 10k instances provide a stable and efficient operating point.

\begin{table}[h]
\centering
\small
\setlength{\tabcolsep}{6pt}
\renewcommand{\arraystretch}{1.15}
\begin{tabular}{lcc}
\toprule
\textbf{Dataset Scale} &
\makecell{\textbf{Real Instance}\\\textbf{Validation (Acc)}} &
\makecell{\textbf{Identity}\\\textbf{Validation (ROC AUC)}} \\
\midrule
5k even split  & 0.860  & \underline{0.97} \\
10k even split & 0.870  & \underline{0.97} \\
20k even split & \underline{0.880} & \textbf{0.98} \\
30k            & \textbf{0.8825}   & 0.91 \\
\bottomrule
\end{tabular}
\caption{\textbf{Effect of training dataset scale on validation performance.}}
\label{tab:dataset_scale_ablation}
\end{table}

Next, we evaluate dataset composition under a fixed 10k training set, comparing an even (1:1:1) mixture to configurations where one subset is made dominant (0.7{:}0.15{:}0.15). As shown in Table~\ref{tab:dataset_composition_ablation}, the even split achieves the best balance across both validation metrics, whereas skewed mixtures improve one metric at the expense of the other.

\begin{table}[h]
\centering
\small
\setlength{\tabcolsep}{3pt}
\renewcommand{\arraystretch}{0.95}
\begin{tabular}{lcc}
\toprule
\textbf{Dataset Composition} &
\makecell{\textbf{Real Instance}\\\textbf{Validation (Acc)}} &
\makecell{\textbf{Identity}\\\textbf{Validation (ROC AUC)}} \\
\midrule
Even Split             & \underline{0.8715}              & \textbf{0.97} \\
Subset 1 dominant      & \textbf{0.8905}     & 0.95 \\
Subset 2a dominant      & 0.8715              & 0.95 \\
Subset 2b dominant      & 0.8685              & \underline{0.96} \\
\bottomrule
\end{tabular}
\caption{\textbf{Effect of dataset composition on validation performance (fixed 10k scale).}}
\label{tab:dataset_composition_ablation}
\end{table}

Based on these findings, we adopt the \textbf{10k even-split} configuration as our final training mixture, providing strong and stable performance across real-instance and identity-level evaluations.

\subsection{Training Ablation}
\label{supp:training_ablation}
All ablations in this section are evaluated using a \textbf{joint CLS+patch embedding} and are trained using the 10k balanced training mixture.

We conduct ablations to isolate the contribution of the backbone, feature losses, and patch similarity metrics. These models are evaluated on two validation sets: (i) Real Instance Validation (accuracy) and (ii) Identity Validation (ROC AUC).

\vspace{1mm}
\subsubsection{Backbone and Input Resolution}
\label{supp:backbone_resolution}

\begin{table}[h]
\centering
\footnotesize
\setlength{\tabcolsep}{3pt}
\renewcommand{\arraystretch}{0.75}
\begin{tabular}{lccc}
\toprule
\textbf{Backbone / Resolution} &
\makecell{\textbf{Real Instance}\\\textbf{Validation (Acc)}} &
\makecell{\textbf{Identity}\\\textbf{Validation (ROC AUC)}} \\
\midrule
DINOv3-L/16 @ 448 (Baseline) & \textbf{0.8715} & \textbf{0.965} \\
DINOv3-L/16 @ 224            & \textbf{0.8715} & \textbf{0.965} \\
DINOv3-B/16 @ 448            & 0.834 & 0.921 \\
DINOv2-L/16 @ 448            & 0.8355 & 0.895 \\
DINOv2-L/16 @ 224            & 0.8135 & 0.819 \\
\bottomrule
\end{tabular}
\caption{\textbf{Backbone and resolution ablation.}}
\label{tab:backbone_resolution_ablation}
\end{table}

DINOv3 performs better than DINOv2 across both validation sets, and resolution mainly affects DINOv2, but higher resolution results in better performance. ViT-L architecture outperforms ViT-B. This supports using \textbf{DINOv3-L/16 at 448px} in the final model.

\vspace{2mm}
\subsubsection{CLS vs.\ Patch vs.\ Joint Training}
\label{supp:cls_patch_joint}

\begin{table}[h]
\centering
\footnotesize
\setlength{\tabcolsep}{4pt}
\renewcommand{\arraystretch}{1.1}
\begin{tabular}{lcc}
\toprule
\textbf{Feature Loss Setting} &
\makecell{\textbf{Real Instance}\\\textbf{Validation (Acc)}} &
\makecell{\textbf{Identity}\\\textbf{Validation (ROC AUC)}} \\
\midrule
CLS + Patch (Baseline) & \textbf{0.8715} & \textbf{0.965} \\
Patch Loss Only        & 0.8665 & 0.908 \\
CLS Loss Only          & 0.8065 & 0.893 \\
\bottomrule
\end{tabular}
\caption{\textbf{Ablation on CLS vs.\ patch vs.\ joint training.}}
\label{tab:feature_loss_ablation}
\end{table}

Joint supervision combines the complementary strengths of CLS and patch features, resulting in stronger overall performance than using either feature in isolation.

\vspace{2mm}
\subsubsection{Loss Function and Patch Metric}
\label{supp:loss_patch_metric}

\begin{table}[h]
\centering
\footnotesize
\setlength{\tabcolsep}{4pt}
\renewcommand{\arraystretch}{1.1}
\begin{tabular}{lcc}
\toprule
\textbf{Objective / Patch Metric} &
\makecell{\textbf{Real Instance}\\\textbf{Validation (Acc)}} &
\makecell{\textbf{Identity}\\\textbf{Validation (ROC AUC)}} \\
\midrule
InfoNCE + Sinkhorn (Baseline)  & \textbf{0.8715} & \textbf{0.965} \\
InfoNCE + Cosine Patch Metric  & 0.8655 & 0.940 \\
Hinge + Sinkhorn               & 0.8705 & 0.945 \\
BCE + Sinkhorn                 & 0.8395 & 0.923 \\
\bottomrule
\end{tabular}
\caption{\textbf{Loss and patch similarity ablation.}}
\label{tab:objective_patchmetric_ablation}
\end{table}

Sinkhorn OT improves patch alignment over cosine distance, and InfoNCE provides the strongest identity separation among the tested objectives. These results support the choice of \textbf{InfoNCE with Sinkhorn} in the final model.

\vspace{2mm}

\subsubsection{Overall Summary}
\label{supp:ablation_summary}

Across all ablations, the combination of DINOv3-L/16, joint CLS+patch training, and Sinkhorn OT produces the most reliable identity-sensitive behavior and is therefore adopted in all main-paper experiments.

\section{Training Details}
\label{supp:training_details}
With the architecture and dataset design fixed as above, we ablated over the key 
training hyperparameters and arrived at the following final configuration.

\subsection{Model Configuration}
\label{supp:model_config}
\begin{itemize}[leftmargin=6mm]
    \item \textbf{Backbone}: DINOv3-ViT-L/16 (stride 16), using CLS and patch features
    \item \textbf{Head}: dual MLPs with 512-dim hidden layers (CLS and patch)
    \item \textbf{LoRA adaptation}: rank 16, $\alpha=32$, dropout 0.05
    \item \textbf{Input resolution}: $448\times448$
    \item \textbf{Precision}: \texttt{bfloat16}
\end{itemize}

\subsection{Optimization}
\label{supp:optimization}
\begin{itemize}[leftmargin=6mm]
    \item \textbf{Optimizer}: AdamW, learning rate $3\times10^{-4}$, weight decay $0$
    \item \textbf{Batch size}: 8 (effective batch size 32 with $\times4$ gradient accumulation)
    \item \textbf{\# epochs}: 3
\end{itemize}

\subsection{Loss}
\label{supp:loss}
\begin{itemize}[leftmargin=6mm]
    \item \textbf{Objective}: InfoNCE with single-negative sampling
    \item \textbf{Margin}: 0.1
    \item \textbf{Feature weighting}: CLS : Patch = 1 : 1
    \item \textbf{Patch alignment}: Sinkhorn optimal transport
\end{itemize}

\subsection{Data Augmentations}
\label{supp:data_aug}
\begin{itemize}[leftmargin=6mm]
    \item Random resized crop (scale 0.9--1.0; aspect ratio 1:1; bicubic)
    \item Color jitter (brightness 0.2, contrast 0.2, saturation 0.08, 
          probability 0.8)
    \item Gaussian blur (kernel $7\times7$, $\sigma\in[0.05,0.6]$, 
          probability 0.5)
\end{itemize}

\subsection{Sinkhorn Patch Metric}
\label{supp:sinkhorn}
\begin{itemize}[leftmargin=6mm]
    \item \textbf{Implementation}: \texttt{geomloss.SamplesLoss} with $p=2$
    \item \textbf{Regularization}: 0.05
    \item \textbf{Blur}: 0.05
    \item \textbf{Maximum tokens}: 1024
    \item Patch features L2-normalized before distance computation
\end{itemize}

\subsection{Data Loading}
\label{supp:data_loading}
\begin{itemize}[leftmargin=6mm]
    \item 4 dataloader workers
    \item Up to 3 concurrent S3 downloads
    \item Train/val splits loaded from S3 parquet files 
\end{itemize}

\section{Evaluation}
\label{supp:evaluation}
\subsection{Evaluation Dataset Details}
\label{supp:eval_datasets}
We summarize here the seven evaluation datasets used in the main paper. Each dataset follows its standard evaluation protocol and is fully disjoint from the training data. The only exception is DeepFashion2, for which a subset of the dataset is used during training; however, the evaluation split employed here is strictly non-overlapping with the training split.

\subsection*{PODS (Instance Retrieval)}
\begin{figure}[h]
  \centering
  \includegraphics[width=\linewidth]{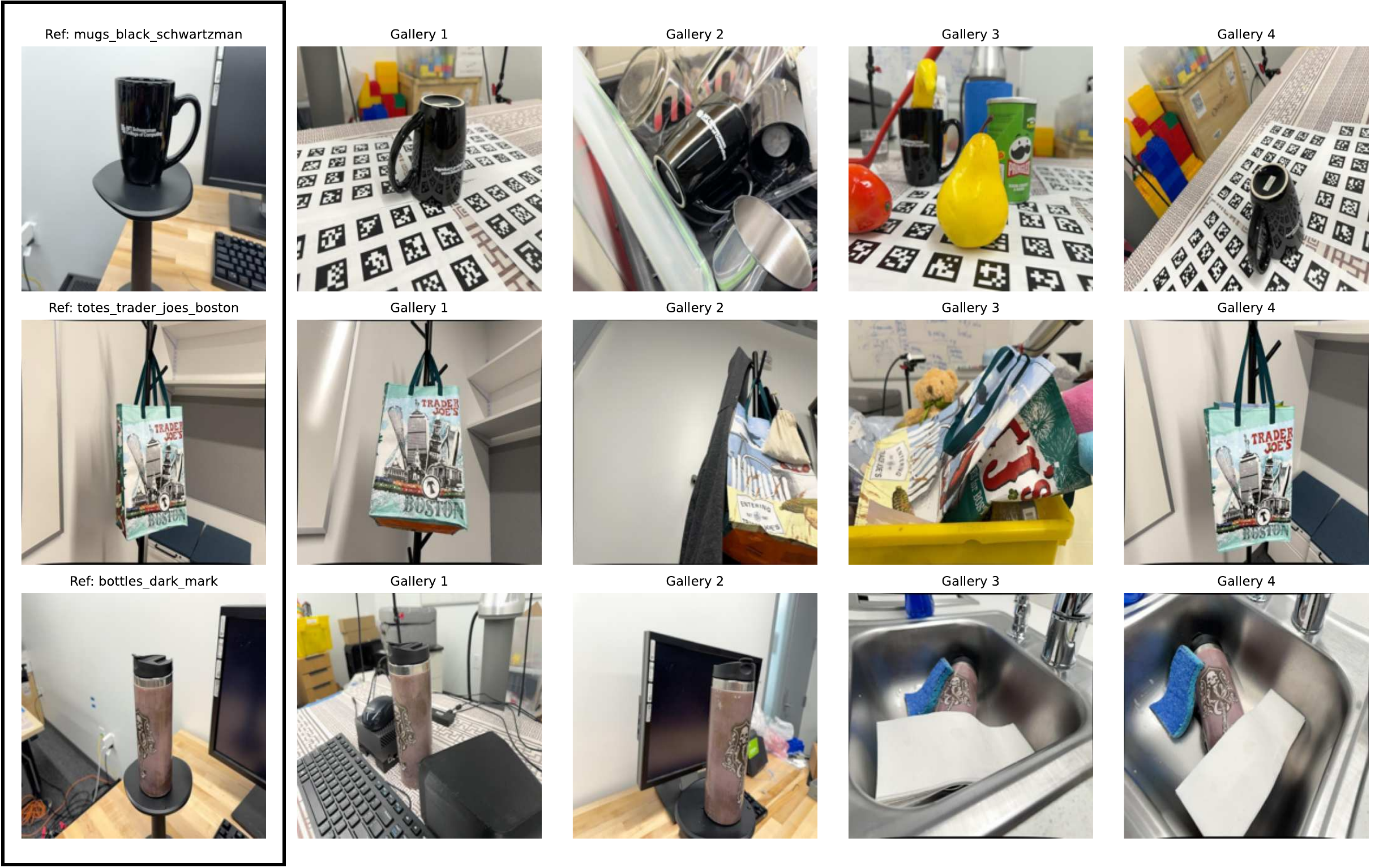}
  \caption{\textbf{PODS Dataset.} The dataset is composed of household objects occurring under different distribution shifts, with varying backgrounds, distractor objects, and poses.}
  \label{fig:pods}
  \vspace{-3mm}
\end{figure}

\begin{itemize}[leftmargin=2em]
\item \textbf{Task:} Instance retrieval for personalized household objects (\Cref{fig:pods}).
\item \textbf{Size:} 1{,}200 query images and 300 gallery images.
\item \textbf{Instances:} 100 object instances appearing in both splits.
\item \textbf{Labels:} Instance-level ID.
\item \textbf{Protocol:} We evaluate using the dataset’s canonical setup: the 1{,}200 images from the \texttt{test\_dense} split serve as queries, and the 300 images from the \texttt{train} split serve as the gallery. Although the split is named “train,” it is only part of the dataset organization and is not used to train our model.
\item \textbf{Metrics:} mAP (main), ROC-AUC, nDCG. Normalized Discounted Cumulative Gain (nDCG) evaluates how well the ranking prioritizes the most relevant matches.
\item \textbf{Notes:} Images show controlled variation in viewpoint, background, and lighting across all instances.
\end{itemize}

\subsection*{DeepFashion2 (Instance Retrieval)}
\begin{figure}[h]
  \centering
  \includegraphics[width=\linewidth]{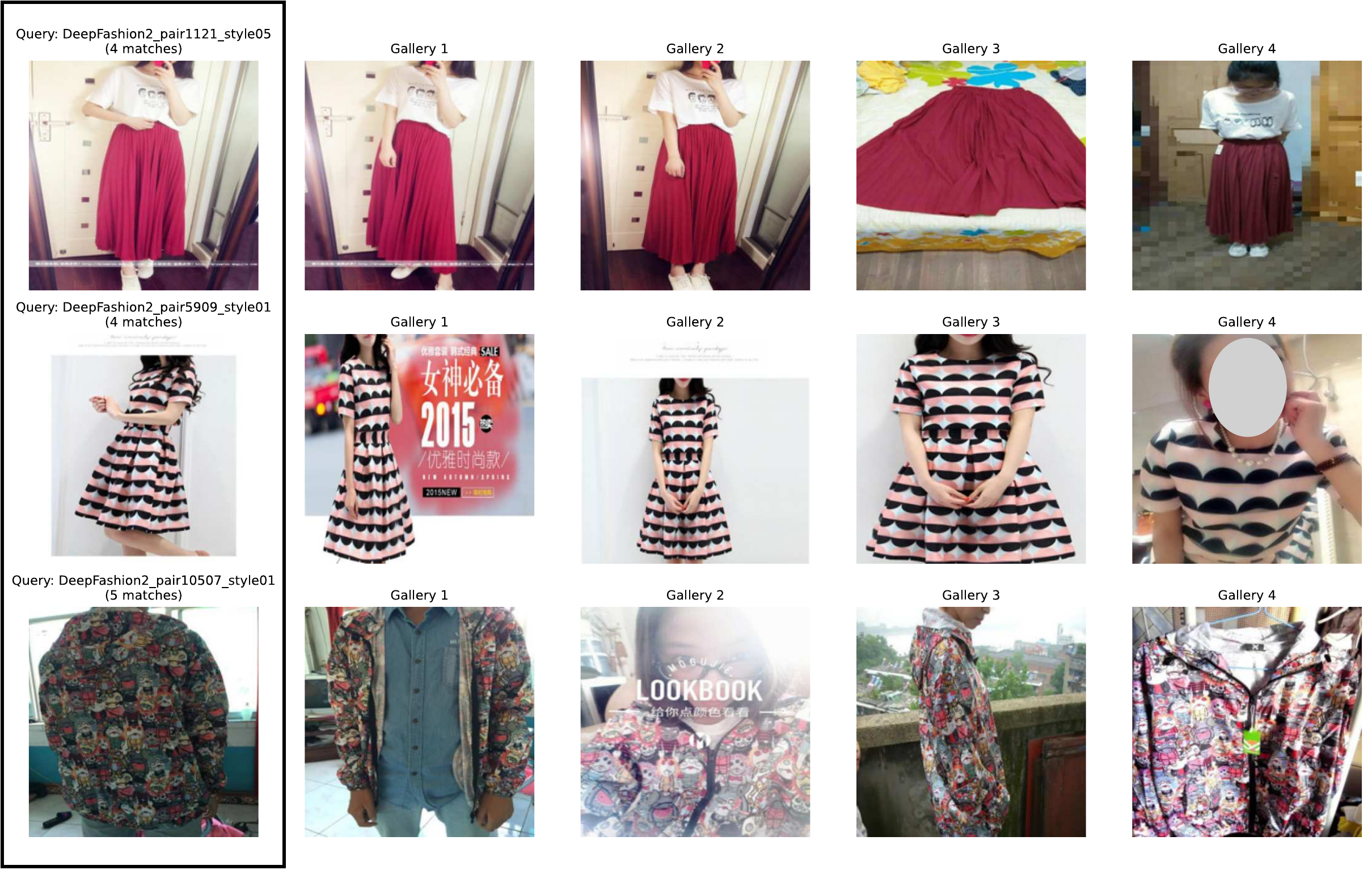}
  \caption{\textbf{DeepFashion2.} The DeepFashion2 dataset features query / gallery images of the same clothing item in-shop as well as worn by consumers.}
  \label{fig:df2}
  \vspace{-3mm}
\end{figure}

\begin{itemize}[leftmargin=2em]
\item \textbf{Task:} Clothing-item instance matching across domains (\Cref{fig:df2}).
\item \textbf{Size:} 1,668 queries; 3,065 gallery images
\item \textbf{Instances:} 1,668 clothing items
\item \textbf{Labels:} Per-item instance ID
\item \textbf{Protocol:} Standard fashion retrieval (each query has at least one gallery match)
\item \textbf{Metrics:} mAP (main), ROC-AUC
\end{itemize}

\subsection*{AerialCattle2017 (Animal Re-ID)}
\begin{figure}[h]
  \centering
  \includegraphics[width=\linewidth]{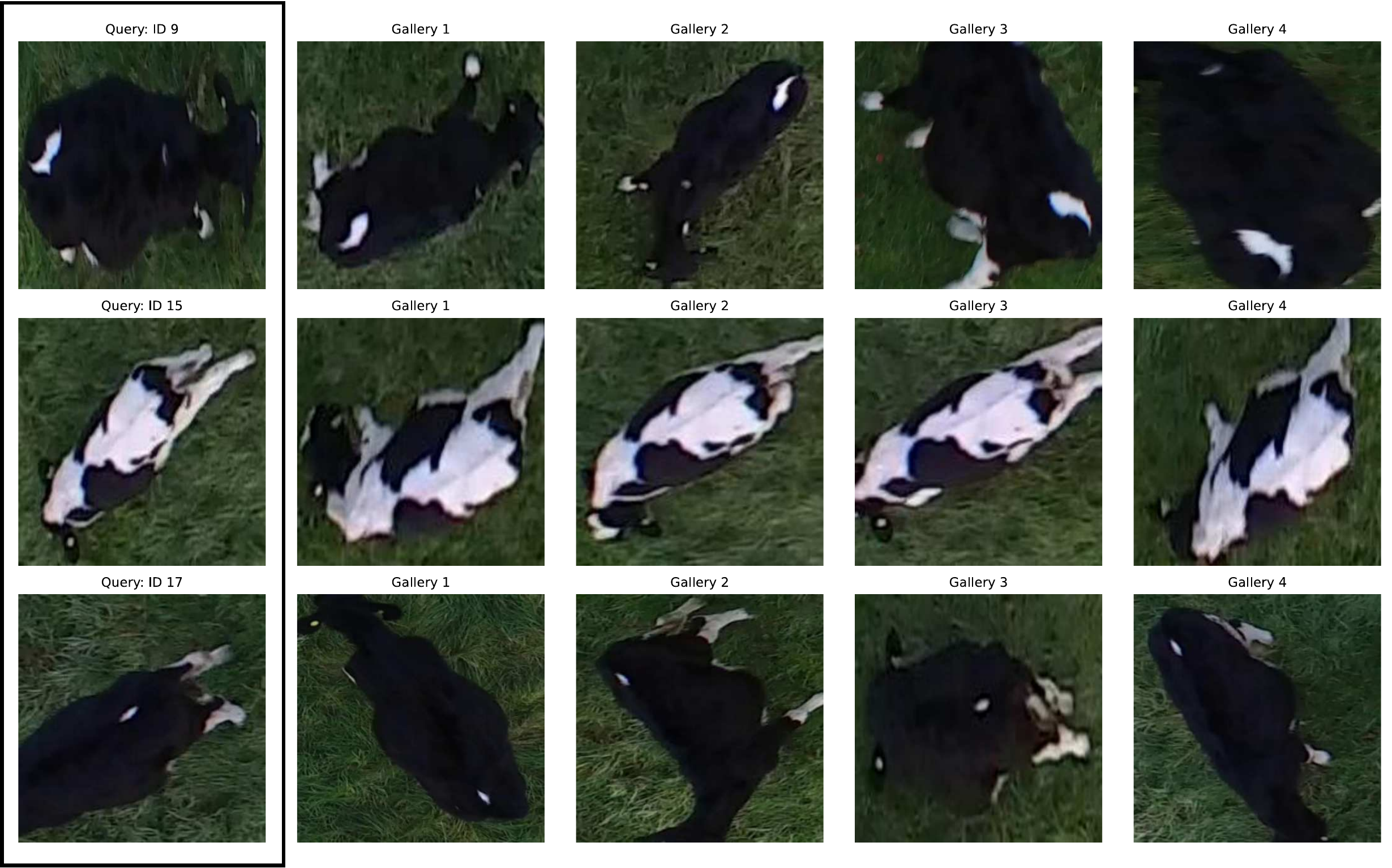}
  \caption{\textbf{AerialCattle2017.} This dataset is composed of aerial imagery of various cows on fields, and the task is to retrieve the same individuals based on a query image.}
  \label{fig:cows}
  \vspace{-3mm}
\end{figure}

\begin{itemize}[leftmargin=2em]
\item \textbf{Task:} Animal identity retrieval from aerial imagery (\Cref{fig:cows})
\item \textbf{Size:} 2,329 filtered images
\item \textbf{Identities:} 23 cattle
\item \textbf{Splits:} 23 queries; 2,306 gallery images
\item \textbf{Labels:} Individual animal ID
\item \textbf{Protocol:} Rank gallery images for each query
\item \textbf{Metrics:} mAP (main).
\end{itemize}

\subsection*{PetFace (Animal Re-ID - Verification)}
\begin{itemize}[leftmargin=2em]
\item \textbf{Task:} Pairwise identity verification across 13 species
\item \textbf{Size:} 3,250 pairs
\item \textbf{Labels:} 1,622 positive; 1,628 negative
\item \textbf{Species:} 13 unique species: cat, chimp, chinchilla, degus, dog, ferret, guineapig, hamster, hedgehog, javasparrow, parakeet, pig, rabbit
\item \textbf{Protocol:} Predict whether two images depict the same individual
\item \textbf{Metrics:} mAP (main)
\end{itemize}

\begin{figure}[h]
  \centering
  \includegraphics[width=\linewidth]{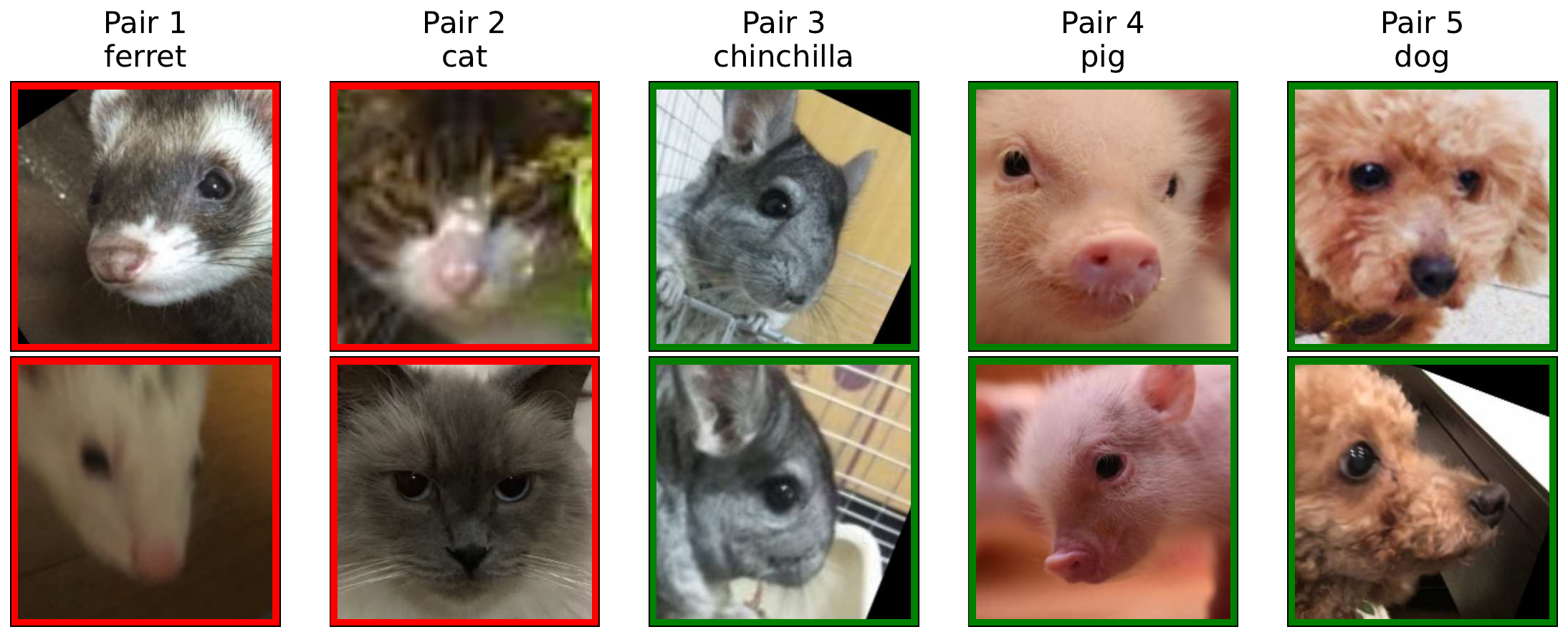}
  \caption{\textbf{Petface.} Evaluation benchmark of 13 unseen animals. Red depicts different individual and green depicts same individual.}
  \label{fig:petface}
  \vspace{-3mm}
\end{figure}

\subsection*{CUTE (Triplet Matching)}
\begin{itemize}[leftmargin=2em]
\item \textbf{Task:} Fine-grained object discrimination using triplet matching
\item \textbf{Size:} 1,800 triplets
\item \textbf{Structure:} Each sample contains an anchor, a positive (same instance), and a negative (different instance)
\item \textbf{Modes:}  
      1) \emph{Easy} mode uses triplets in which all three images come from the same scene, testing discrimination between instances under identical background and context,
      2) \emph{Hard} mode selects the anchor from a different scene whenever possible while keeping the positive and negative in the same scene; this requires recognizing the same instance across scene changes while rejecting a same-scene negative \textbf{}
\item \textbf{Protocol:} Predict whether the anchor is more similar to the positive than the negative
\item \textbf{Metrics:} Accuracy (main). We report \textbf{Hard-mode results} in the main paper and provide both modes in the supplemental
\end{itemize}

\begin{figure}[h]
  \centering
  \includegraphics[width=\linewidth]{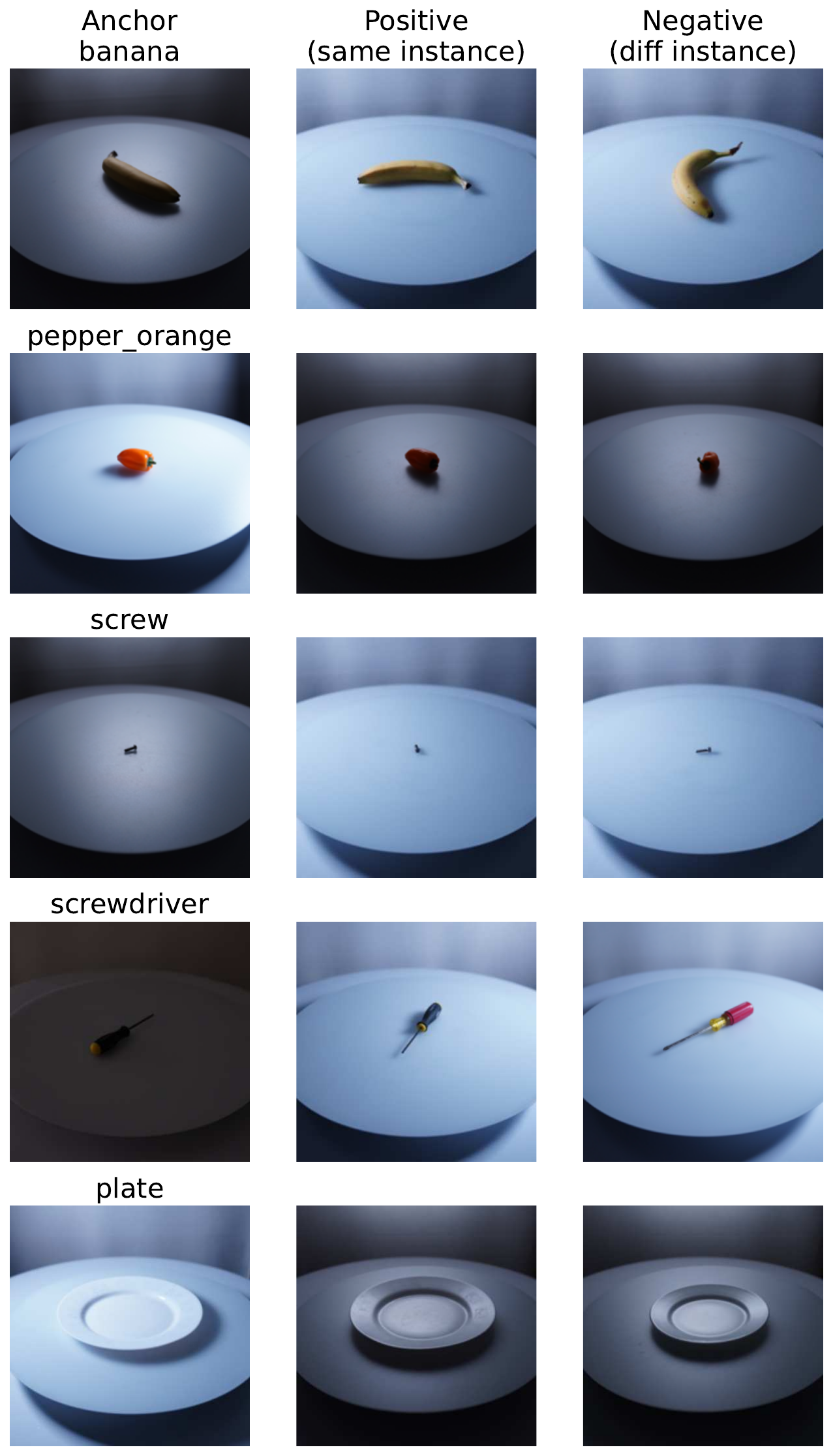}
  \caption{\textbf{CUTE triplets.} Examples of triplets selected for "Hard" mode. The positive and negative examples are drawn from the same scene type, which may be different from the scene type of the anchor, forcing a match across extrinsic characteristics.}
  \label{fig:cute-triplets}
  \vspace{-3mm}
\end{figure}

\subsection*{Subjects2k (Binary Verification for Generative Model)}
\begin{itemize}[leftmargin=2em]
\item \textbf{Task:} Human-validated concept preservation
\item \textbf{Size:} 2,000 pairs
\item \textbf{Labels:} 473 positive; 1,527 negative
\item \textbf{Source:} Curated from Subjects200k using GPT-4V filtering + human annotation
\item \textbf{Protocol:} Predict whether the target preserves the identity of the reference
\item \textbf{Metrics:} AP (main), ROC-AUC
\end{itemize}
\begin{figure}[h]
  \centering
  \includegraphics[width=\linewidth]{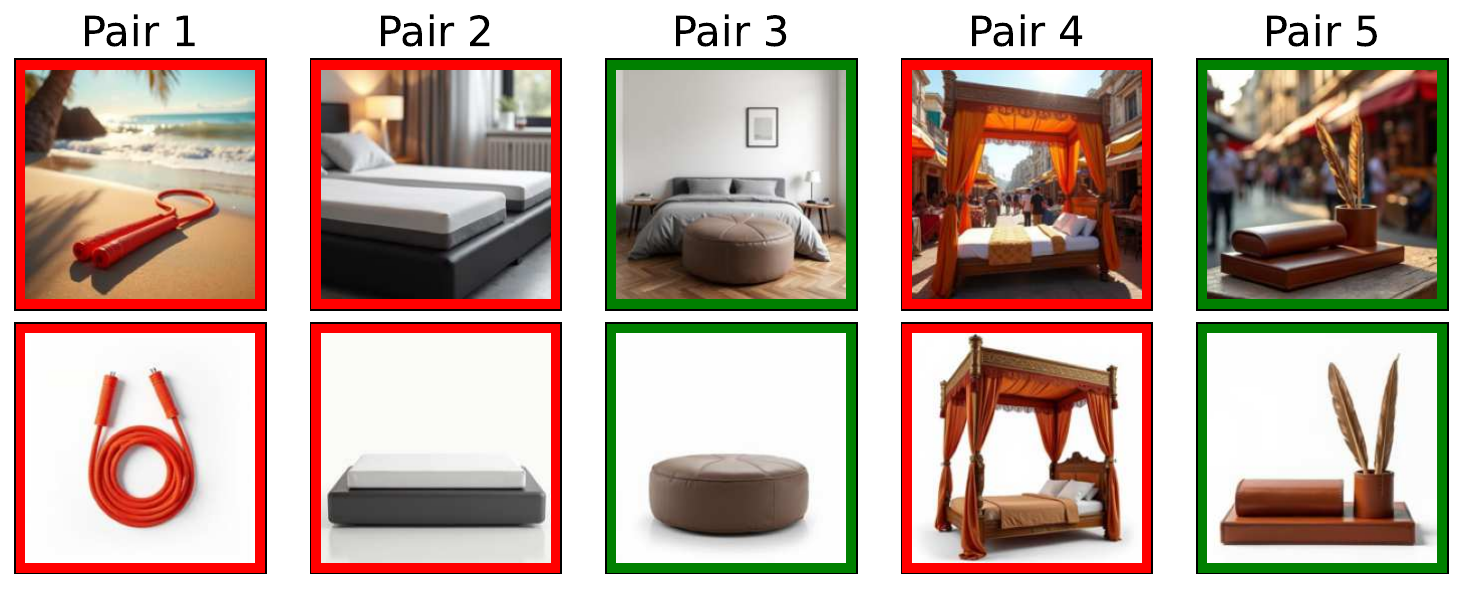}
  \caption{\textbf{Subjects2k Pairs.} Newly annotated 2k subset of Subjects200k \cite{tan2024omini}. Green depicts same instance, red is different.}
  \label{fig:subjects2k}
  \vspace{-3mm}
\end{figure}

\subsection*{DreamBench++ (Discrete Rating for Generative Model)}
\begin{itemize}[leftmargin=2em]
\item \textbf{Task:} Identity preservation in subject-driven image generation
\item \textbf{Size:} 6,921 valid pairs (after filtering)
\item \textbf{Ratings:} Discrete identity score in \([0,4]\)
\item \textbf{References:} 110 reference subjects
\item \textbf{Protocol:} Rank generated images by predicted similarity to the reference
\item \textbf{Metrics:} Spearman correlation (main), Kendall correlation
\end{itemize}

\begin{figure}[h]
  \centering
  \includegraphics[width=\linewidth]{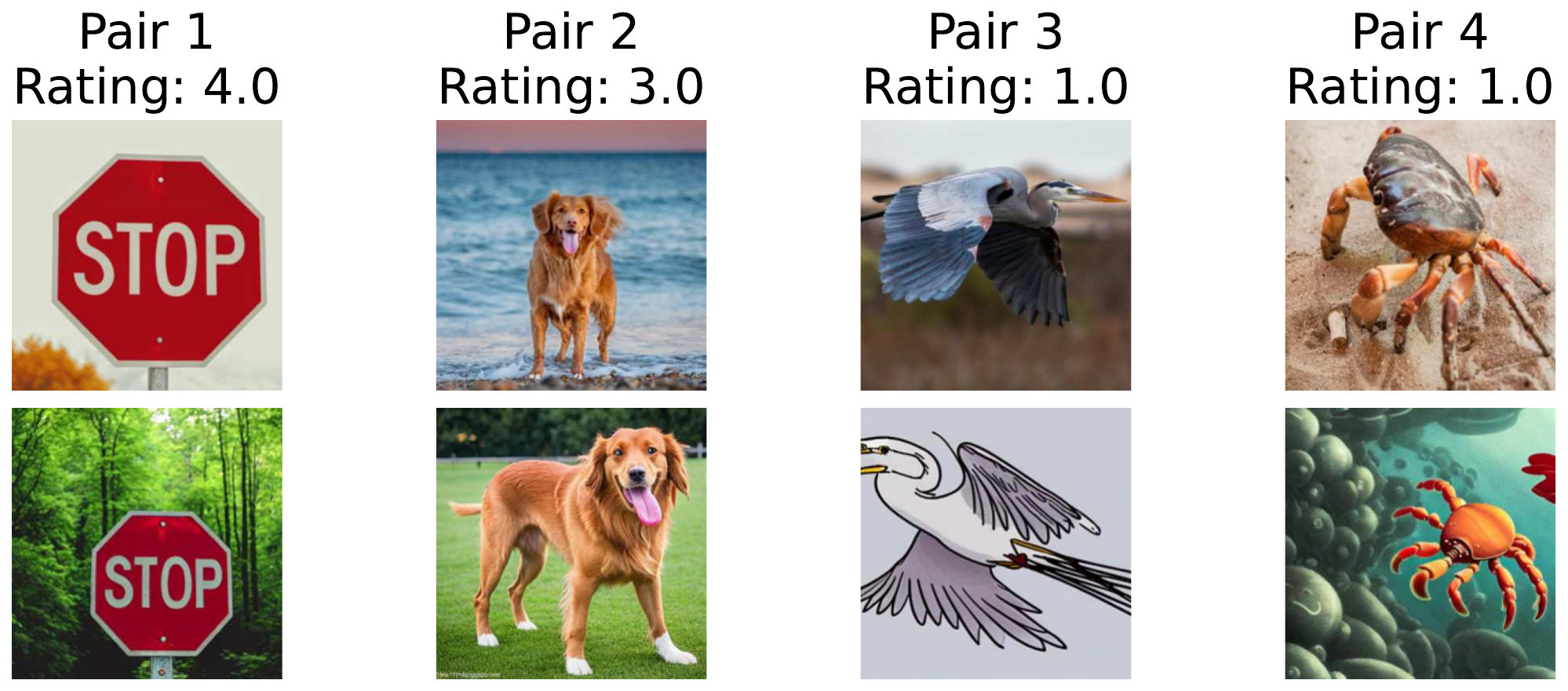}
  \caption{\textbf{DreamBench Pairs.} DreamBench images are accompanied by human annotations out of 4.}
  \label{fig:dreambench}
  \vspace{-3mm}
\end{figure}

\subsection{Subjects2k Human Annotation Pipeline}
\label{supp:subjects2k_pipeline}
\paragraph{Motivation} 
DreamBench++ \cite{peng2024dreambench} is one of the most widely used human benchmark for evaluating concept
preservation in personalized generation, but its annotation design introduces significant noise. Each image receives only two human ratings, and annotators
provide a 0--4 rubric score rather than answering a direct same/different or
pairwise comparison questions. As shown in \Cref{fig:dreambench_limitations}, this results in
both (i) large identity variation among images with identical DreamBench scores,
and (ii) highly variable scores for images with higher identity
similarity. These inconsistencies motivate the need for a cleaner, better-calibrated
evaluation set. We therefore construct \textbf{Subjects2k}, a new human-annotated
subset of Subjects200k \cite{tan2024omini} designed to provide more reliable identity-preservation
judgments.

\paragraph{Subjects2k: Setup.}
Subjects2k is derived from the GPT-annotated \cite{hurst2024gpt}, Flux-generated \cite{flux2024} Subjects200k \cite{tan2024omini} dataset
used for high-fidelity image editing evaluation. Subjects200k provides a
0–5 score per image indicating GPT’s assessment of identity preservation. From
this pool, we construct a balanced human-evaluation subset by sampling
1{,}000 images with GPT score~5, and 200 images from each of the remaining
scores~0-4, yielding 2,000 images total. We built a lightweight web interface
(custom server shown below) and collected human annotations on Prolific.

\begin{figure}[h]
  \centering
  \includegraphics[width=\linewidth]{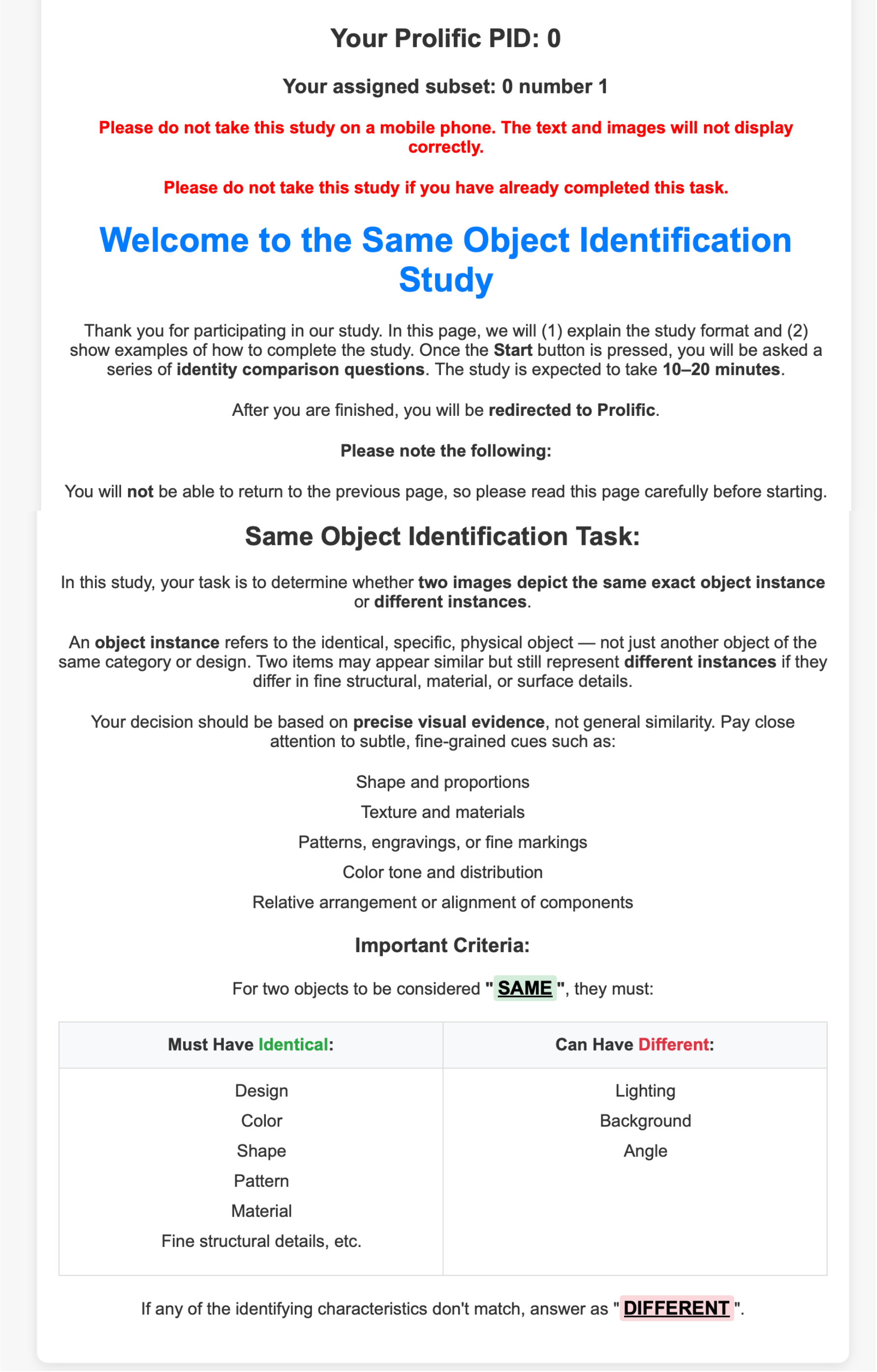}
  \caption{\textbf{Introduction Page for Subjects2k Annotation Server.} We provide a clear definition of an instance to all participants prior to starting their annotations.}
  \label{fig:prolific_landing}
  \vspace{-3mm}
\end{figure}

\begin{figure}[h]
  \centering
  \includegraphics[width=\linewidth]{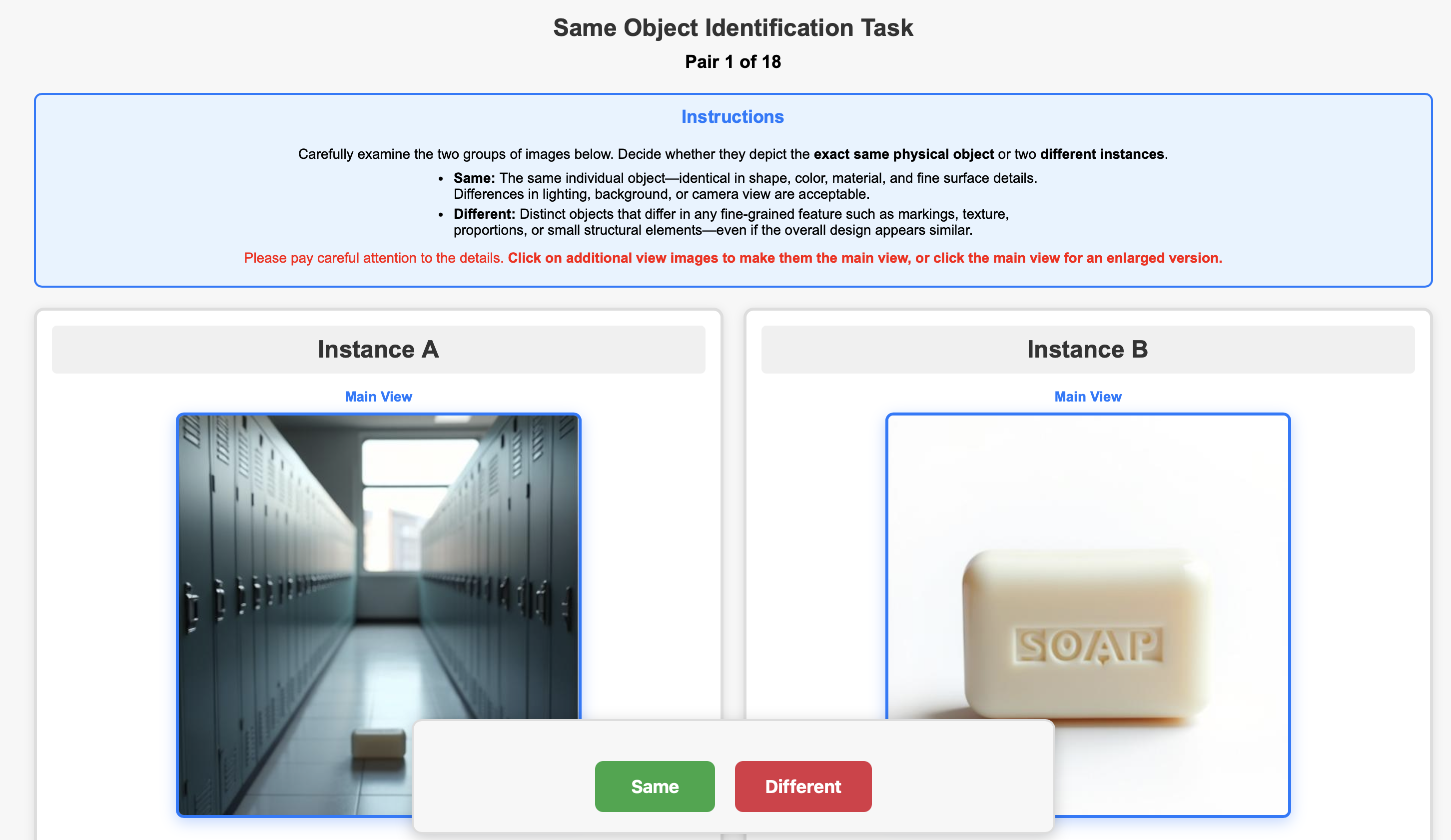}
  \caption{\textbf{Subjects2k Annotation Server Task Page.} Example of a task page for our annotators.}
  \label{fig:prolific_task}
  \vspace{-3mm}
\end{figure}

\paragraph{Subjects2k: Human Annotation Summary}
\label{supp:subjects2k_stats}

We collected human judgments for all 2{,}000 image pairs in Subjects2k and inserted
7 manually-verified sentinel pairs with known ground-truth labels. After each
annotation batch, we filtered annotators by requiring perfect accuracy on all
sentinel questions; responses from any annotator who missed one or more sentinels
were discarded. This procedure ensured a high-quality, reliable annotation set. 
Each pair was annotated in batches: we first obtained labels from three annotators (post-filtering).  
If all three agreed (all ``same'' or all ``different''), we stopped for that pair.  
If there was any disagreement, we collected up to five additional annotations,  
for a maximum of nine annotations per pair.  
This procedure yields an average of $5.01$ annotations per pair  
(min.\ 3, max.\ 9).

\vspace{1mm}
\noindent\textbf{Agreement measure.}
For each pair, let $p$ be the fraction of annotators voting ``same instance''.  
We define agreement as $\max(p, 1-p)$, i.e., the fraction of annotators  
supporting the majority label. Averaged over all 2{,}000 pairs,  
the agreement is $0.864$.

\vspace{1mm}
\noindent\textbf{Continuous labels and binarization.}
For each pair, we define a continuous label
\[
\ell = \frac{\#\text{``same'' votes}}{\#\text{total votes}} \in [0, 1].
\]
The empirical distribution of $\ell$ is summarized in
Table~\ref{tab:subjects2k_label_hist}.  
We then derive a binary label $\mathrm{bin\_label}$ by thresholding at $0.8$:  
pairs with $\ell > 0.8$ are assigned $\mathrm{bin\_label}=1$ (``same''),  
and all others are assigned $\mathrm{bin\_label}=0$ (``different'').  
This yields 1{,}527 negative pairs and 473 positive pairs.

\begin{table}[h]
  \centering
  \small
  \setlength{\tabcolsep}{6pt}
  \renewcommand{\arraystretch}{0.95}
  \caption{\textbf{Subjects2k continuous label histogram.}  
  Counts of image pairs falling into each bin of the average human ``same'' vote
  fraction $\ell$.}
  \label{tab:subjects2k_label_hist}
  \begin{tabular}{lc}
    \toprule
    Label range & \# pairs \\
    \midrule
    $[0.00, 0.09)$ & 788 \\
    $[0.09, 0.18)$ & 94 \\
    $[0.18, 0.27)$ & 111 \\
    $[0.27, 0.36)$ & 96 \\
    $[0.36, 0.45)$ & 141 \\
    $[0.45, 0.55)$ & 70 \\
    $[0.55, 0.64)$ & 143 \\
    $[0.64, 0.73)$ & 68 \\
    $[0.73, 0.82)$ & 105 \\
    $[0.82, 0.91)$ & 77 \\
    $[0.91, 1.00)$ & 307 \\
    \bottomrule
  \end{tabular}
\end{table}

\noindent\textbf{Binary label distribution.}
Using the above threshold, the binary label counts are:
1{,}527 pairs with $\mathrm{bin\_label} = 0$ and
473 pairs with $\mathrm{bin\_label} = 1$.

\begin{figure}[h]
  \centering
  \includegraphics[width=\linewidth]{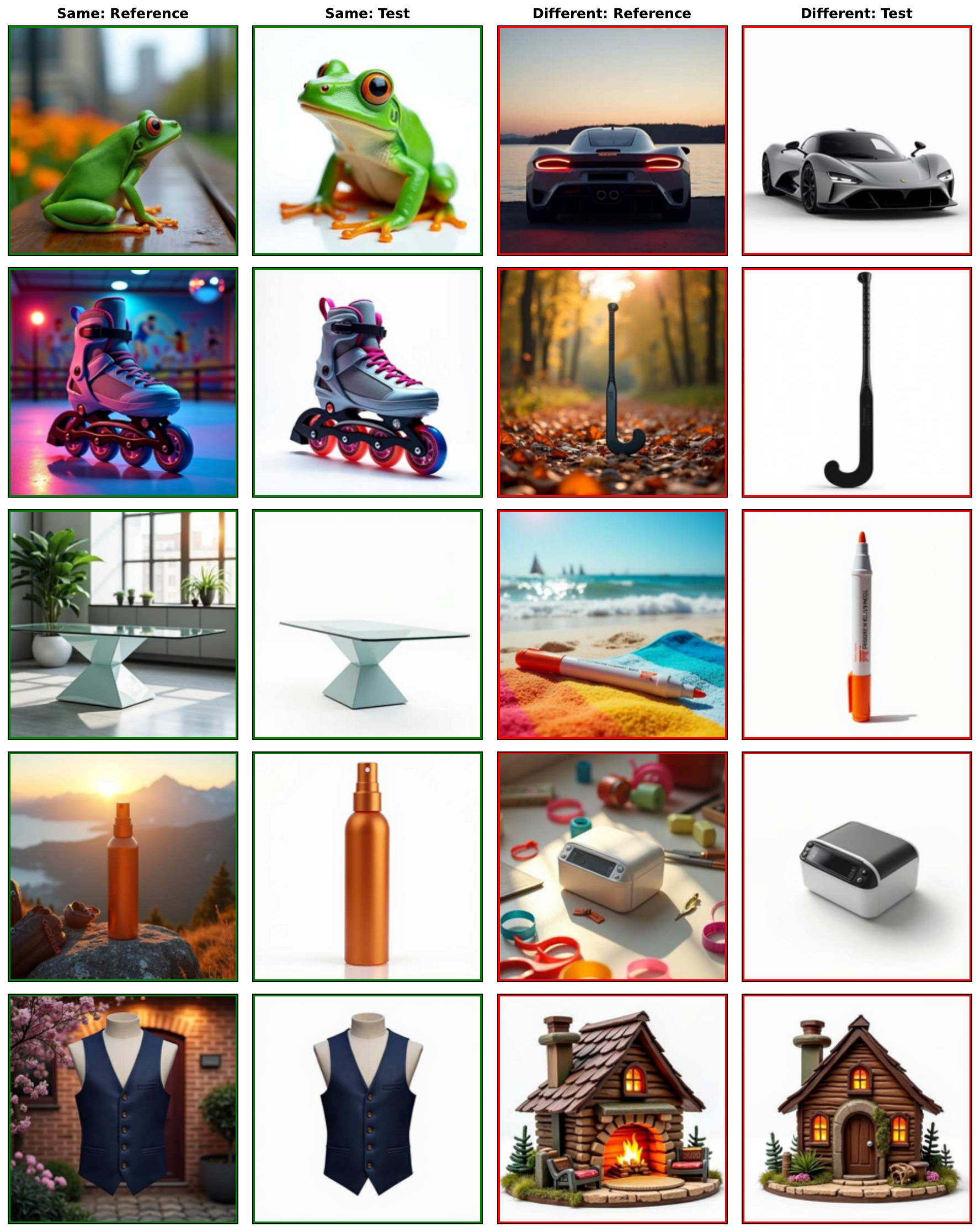}
  \caption{\textbf{Human-labeled identity pairs.} 
\textbf{Left:} Examples annotated as “Same.” 
\textbf{Right:} Examples annotated as “Different.”  
Human annotators reliably pick up subtle, fine-grained cues—such as texture, pattern, and small structural differences.}

  \label{fig:prolific_landing}
  \vspace{-3mm}
\end{figure}

\subsection{MLLM Evaluation Criteria}
\label{supp:mllm_eval}
To ensure fair comparison with MLLMs, we employ structured evaluation protocols including DreamBench++. The MLLM* row of Table~2(a) reports results using the original prompts and models from Subjects200K and DreamBench++, reflecting the annotations provided with the released datasets. In particular, DreamBench++ prompts follow a rubric-based protocol designed for identity-consistency scoring. Although the exact filtering prompts for Subjects200K are not publicly released, the authors describe a rigorous MLLM-based quality control process that explicitly verifies subject consistency. During dataset construction, each generated sample ``underwent five independent evaluations by ChatGPT-4o,'' and ``only images that passed all five evaluations were included'' in the final dataset~\cite{tan2024omini}.

\begin{figure}[t]
\centering
\fbox{
\begin{minipage}{0.95\columnwidth}
\small
\textbf{Prompt.}
You are a visual identity metric. Given two input images, decide if they depict the same instance (e.g., the same animal individual or the same exact object). Focus on stable, instance-specific features and ignore differences due to pose, background, and lighting.

\vspace{4pt}
\textbf{Output format:} Return \textbf{only} a single JSON object with exactly these fields:

\vspace{4pt}
\texttt{\{}\\
\texttt{\ \ "same\_instance": 0 or 1,} // binary decision (1 = same, 0 = different) \\
\texttt{\ \ "confidence": float in [0,1],} // confidence in the decision \\
\texttt{\ \ "similarity": float in [0,1]} // similarity score for ranking/mAP (higher is more similar) \\
\texttt{\}}
\end{minipage}
}
\caption{GPT-Generated prompt used for MLLM standardized evaluation.}
\label{fig:mllm_prompt}
\end{figure}

\section{Results}
\label{supp:results}
\subsection{Dense Results}
\label{supp:dense_results}
We show qualitative results for instance segmentation below, comparing ID-Sim and DINOv3.  

\begin{figure}[h]
  \centering
  \includegraphics[width=\linewidth]{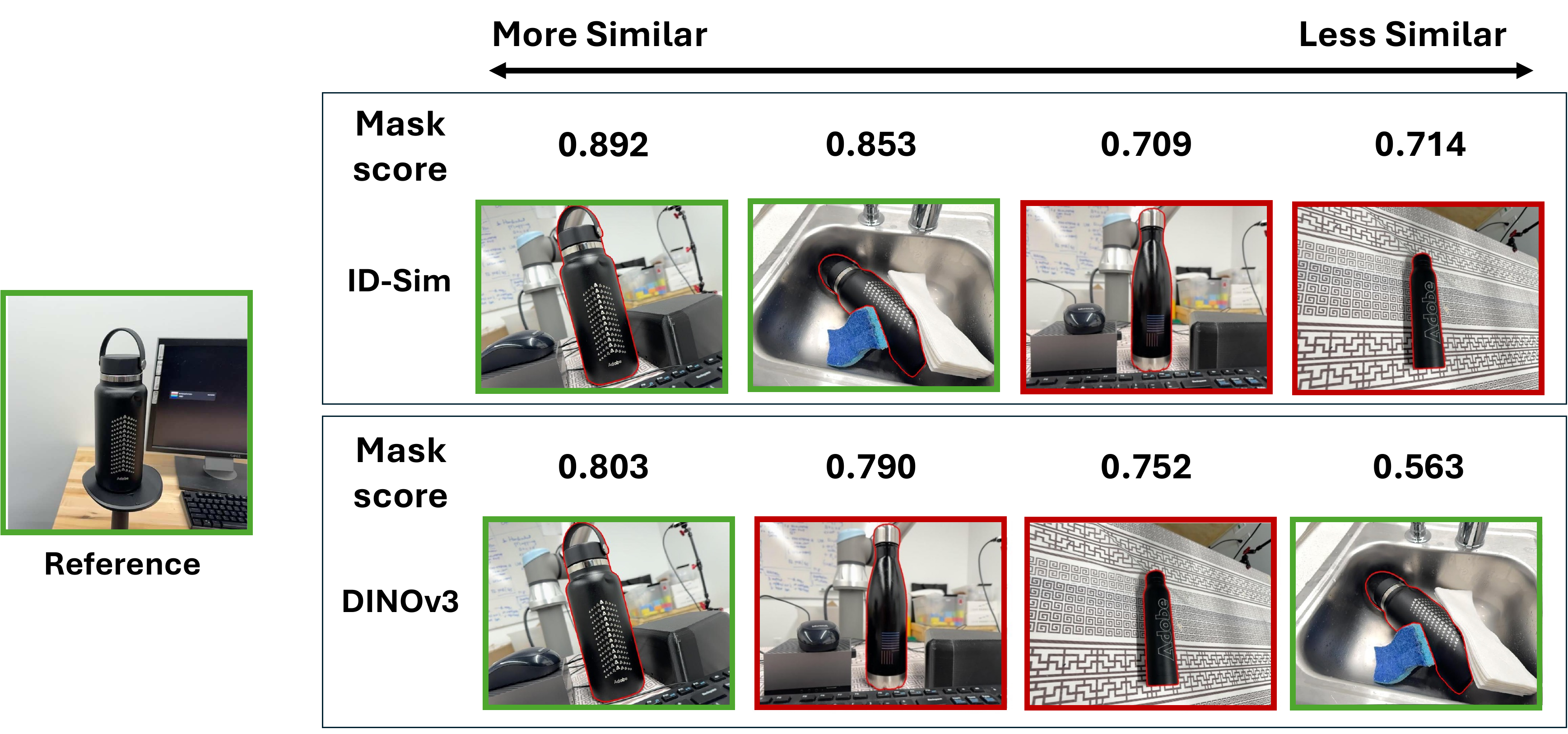}
  \caption{\textbf{Qualitative Results for Per-SAM.} We show predicted segmentation masks and corresponding predicted confidence scores, ordered in highest to lowest with respect to a reference object. First, we observe that when combined with PerSAM, both ID-Sim and DINOv3 are able to produce reliable segmentation mask predictions (mask drawn in red around the instance). However, we observe that ID-Sim is significantly better at recognizing instances across distribution shifts and discriminating fine-grained neighbors compared to DINOv3, as shown by the predicted mask scores and the ordering.}
  \label{fig:heatmap}
  \vspace{-3mm}
\end{figure}

In addition to being useful for spatial tasks, ID-Sim's dense features can be integrated with additional conditioning to extend ID-Sim's capabilities in more complex scenarios. This is clearly demonstrated in \Cref{fig:heatmap}, where we can see that ID-Sim \textit{learns spatially-localized identity features} that remain informative even in multi-entity scenes. This is particularly helpful in resolving ambiguity in multi-entity scenes, which require additional user conditioning to specify which instances the metric should be applied towards. While conditioning is not part of our metric, this shows ID-Sim is naturally compatible with external conditioning signals (e.g., spatial masks or region selection) for specifying user intent.

\begin{figure}[h]
  \centering
  \includegraphics[width=\linewidth]{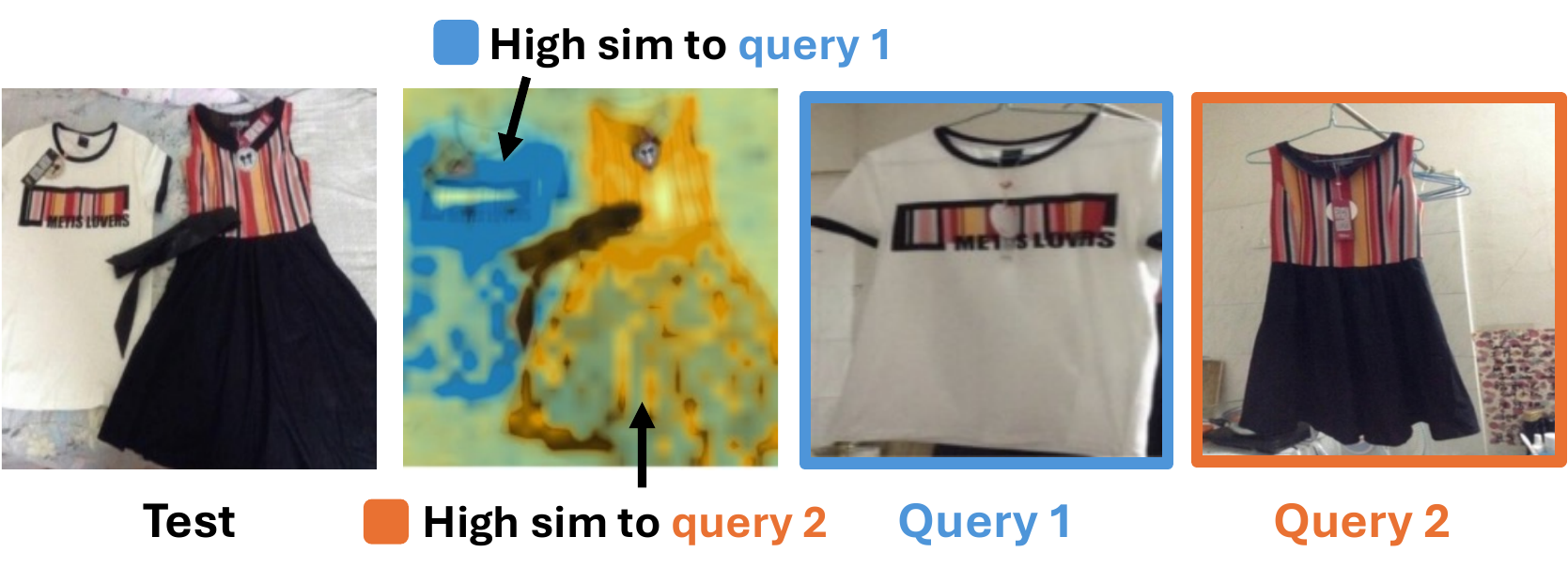}
  \caption{\textbf{Dense masks can resolve ambiguities in multi-object Scenes.} Given the test image with 2 shirts (left), ID-Sim features are sensitive to the identity of the query image (right 2 images), evidenced by the patch-level similarity heatmaps (2nd to left).}
  \label{fig:heatmap}
  \vspace{-3mm}
\end{figure}

\begin{figure*}[t]
\centering

\vspace{10mm}
\begin{minipage}{0.95\textwidth}
  \centering
  \includegraphics[width=\linewidth]{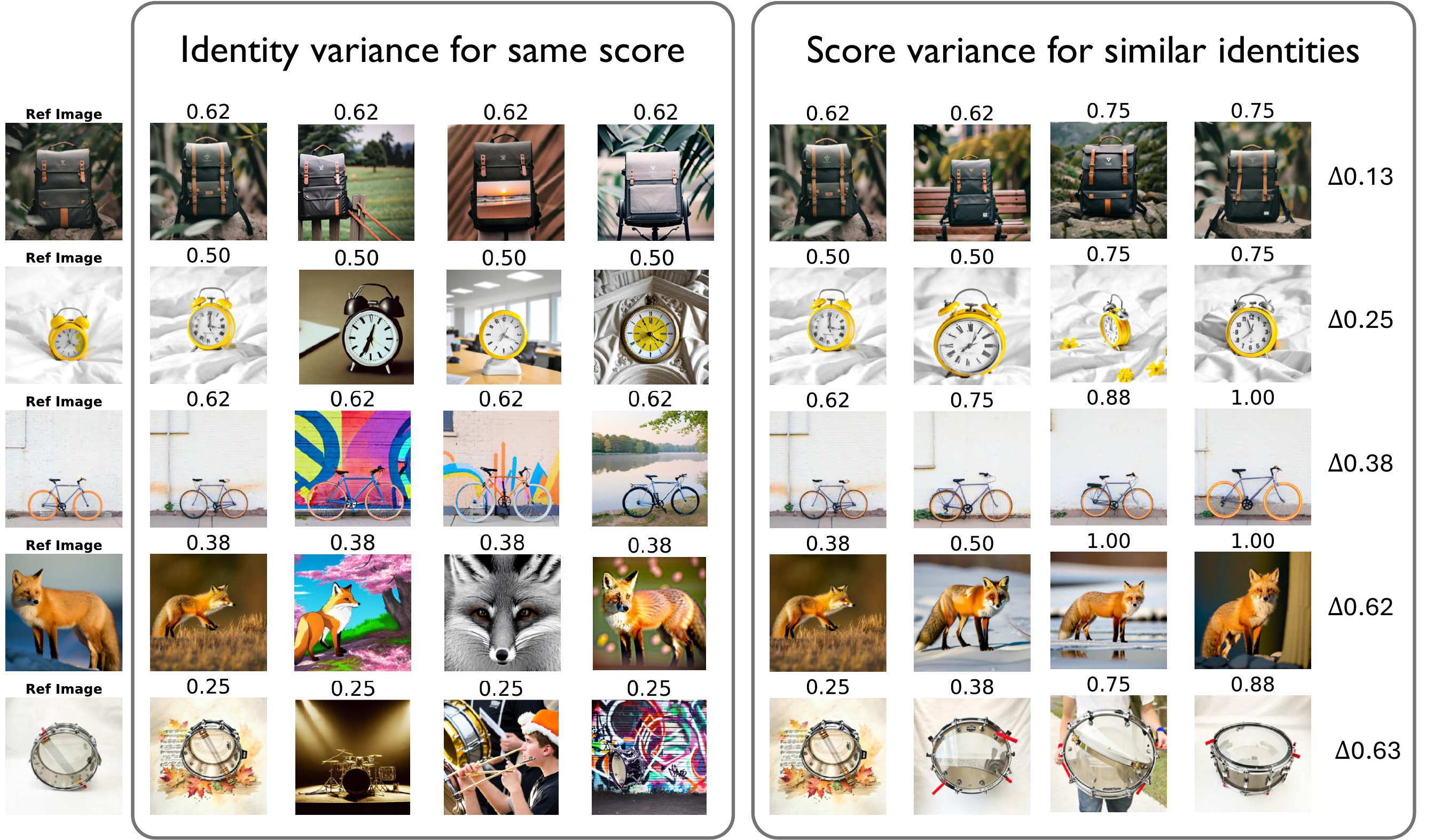}
  \caption{\textbf{Limitations of DreamBench++ annotations.}  
    DreamBench++ assigns only two human rubric scores (0--4) per image, which leads to substantial noise in concept-preservation evaluation. As shown above, (i) images with the same DreamBench score can exhibit large variation in identity similarity, and (ii) images with high identity similarity may still receive widely different DreamBench scores. These inconsistencies highlight the need for a cleaner and more discriminative human benchmark, motivating the construction of our Subjects2k dataset.}
  \label{fig:dreambench_limitations}
\end{minipage}

\vspace{4mm} %

\begin{minipage}{0.95\textwidth}
  \centering
  \scriptsize
  \setlength{\tabcolsep}{5pt}
  \renewcommand{\arraystretch}{1.35}

  \begin{tabular}{l
    ccc      %
    c        %
    c        %
    cc       %
    cc       %
    cc       %
    cc}      %
    \toprule
    
    & \multicolumn{3}{c}{\textbf{PODS}}
    & \multicolumn{1}{c}{\textbf{DeepFashion2}}
    & \multicolumn{1}{c}{\textbf{AerialCattle}}
    & \multicolumn{2}{c}{\textbf{PetFace}}
    & \multicolumn{2}{c}{\textbf{CUTE}}
    & \multicolumn{2}{c}{\textbf{SS200k}}
    & \multicolumn{2}{c}{\textbf{DreamBench}} \\
    
    \cmidrule(lr){2-4}
    \cmidrule(lr){5-5}
    \cmidrule(lr){6-6}
    \cmidrule(lr){7-8}
    \cmidrule(lr){9-10}
    \cmidrule(lr){11-12}
    \cmidrule(lr){13-14}
    
    \textbf{Model}
    & AUC & AP AUC & nDCG
    & mAP
    & mAP
    & AUC & AP AUC
    & Easy Acc & Hard Acc
    & AUC & AP AUC
    & Spear. & Kend. \\
    \midrule
    
    \multicolumn{14}{l}{\textit{Foundation models}} \\
    
    DINOv3      
    & 0.929 & 0.424 & 0.744
    & 0.519
    & \underline{0.516}
    & 0.879 & 0.884
    & \underline{0.827} & \underline{0.813}
    & 0.642 & 0.323
    & 0.576 & 0.437 \\
    
    CLIP        
    & 0.862 & 0.294 & 0.656
    & 0.408
    & 0.368
    & 0.754 & 0.776
    & 0.779 & 0.687
    & 0.594 & 0.296
    & 0.625 & 0.478 \\
    
    OpenCLIP    
    & 0.887 & 0.359 & 0.705
    & 0.488
    & 0.430
    & 0.753 & 0.772
    & 0.796 & 0.699
    & 0.584 & 0.294
    & 0.666 & 0.516 \\
    \midrule
    
    \multicolumn{14}{l}{\textit{Perceptual similarity models}} \\
    
    DreamSim    
    & 0.897 & 0.317 & 0.672
    & 0.529
    & 0.593
    & 0.814 & 0.824
    & 0.770 & 0.734
    & 0.603 & 0.289
    & \textbf{0.716} & \textbf{0.561} \\
    
    LPIPS       
    & 0.603 & 0.067 & 0.387
    & 0.309
    & 0.442
    & 0.752 & 0.769
    & 0.651 & 0.625
    & 0.483 & 0.235
    & 0.482 & 0.354 \\
    \midrule
    
    \multicolumn{14}{l}{\textit{Instance retrieval model}} \\
    
    UNED        
    & \underline{0.944} & \underline{0.671} & \underline{0.871}
    & \underline{0.714}
    & 0.468
    & 0.784 & 0.800
    & 0.815 & 0.777
    & \underline{0.654} & \underline{0.356}
    & 0.672 & 0.523 \\
    \midrule
    
    \multicolumn{14}{l}{\textit{Ours}} \\
    
    ID-Sim 
    & \makecell{\textbf{0.9642} \\ {\tiny$\pm$ 0.0035}}
    & \makecell{\textbf{0.7727} \\ {\tiny$\pm$ 0.0106}}
    & \makecell{\textbf{0.9161} \\ {\tiny$\pm$ 0.0050}}
    & \makecell{\textbf{0.8045} \\ {\tiny$\pm$ 0.0119}}
    & \makecell{\textbf{0.6786} \\ {\tiny$\pm$ 0.0123}}
    & \makecell{\textbf{0.9002} \\ {\tiny$\pm$ 0.0072}}
    & \makecell{\textbf{0.8958} \\ {\tiny$\pm$ 0.0101}}
    & \makecell{\textbf{0.8887} \\ {\tiny$\pm$ 0.0077}}
    & \makecell{\textbf{0.8559} \\ {\tiny$\pm$ 0.0124}}
    & \makecell{\textbf{0.7053} \\ {\tiny$\pm$ 0.0048}}
    & \makecell{\textbf{0.4113} \\ {\tiny$\pm$ 0.0060}}
    & \makecell{\underline{0.6856} \\ {\tiny$\pm$ 0.0103}}
    & \makecell{\underline{0.5305} \\ {\tiny$\pm$ 0.0100}} \\
    \bottomrule
    
    \end{tabular}
    \vspace{2mm}
    \caption{\textbf{Full quantitative comparison across all benchmarks.}  
We report complete numerical results for all datasets and baselines. For ID-Sim, we show mean~$\pm$~standard deviation over 10 independent training runs. All evaluations use the CLS embedding at inference, consistent with the main paper.}
\label{tab:global_results}

\end{minipage}

\end{figure*}

\subsection{Full results}
\label{supp:full_results}
In \Cref{tab:global_results}, we report full numerical results across all datasets and baselines. Beyond the metrics shown in the main paper, we include additional evaluation metrics and settings for several benchmarks. For ID-Sim, we also report variance over 10 independent runs, each trained with a different random seed.

\section{Analysis}
\label{supp:sensitivity}

We use 100 held-out object instances from MVImgNet, a multi-view dataset that does not appear in any training or evaluation set. For each object, we generate a dense grid of edited images that vary jointly along identity change and one additional dimension (background, viewpoint, or lighting). This provides controlled perturbations for measuring how similarity scores behave under specific visual changes.

\paragraph{Per-instance regression.}
For each instance, we fit a linear model to all similarity scores \(\{\text{sim}_i\}\):
\[
\text{sim}_i = \beta_0
\;+\; \beta_1 \,\text{factor-change}_i
\;+\; \beta_2 \,\text{identity-change}_i
\;+\; \varepsilon_i.
\]
Sensitivity to each dimension is defined as the negative slope \((-\beta_1\) or \(-\beta_2)\), which gives the amount of similarity reduction per unit change. Because the regression uses all points in the joint edit grid, it produces stable directional sensitivity estimates while avoiding the noise that arises when using only axis-restricted slices. We also record the regression \(R^2\) value for each instance.

\paragraph{Aggregation across objects.}
Dataset-level sensitivities are obtained by averaging per-instance slopes across the 100 MVImgNet objects. The same grid construction, regression fitting, and aggregation are applied independently to each evaluated model.

\paragraph{Bootstrap uncertainty.}
To estimate uncertainty, we perform bootstrap resampling over object identities. In each of 1,000 bootstrap iterations, we sample the 100 instances with replacement, recompute all regression coefficients, and compute the mean sensitivity for that resample. For each model and each dimension, we report:
\begin{itemize}[leftmargin=6mm]
    \item the bootstrap mean,
    \item the bootstrap standard deviation,
    \item the 95\% confidence interval (2.5 to 97.5 percentile).
\end{itemize}
These intervals capture variability across object identities and provide a reliable measure of uncertainty for the estimated sensitivities.

\subsection{Analysis Image Generation}
\label{supp:analysis_image_gen}
We generate three types of edits (identity, lighting, and background) using dedicated Qwen-Image-Edit pipelines. These images are used only for sensitivity analysis and are fully separate from all training data.

\paragraph{Identity edits.}
For each anchor image:
\begin{itemize}[leftmargin=6mm]
    \item Qwen-Image-Edit operates in inpainting mode over the foreground mask,
    \item We use seven edit strengths \(0.4, 0.5, 0.6, 0.7, 0.8, 0.9, 1.0\) to produce a graded sequence of identity change,
    \item The prompt instructs Qwen to alter internal appearance while preserving overall structure and silhouette,
    \item Only the foreground region is modified while the background remains unchanged,
    \item A fixed seed is used for reproducibility.
\end{itemize}

\paragraph{Lighting edits.}
Lighting variations are generated with global edits (no masking):
\begin{itemize}[leftmargin=6mm]
    \item Eight lighting prompts (shown below)
    \item Prompts adjust illumination, color temperature, and shading while keeping geometry and texture fixed,
    \item Qwen-Image-Edit is run with 8 inference steps and a fixed seed.
\end{itemize}

\paragraph{Background edits.}
Background replacements are created with mask-guided editing:
\begin{itemize}[leftmargin=6mm]
    \item The background is removed using a mask and replaced with a white canvas prior to editing,
    \item Eleven background prompts of varying intensity (see below)
    \item The prompt specifies that only background pixels may change and that the object must remain unchanged in geometry, pose, and fine appearance,
    \item Qwen adjusts shading to maintain foreground and background consistency,
    \item Deterministic seeds produce reproducible outputs.
\end{itemize}

These edit types provide controlled and interpretable variations for quantifying how models respond to identity changes, contextual changes, and illumination changes.

\paragraph{Prompt sets used for analysis.}
For completeness, we list the exact background and lighting prompts used to generate the edit grids described in this section. These prompts correspond directly to the options indexed in our code and are referenced when constructing the background--identity grid, the lighting--identity grid, and the viewpoint--identity grid.

\subparagraph{Background prompts (11).}
\begin{enumerate}[leftmargin=6mm, itemsep=1mm]
    \item Soft matte off-white plaster wall with subtle imperfections and even diffused daylight.
    \item Coastal scene with overcast bright daylight, pale sandy boardwalk, soft gray ocean, and light cloudy sky.
    \item Contemporary office interior with white walls, light wood, glass partitions, and soft diffuse daylight.
    \item Indoor greenery in white pots with bright filtered daylight from a large window and light walls.
    \item Urban street wall with faded or pastel graffiti on light concrete under overcast daylight.
    \item Artistic studio with neutral-toned canvases, minimal paint splatter, and soft shadow-free daylight.
    \item Bright modern kitchen with white or light-gray surfaces, minimal decor, and soft natural daylight.
    \item Minimalist boutique or gallery space with light walls, wood floors, neutral displays, and even ambient lighting.
    \item Sunlit desert landscape with pale sand, warm beige rock formations, and soft shadows under clear daylight.
    \item Industrial loft with light-exposed brick, large windows with daylight, and pale metal beams.
    \item Warm-toned library interior with light wood shelves, muted books, and soft warm ambient lighting.
\end{enumerate}

\subparagraph{Lighting prompts (8).}
\begin{enumerate}[leftmargin=6mm, itemsep=1mm]
    \item Neutral, balanced front lighting with soft shadows and natural highlights (reference lighting).
    \item Warm front-directional lighting with soft elongated shadows and a gentle amber color cast.
    \item Strong directional side lighting with pronounced contrast between lit and shaded regions.
    \item Bright neutral-cool lighting with soft-edged shadows and crisp highlights.
    \item Extremely soft front lighting with faint highlights and very low-contrast shadows.
    \item Bright front lighting with well-defined shadows and accentuated surface detail.
    \item Very low-level front illumination that preserves shape and color with shadow dominance.
    \item Intense front lighting with high brightness, strong highlights, and deep detailed shadows.
\end{enumerate}

\begin{figure*}[t]
  \centering
  \includegraphics[width=\textwidth]{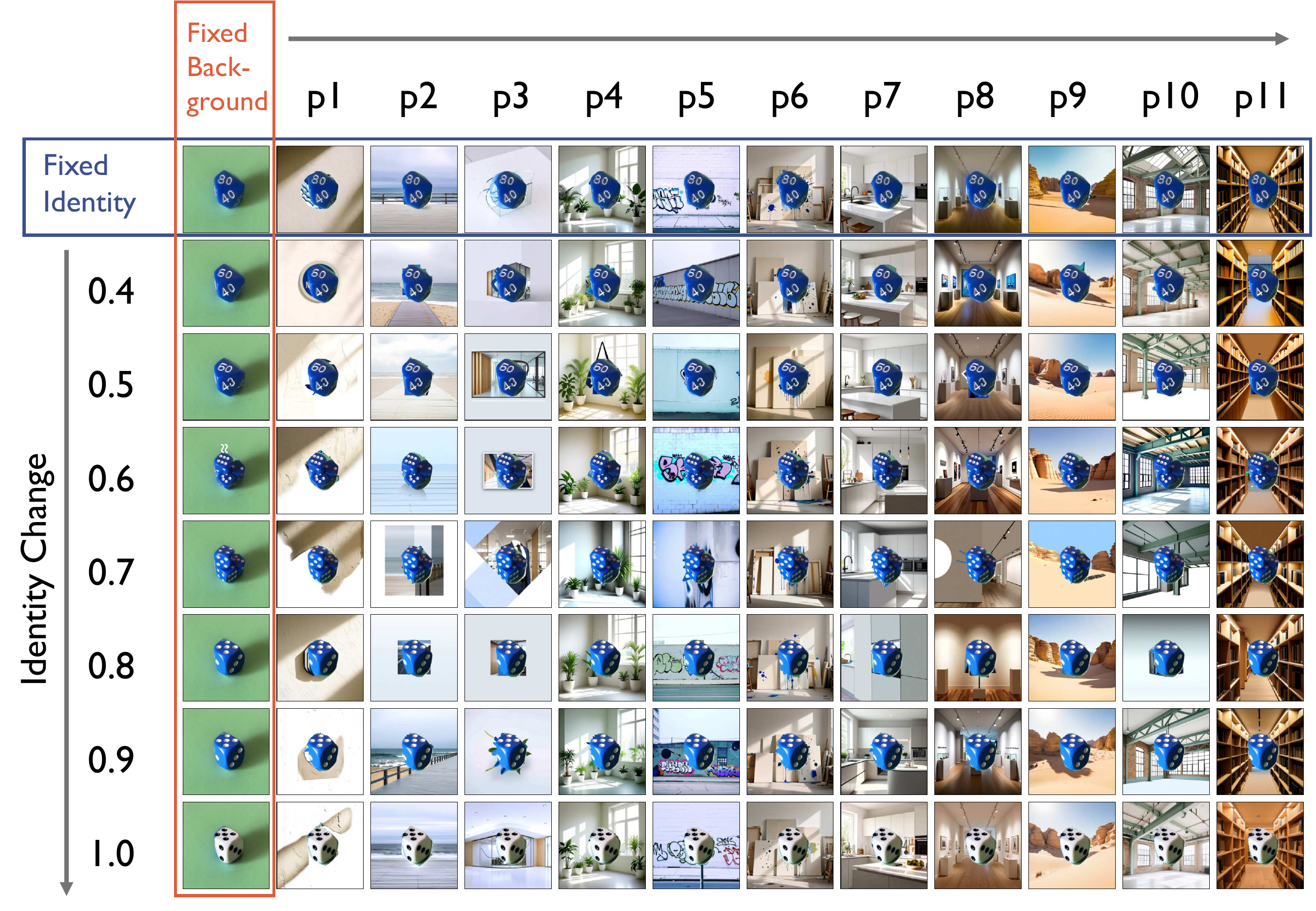}
  \caption{\textbf{Background vs.\ Identity Variation Grid.} 
Rows vary the foreground identity through Qwen-Edit inpainting at increasing edit strengths, while columns vary the scene background using inpainting prompts. Each cell shows the similarity of the edited image to the original anchor. This grid isolates how models respond jointly to identity changes and background shifts.}
  \label{fig:background_grid}
  \vspace{-3mm}
\end{figure*}

\begin{figure*}[t]
  \centering
  \includegraphics[width=\textwidth]{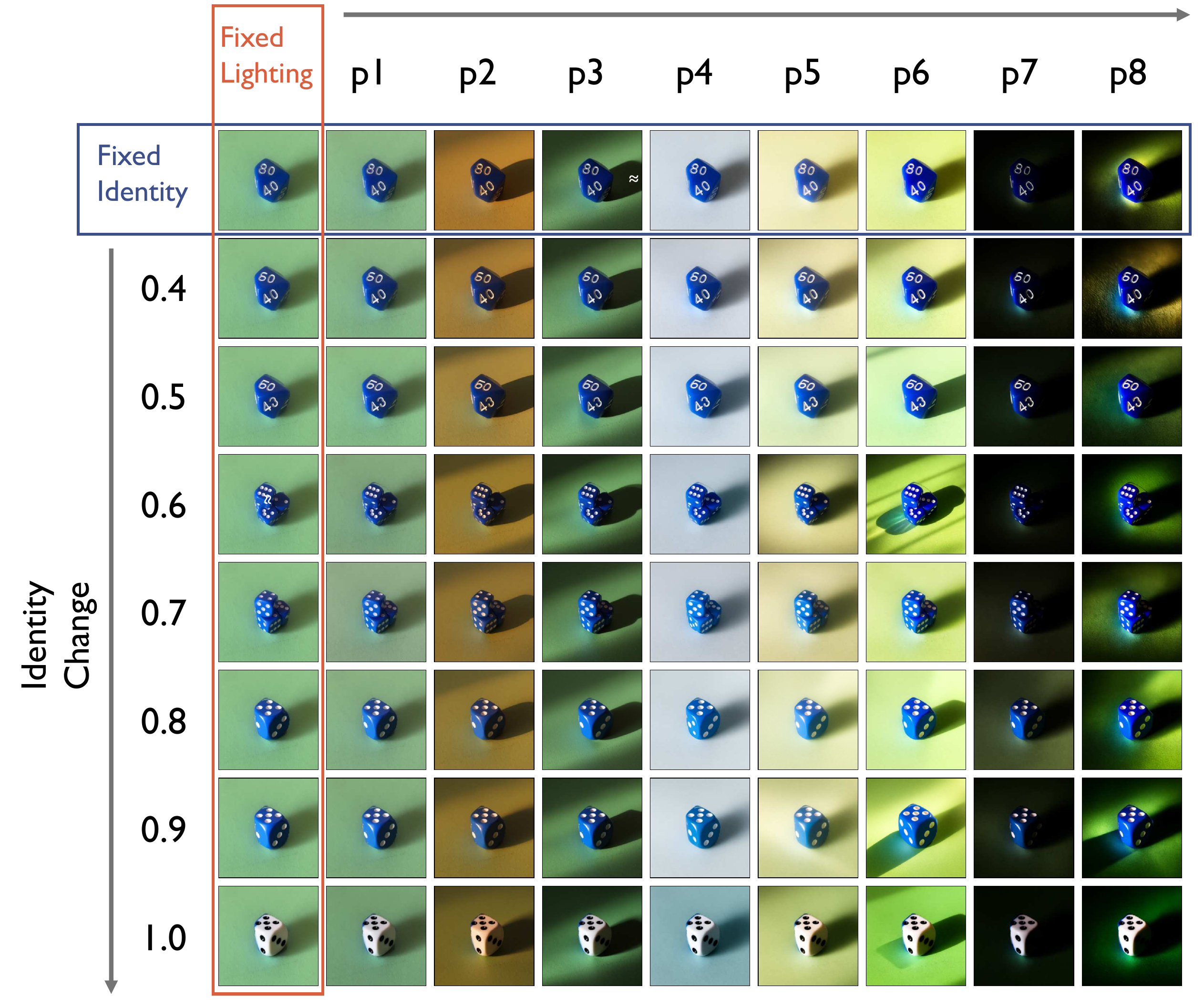}
  \caption{\textbf{Lighting Variation Grid.} 
    Rows correspond to increasing levels of identity change, while columns apply eight different lighting edits using Qwen-Edit. This grid tests whether models remain stable under illumination changes while remaining sensitive to small identity perturbations.}
  \label{fig:lighting_grid}
  \vspace{-3mm}
\end{figure*}

These prompts define the discrete levels of background variation and illumination used to construct the edit grids in Figures~\ref{fig:background_grid} and  \ref{fig:lighting_grid}.  
They are applied consistently across all MVImgNet objects to ensure comparable and fully reproducible sensitivity measurements.

\clearpage
\begin{figure*}[t]
  \centering
  \includegraphics[width=\textwidth]{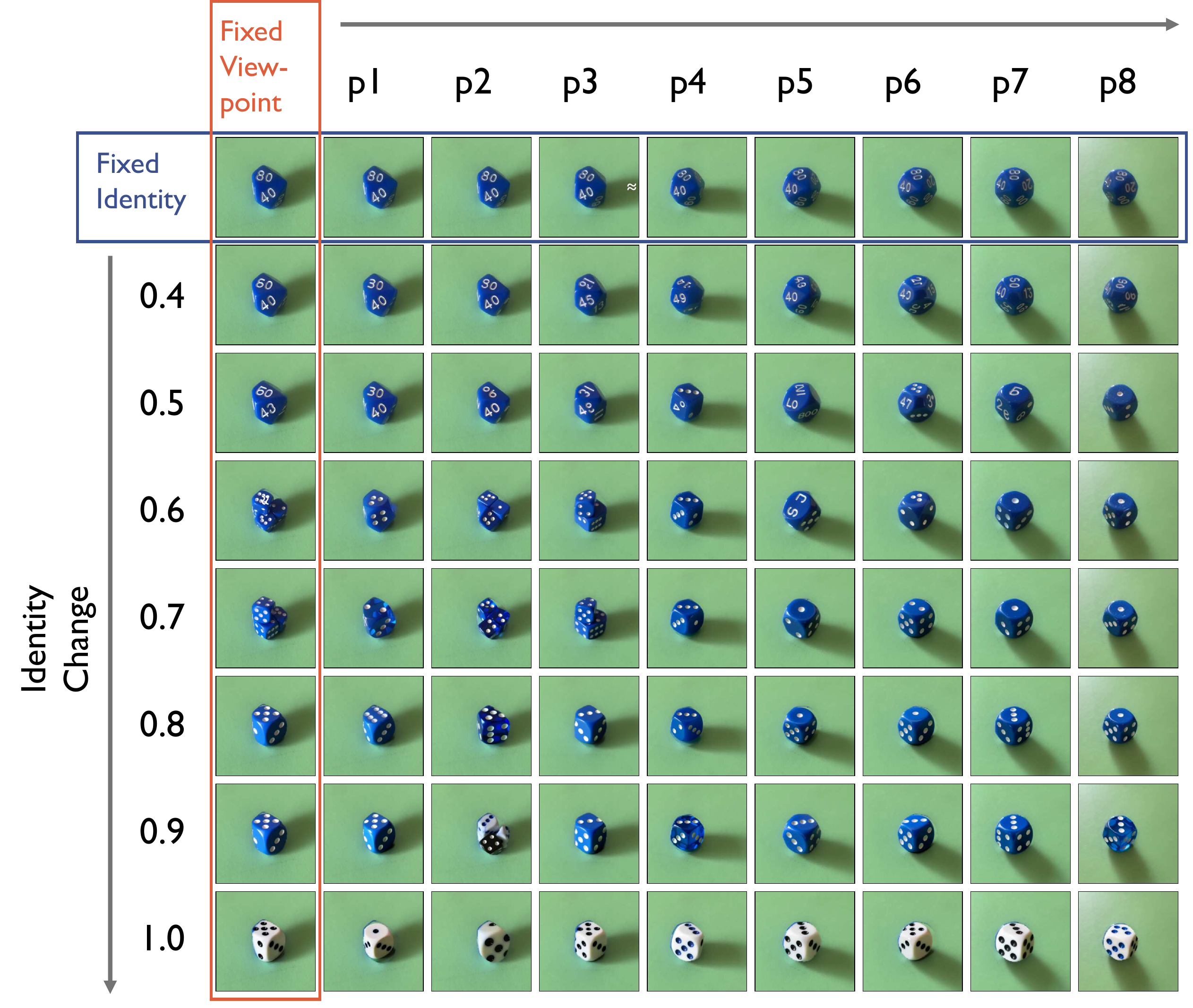}
  \caption{\textbf{Viewpoint Variation Grid.} 
    Rows vary identity strength and columns sweep natural viewpoint changes using the multi-view MVImgNet sequence. This grid evaluates how well each model maintains invariance to viewpoint while still detecting identity-altering edits.}
  \label{fig:viewpoint_grid}
  \vspace{-3mm}
\end{figure*}

\clearpage

\end{document}